%% file: sn-article.tex
\documentclass[11pt]{article}
\usepackage{mmedagent}
\usepackage{mathpazo}
\usepackage[most]{tcolorbox}
\usepackage{authblk}
\usepackage{xcolor}
\usepackage{fancyhdr}
\usepackage[T1]{fontenc}
\usepackage{ltablex}
\usepackage{hyperref}
\keepXColumns
\definecolor{myorange}{RGB}{252,129,59}

\usepackage{graphicx}
\usepackage{caption}

\usepackage{amsfonts}
\usepackage[title]{appendix}
\usepackage{textcomp}
\usepackage{listings}
\usepackage{tabularx}
\usepackage{xspace}

\usepackage{array}
\usepackage{xurl}
\usepackage{hyperref}

\newcolumntype{Y}{>{\raggedright\arraybackslash}X}
\newcolumntype{P}[1]{>{\raggedright\arraybackslash}p{#1}}

\title{
\raisebox{-0.15em}{%
    \includegraphics[height=2.5em]{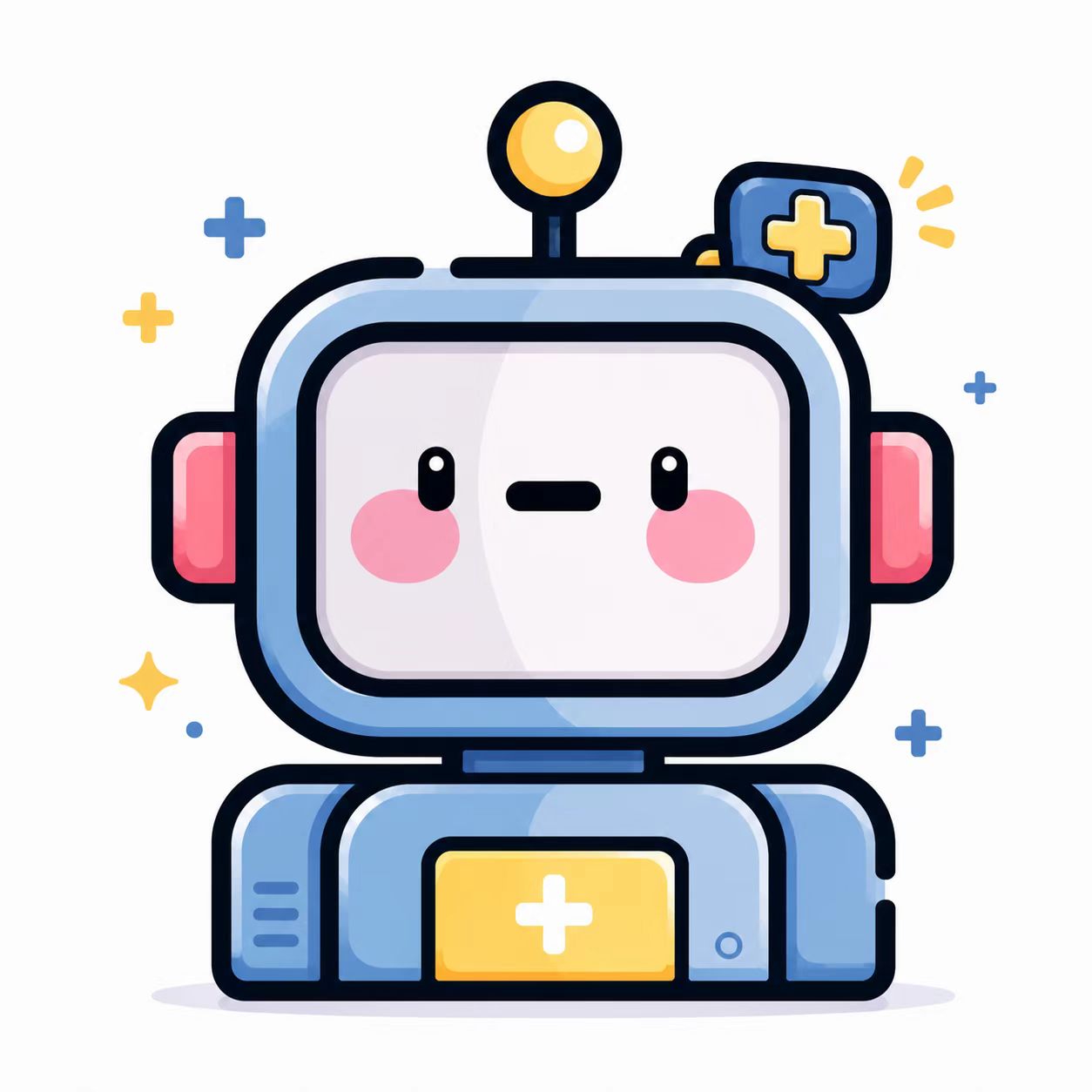}%
  }\hspace{0.1em}%
\textbf{MMedAgent-Plus: Scaling Multimodal Medical \\ Agents across Modalities, Tasks and Workflows}}

\author[1,2]{Ruoxin Wang}
\author[2,3]{Yuanting Pan}
\author[2,4]{Binxu Li}
\author[5]{Ruiyang Ji}
\author[1]{Yaqian Chen}
\author[6]{Jiayuan Ding}
\author[7]{Qifan Wang}
\author[8]{Lifu Huang}
\author[9]{Mingjie Li}
\author[2]{Yu Zhang}
\author[2]{Ehsan Adeli}
\author[2]{Kilian Pohl}
\author[10]{Zhe Xu}
\author[8,*]{Zihao Lin}
\author[1,*]{Haoyu Dong}
\author[2,*]{Yixin Wang}
\affil[ ]{
\textsuperscript{1}Duke University,
\textsuperscript{2}Stanford University,
\textsuperscript{3}Yale University,
\textsuperscript{4}Princeton University,
\textsuperscript{5}Virginia Tech,
\textsuperscript{6}Hippocratic AI,
\textsuperscript{7}Meta,
\textsuperscript{8}UC Davis,
\textsuperscript{9}Shanghai Jiao Tong University,
\textsuperscript{10}Tongji University
}

\date{}

\begin{document}
\maketitle
\vspace{-10mm}

\begingroup
\renewcommand\thefootnote{}\footnotetext{\footnotesize \textsuperscript{*}Corresponding authors}%
\renewcommand\thefootnote{}\footnotetext{\footnotesize Emails: \href{mailto:ruoxinwaaang@gmail.com}{ruoxinwaaang@gmail.com},
\href{mailto:yxinwang@stanford.edu}{yxinwang@stanford.edu}}%
\addtocounter{footnote}{-1}%
\endgroup

\begin{center}
\href{https://mmedagent.org}{
  \hspace{0.1em}\textcolor{myorange}{Project Website: https://mmedagent.org}%
}
\end{center}

\begin{abstract}
{\normalsize
\input{sections/Abstract}

}
\end{abstract}

\begin{center}
{\small\textbf{Keywords:} Agent, Instruction Tuning, Vision-Language Model, Multimodal Medicine}
\end{center}

\input{sections/Main}
\input{sections/Results}

\input{sections/Discussion}
\input{sections/Methods}

\input{sections/Data_Availability}

\input{sections/Code_Availability}

\bibliographystyle{unsrt}
\bibliography{sn-bibliography}

\newpage
\input{sections/Supplementary}

\end{document}

%% file: sections/Abstract.tex
In healthcare, medical tasks often involve multiple interdependent steps, where each outcome informs subsequent decisions. Specialist AI models achieve state-of-the-art performance on individual tasks, whereas generalist vision-language models cover more tasks but typically lag behind specialists. Neither is designed to decide which analyses are needed, execute them in order, or pass intermediate results between steps. We proposed \textsc{MMedAgent-Plus}, a multimodal medical tool-use agent that couples a vision-language planner with a library of 18 specialist models spanning 12 medical tasks and 8 modalities, each selected as a leading publicly available model for its task. Rather than approximating expert capabilities within a single model, the planner learns to orchestrate these specialists, deciding which to call and in what order, and to generate a final response grounded in their outputs. All tools share a unified interface declaring its inputs, parameters, and outputs, so that models with different architectures are invoked in one call format. The backbone planner is instruction-tuned on 208,928 samples curated from 32 public medical benchmarks, spanning diverse tool-use scenarios involving single or multiple tools and varying numbers of interaction rounds. We evaluate tool selection, response synthesis, and end-to-end workflow execution separately on held-out evaluation samples. \textsc{MMedAgent-Plus} selects the correct specialist for 91.0\%, 98.1\%, and 80.7\% of individual tool calls in these three settings, respectively. Given correct tool outputs, \textsc{MMedAgent-Plus} achieves scores of 14.4-14.8 out of 15 for faithfulness, completeness, and conciseness across 600 evaluation samples. In end-to-end execution without pre-supplied tool outputs, \textsc{MMedAgent-Plus} correctly invokes the required specialist in 90.6\% of 350 evaluation cases spanning seven tasks. Its final-answer accuracy reaches 0.98, matching the dedicated leading specialist and exceeding the strongest baseline (0.28).

%% file: sections/Main.tex
\section{Main}

Over the past decade, deep learning has achieved expert-level performance on selected medical tasks, including radiograph interpretation~\cite{rajpurkar2017chexnet}, histopathology image classification~\cite{lu2024visual}, and anatomical structure segmentation~\cite{ma2024segment}. These advances have largely been driven by specialist models~\cite{moor2023foundation,zhang2024ultramedical} designed for specific modalities and tasks. However, real-world clinical workflows often require multiple interdependent analyses rather than isolated predictions~\cite{tu2024towards,sounderajah2025stard}.For example, a radiologist may identify and characterize suspicious lesions before integrating the findings into a diagnostic report. Similarly, a clinician may assess symptoms and medical history, select diagnostic tests, interpret the results, and develop a management plan. These tasks require diverse input modalities and specialized expertise, with intermediate findings guiding subsequent analyses and decisions. Although specialist models excel at individual tasks, they typically operate in isolation, lacking the ability to determine which analyses are needed, coordinate their execution, and integrate results into a coherent medical workflow.

Medical foundation models~\cite{luo2024biomedgpt, nath2025vila, xu2025lingshu, lu2024multimodal} have emerged to address the fragmentation of specialist systems by extending task and modality coverage within unified architectures. Models such as MedSAM~\cite{ma2024segment} and BiomedCLIP~\cite{zhang2023biomedclip} improve generalization across medical imaging modalities and biomedical image-text representations, respectively. Generative vision-language models (VLMs), including LLaVA-Med~\cite{li2023llava} and LLaVA-Rad~\cite{chaves2024towards}, enable open-ended interactions with medical images, while generalist models such as BiomedGPT~\cite{luo2024biomedgpt}, RadFM~\cite{xie2023towards}, and HealthGPT~\cite{lin2025healthgpt} support a broader range of tasks within unified frameworks. Building on this progress, recent models, including VILA-M3~\cite{nath2025vila}, Lingshu~\cite{xu2025lingshu}, Hulu-Med~\cite{jiang2025hulu}, OmniV-Med~\cite{jiang2025omniv}, Fleming-VL~\cite{shu2025fleming}, and UniBiomed~\cite{wu2025unibiomed}, have further expanded task and modality coverage within unified architectures. However, broad task coverage does not inherently provide the ability to coordinate multiple specialized analyses. Complex medical workflows require selecting appropriate tools, executing them in sequence, and adapting subsequent steps based on intermediate results, capabilities that extend beyond supporting diverse tasks within a single model.

Medical agents offer a promising approach to workflow orchestration by dynamically selecting and coordinating specialized capabilities. Existing approaches can be broadly divided into two paradigms. The first emphasizes collaboration among multiple language or vision-language models, where agents assume specialized roles and exchange information through dialogue or structured discussions to reach a shared conclusion~\cite{tang2024medagents,kim2024mdagents,wang2025colacare,zhou2025mam,xia2025mmedagent,mishra2025teammedagents,ma2026medla}. While these systems facilitate collaborative medical reasoning, their coordination primarily occurs through model-to-model interactions rather than the execution of external specialist tools. Unlike model-based collaborative agents, these systems generally do not directly execute specialized medical operations that produce outputs such as segmentation masks or processed medical images.
A second paradigm integrates external specialist tools, enabling medical agents to invoke task-specific models and obtain their outputs directly. Representative systems include MMedAgent~\cite{li2024mmedagent}, MedRAX~\cite{fallahpour2025medrax}, MedAgent-Pro~\cite{wang2025medagent}, and VoxelPrompt~\cite{hoopes2024voxelprompt}. By incorporating executable tools, these systems support workflows that extend beyond text-based reasoning to specialized image analysis and processing.Despite this progress, many tool-augmented medical agents have been developed or evaluated within particular modalities or clinical domains, including general medical imaging~\cite{fathi2025aura,huang2026medsegagent}, histopathology~\cite{lyu2025wsi,chen2025pathagent}, endoscopy~\cite{tang2025endoagent}, chest radiography~\cite{sharma2024cxr,zhang2025radagents,qi2026eviagent,lin2026march,wang2025mrgagents,shen2026evo}, and Alzheimer's disease~\cite{hou2025adagent}. Recent studies have also investigated tool selection and specialized workflow execution, including ToolSelect~\cite{saha2026picking} and MedScope~\cite{li2026medscope}. However, the reliability of tool orchestration across heterogeneous medical tasks, modalities, and multi-step interactions remains insufficiently characterized.

We introduce \textsc{MMedAgent-Plus} (denoted as $\mathcal{M}$), a multimodal medical tool-use agent designed to orchestrate leading specialist models across diverse medical workflows. Building on MMedAgent~\cite{li2024mmedagent}, $\mathcal{M}$ combines a vision-language planner with a library of 18 specialist models covering 12 medical tasks and 8 modalities (Figure~\ref{fig:overview}a; Supplementary Tables~\ref{tab:tool_cards} and~\ref{tab:modality_matrix}). These tasks span language understanding, visual analysis, image processing, and clinical report generation, across text, Computed Tomography (CT), Magnetic Resonance Imaging (MRI), X-ray, ultrasound (US), retinal imaging (including fundus and Optical Coherence Tomography (OCT)), and pathology. Each specialist was selected as the leading publicly available model for its task at the time of construction, based on reported benchmark performance (Supplementary Table~\ref{tab:tool_cards}).

To enable reliable tool orchestration, we construct a large-scale instruction-tuning corpus $\mathcal{S}$ containing 208,928 samples drawn from 32 publicly available medical benchmarks (Supplementary Table~\ref{tab:dataset_details}). The dataset is organized into three dialogue settings that reflect different workflow patterns, including \textit{single-tool multi-round} dialogues for iterative individual analyses, \textit{multi-tool multi-round} dialogues for progressive workflow execution and for successive independent requests across multiple dialogue turns, and \textit{multi-tool single-round} dialogues for compositional workflows within one interaction (Figure~\ref{fig:overview}d). The tool chains covered by each multi-tool setting are listed in Supplementary Tables~\ref{tab:chains_s2} and~\ref{tab:chains_s3}, and one example dialogue per task is shown in Supplementary Figure~\ref{fig:schema_pmc_llama}-\ref{fig:schema_unigradicon}. This training enables $\mathcal{M}$ to select specialist tools, coordinate their execution, and integrate tool-generated evidence into final responses.

The effectiveness of \textsc{MMedAgent-Plus} depends on two complementary components: the planner's ability to select and coordinate appropriate specialist tools, and the performance of the specialists themselves. Across the three dialogue settings, \textsc{MMedAgent-Plus} achieves tool-call accuracies of 91.0\%, 98.1\%, and 80.7\% for single-tool multi-round, multi-tool multi-round, and multi-tool single-round interactions, respectively. Its final responses also outperform those of four recent vision-language models, with improvements in faithfulness (consistency with specialist outputs) and completeness (coverage of relevant information) (Figure~\ref{fig:overview}b).
We further evaluate $\mathcal{M}$ on end-to-end medical workflows requiring the selection and coordination of complementary specialist tools across modalities, followed by the synthesis of tool-grounded responses. Together, these results demonstrate the potential of learned tool orchestration to support multi-step medical workflows across diverse tasks and imaging modalities, extending beyond isolated specialist predictions and direct-answer generation (Figure~\ref{fig:overview}c). Our work establishes a framework for coordinating leading specialists through a learned planner, so that expert-level performance on individual tasks carries over into multi-step clinical workflows.

\begin{figure}[!htbp]
    \centering
    \includegraphics[width=0.97\linewidth]{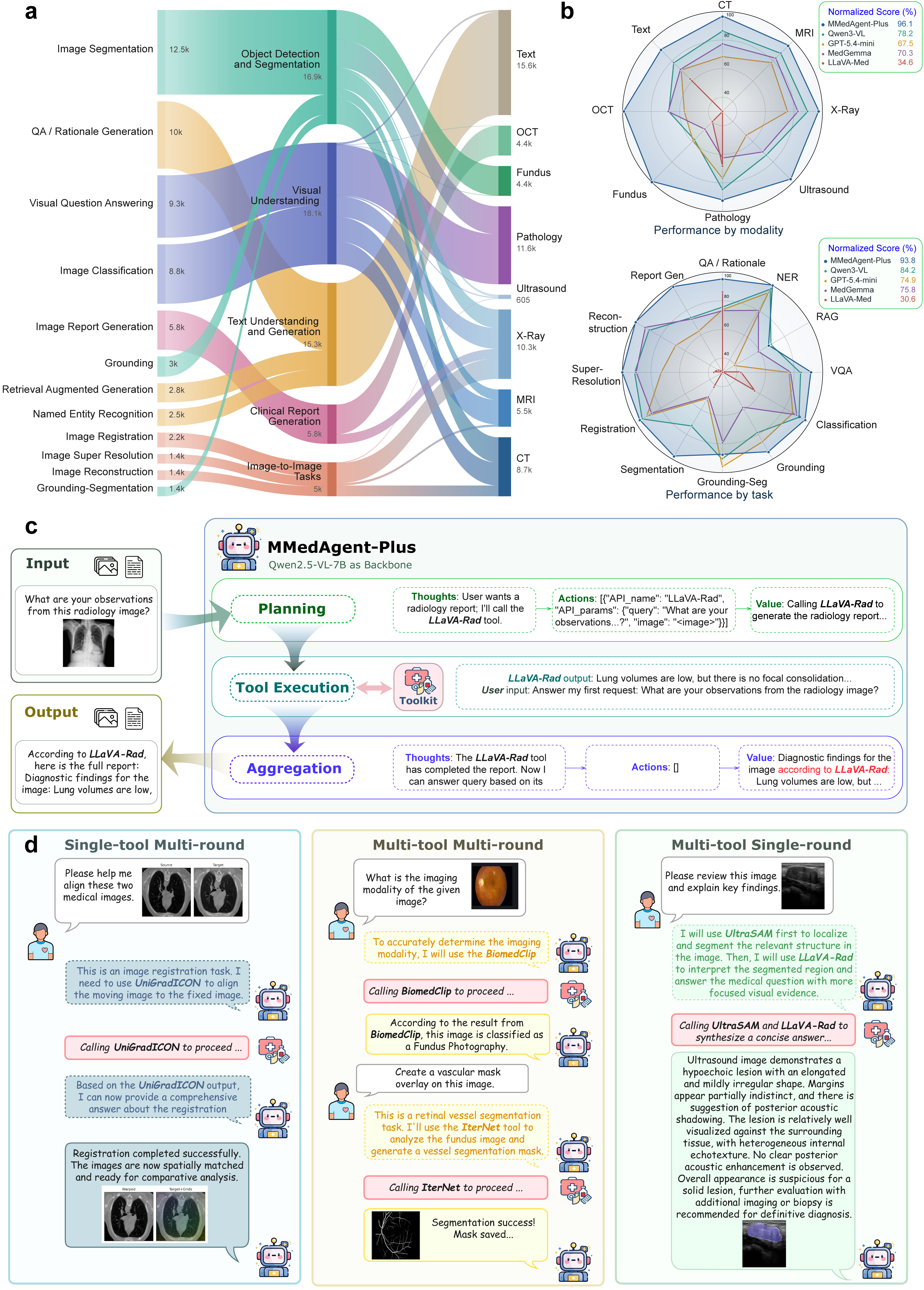}
    \vspace{-2mm}
    \caption{%
    \textbf{Overview of \textsc{MMedAgent-Plus}.}
    (a) Medical tool library across tasks, functions, and modalities.
    (b) Comparison with four vision-language baselines. (c) \textsc{MMedAgent-Plus} work-
    }
    \label{fig:overview}
\end{figure}

\begin{figure}[!tp]
    \ContinuedFloat
    \captionsetup{labelformat=empty}
    \caption[]{%
    flow: planning, tool execution,
    and aggregation.
    (d) Example single- and multi-tool workflows.
    In a single-tool workflow, a request is resolved by one
    specialist call; in a multi-tool workflow, several specialists
    are called in sequence, either within one request (single-round)
    or across successive dialogue turns (multi-round), and earlier
    tool outputs can be passed to later steps.
    }
\end{figure}

%% file: sections/Results.tex
\section{Results}

The performance of \textsc{MMedAgent-Plus} relies on three capabilities: selecting and invoking appropriate specialist tools, executing these tools successfully, and synthesizing their outputs into coherent responses. We assess these capabilities through complementary component-level and end-to-end evaluations. At the component level, we evaluate tool orchestration accuracy on 1,500 held-out dialogues(Section~\ref{sec:orchestration}), and response synthesis quality on 600 cases with correct specialist outputs provided (Section~\ref{sec:final_answer_quality}). We then assess end-to-end performance on 350 single-tool cases, requiring the model to independently select and invoke the appropriate specialist and integrate its output into the final response (Section~\ref{sec:end-to-end}). This evaluation design allows us to distinguish performance at individual stages from that of the integrated system, and it also tests whether component-level results are maintained when these stages are combined end to end. Evaluation metrics are detailed in Section~\ref{sec:evaluation}.

\paragraph{Evaluation data.}
\label{sec:eval_data}

All evaluation cases are drawn from held-out partitions established before training. Approximately 10\% of dialogues from each setting were reserved for evaluation, with no overlap in dialogues or source images between training and evaluation (Supplementary Section~\ref{sup:split_protocol}). This yielded held-out pools of 9{,}128 single-tool multi-round, 10{,}000 multi-tool multi-round, and 2{,}430 multi-tool single-round dialogues.

Using a fixed random seed, we sampled three evaluation sets: (i) 500 dialogues per setting for tool orchestration (Section~\ref{sec:orchestration}); (ii) 600 disjoint response-synthesis cases, comprising 200 per setting, each extracted as a single tool-return and response round from a held-out dialogue (Section~\ref{sec:final_answer_quality}); and (iii) 350 end-to-end cases, comprising 50 cases for each of seven specialist tools, drawn from the single-tool pool (Section~\ref{sec:end-to-end}).

\begin{figure}[!h]
    \centering
    \includegraphics[width=\linewidth]{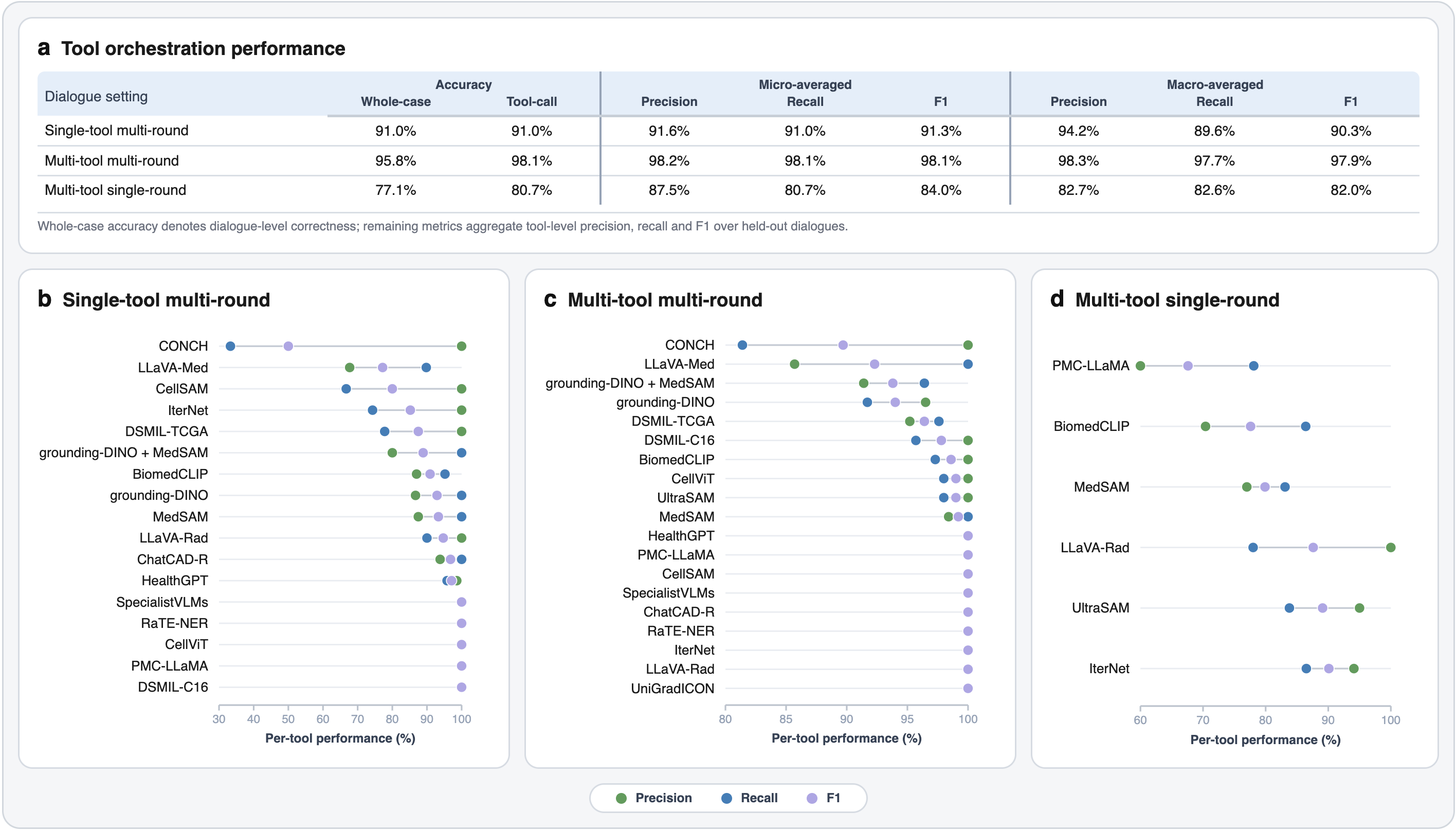}
    \caption{\textbf{Tool orchestration performance across dialogue settings.}
    \textbf{a,} Aggregate performance across single-tool multi-round, multi-tool multi-round, and multi-tool single-round dialogues, evaluated on 500 held-out dialogues per setting (Section~\ref{sec:eval_data}). Whole-case accuracy (Eq.~\ref{eq:case_acc}) requires the complete ordered sequence of tool calls within a dialogue to match the reference sequence, whereas tool-call accuracy (Eq.~\ref{eq:tool_acc}) measures correctness at the level of individual required tool invocations. Tool-level precision, recall, and F1 are summarized using micro-averaging, which pools tool decisions across all dialogues, and macro-averaging, which assigns equal weight to each tool type.    
    \textbf{b-d,} Per-tool precision, recall, and F1 for \textbf{b,} single-tool multi-round, \textbf{c,} multi-tool multi-round, and \textbf{d,} multi-tool single-round dialogues. Each row represents a specialist tool, with connected markers indicating its precision, recall, and F1. Tools are ordered by F1 within each panel. Precision captures the extent of unnecessary tool selection, whereas recall captures missed required tool invocations. Full metric definitions are provided in Section~\ref{sec:evaluation}.}    
    \label{fig:tool_orchestration}
\end{figure}

The evaluation sets are smaller than the training corpus, which spans 18 specialist tools, 8 modalities, and thousands of tool-chain combinations (Supplementary Tables~\ref{tab:chains_s2} and~\ref{tab:chains_s3}). Nevertheless, the 95\% Wilson confidence intervals for tool-orchestration accuracy have margins of error within 4 percentage points across all three settings (Figure~\ref{fig:tool_orchestration}a). Representative dialogues and qualitative end-to-end comparisons are provided in Supplementary Figure~\ref{fig:schema_pmc_llama}-\ref{fig:schema_unigradicon} and~\ref{fig:qual_qa}-\ref{fig:qual_path_3}.

\begin{figure}[!h]
    \centering
    \includegraphics[width=\linewidth]{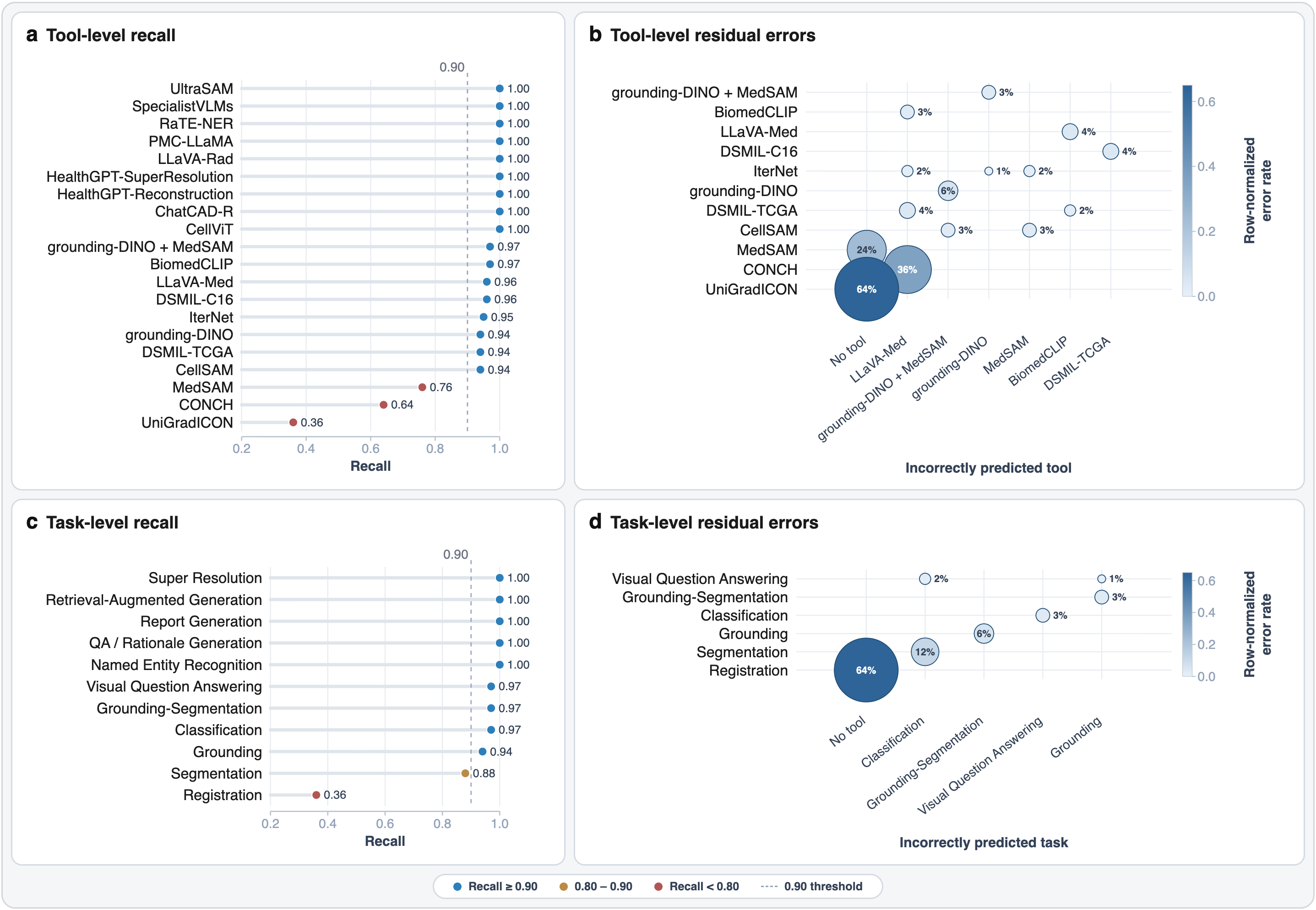}
    \caption{\textbf{Tool- and task-level routing performance across dialogue settings.}
    Row-normalized recall and incorrect routing assignments aggregated across three dialogue settings (1{,}500 held-out dialogues).
    \textbf{a,} Per-tool recall, with the dashed line indicating 0.90.
    \textbf{b,} Nonzero incorrect tool assignments, showing how missed or misrouted tool calls are distributed across alternative routing outcomes. \texttt{No tool} indicates that $\mathcal{M}$ responds without invoking a specialist.
    \textbf{c,} Per-task recall.
    \textbf{d,} Nonzero incorrect task assignments.
    For Registration, the proportion assigned to \texttt{No tool} equals $1-$recall, corresponding to the missed UniGradICON invocations in \textbf{b}.}
    \label{fig:routing}
\end{figure}

\subsection{\textsc{MMedAgent-Plus} demonstrates effective tool orchestration across diverse dialogue settings}
\label{sec:orchestration}

We evaluate $\mathcal{M}$ on 500 held-out dialogues for each of the three dialogue types (1{,}500 in total; Section~\ref{sec:eval_data}), with full results shown in Figure~\ref{fig:tool_orchestration}a. Each \textit{dialogue} represents a complete interaction, potentially spanning multiple \textit{round} of user requests and system responses, with tool calls occurring within or across rounds. We assess orchestration performance using two complementary metrics: \textit{Tool-call accuracy} (Eq.~\ref{eq:tool_acc}), which measures the correctness of individual tool selections, and \textit{Whole-case accuracy} (Eq.~\ref{eq:case_acc}), which requires all tool calls within a dialogue to be correct. We additionally report micro- and macro-averaged precision, recall, and F1~\cite{sokolova2009systematic} to characterize tool-level performance. Figure~\ref{fig:tool_orchestration}b-d presents aggregate and per-tool results across the three dialogue settings, while Figure~\ref{fig:routing} summarizes tool- and task-level recall and incorrect routing assignments.

\paragraph{Single-tool multi-round.}
In the single-tool multi-round setting, $\mathcal{M}$ achieves 91.0\% whole-case and tool-call accuracy (95\% CI, 88.2-93.2\%), indicating consistent tool selection across successive interactions involving individual specialists. Figure~\ref{fig:tool_orchestration}b shows that achieve precision, recall, and F1 scores above 99.0\%, including PMC-LLaMA~\cite{wu2024pmc} (text QA), RaTE-NER~\cite{zhao2024ratescore} (named entity recognition (NER)), SpecialistVLMs~\cite{holland2024specialized} (retinal OCT report generation), CellViT~\cite{horst2023cellvit} (WSI cell segmentation), and DSMIL-C16~\cite{li2021dual} (breast tumour classification). Other tools, including HealthGPT~\cite{lin2025healthgpt} (reconstruction and super-resolution), LLaVA-Rad~\cite{chaves2024towards} (radiology report generation), and ChatCAD-R~\cite{wang2024interactive} (retrieval-augmented generation (RAG)) also exceed 90.0\% across all three metrics, indicating strong routing performance across diverse medical tasks and modalities.

\paragraph{Multi-tool multi-round.}
The multi-tool multi-round setting achieves the highest overall performance, with 98.1\% tool-call accuracy (95\% CI, 97.2-98.7\%; 1{,}251 tool calls) and 95.8\% whole-case accuracy (95\% CI, 93.7-97.2\%) (Figure~\ref{fig:tool_orchestration}a). Micro-averaged precision, recall, and F1 are 98.2\%, 98.1\%, and 98.1\%, respectively, matching the macro-averaged scores of 98.3\%, 97.7\%, and 97.9\% (Figure~\ref{fig:tool_orchestration}c). This agreement suggests that performance is distributed across specialist tools rather than dominated by frequently selected tools. At the individual-tool level, nine specialists achieve precision, recall, and F1 scores above 99.0\%, including HealthGPT, PMC-LLaMA, LLaVA-Rad, CellSAM~\cite{marks2025cellsam} (2D cell segmentation), IterNet~\cite{li2020iternet} (fundus segmentation), and UniGradICON~\cite{tian2024unigradicon} (image registration), while five additional tools exceed 90.0\% across all three metrics.

\paragraph{Multi-tool single-round.}
The multi-tool single-round setting yields the lowest orchestration accuracy among the three settings, reflecting the challenge of coordinating multiple specialist tools within a single interaction. $\mathcal{M}$ achieves 80.7\% tool-call accuracy (95\% CI 77.0-83.9\%) and 77.1\% whole-case accuracy (73.2-80.6\%). Micro-averaged precision, recall, and F1 are 87.5\%, 80.7\%, and 84.0\%, respectively, compared with macro-averaged scores of 82.7\%, 82.6\%, and 82.0\% (Figure~\ref{fig:tool_orchestration}d). Despite the lower aggregate performance, specialists such as UltraSAM~\cite{meyer2024ultrasam}, IterNet maintain high precision and recall. The difference between tool-call and whole-case accuracy suggests that some dialogues contain a mixture of correct and incorrect tool selections, highlighting the challenge of accurately composing multiple specialist operations.

\paragraph{Aggregate error structure.}

Across the three dialogue settings, routing errors are concentrated in a small subset of specialist tools and tasks (Figure~\ref{fig:routing}). All but three tools achieve recall above 0.90, with nine reaching 1.00. The lowest recall is observed for UniGradICON (0.36), CONCH (0.64), and MedSAM (0.76) (Figure~\ref{fig:routing}a). Two dominant error patterns emerge at the tool level (Figure~\ref{fig:routing}b). First, the model occasionally confuses tools with related capabilities, most notably routing 36\% of CONCH cases to LLaVA-Med (VQA), potentially reflecting overlap between pathology classification and medical visual question answering. Second, the model sometimes bypasses required specialist tools and responds directly, as observed in 64\% of UniGradICON cases and 24\% of MedSAM cases routed to \texttt{No tool}. Similar patterns emerge at the task level, where most tasks achieve recall of at least 0.94, except Registration (0.36) and Segmentation (0.88) (Figure~\ref{fig:routing}c). Registration errors consist almost entirely of omitted tool invocations, consistent with the UniGradICON results. Other errors include Segmentation routed to Classification (12\%), Visual Question Answering routed to Classification (2\%), and confusion between Grounding and Grounding-Segmentation (Figure~\ref{fig:routing}d).

\subsection{\textsc{MMedAgent-Plus} achieves high-quality tool-grounded responses}
\label{sec:final_answer_quality}

\begin{table}[ht]
\centering
\caption{\textbf{Final response quality across dialogue settings.} Responses are evaluated for Faithfulness (Faith.) / Completeness (Compl.) / Conciseness (Conci.) on a 1-5 scale. Avg denotes the mean of the three scores. Dialogue settings includes single-tool multi-round (SM), multi-tool multi-round (MM), and multi-tool single-round (MS).}
\label{tab:final_answer_quality}
\footnotesize
\setlength{\tabcolsep}{2pt}
\begin{tabular}{l cccc cccc cccc}
\toprule
\textbf{Model}
& \multicolumn{4}{c}{\textbf{SM \textnormal{(Single-Tool MR)}}}
& \multicolumn{4}{c}{\textbf{MM \textnormal{(Multi-Tool MR)}}}
& \multicolumn{4}{c}{\textbf{MS \textnormal{(Multi-Tool SR)}}} \\
\cmidrule(lr){2-5} \cmidrule(lr){6-9} \cmidrule(lr){10-13}
& Avg & Faith. & Compl. & Conci.
& Avg & Faith. & Compl. & Conci.
& Avg & Faith. & Compl. & Conci. \\
\midrule
LLaVA-Med~\cite{li2023llava}
& 2.68 & 3.14 & 2.15 & 2.76
& 1.27 & 1.03 & 1.01 & 1.77
& 2.45 & 2.03 & 2.15 & 3.18 \\

GPT-5.4-mini~\cite{openai2026gpt54mini}
& 3.80 & 3.46 & 3.74 & 4.21
& 4.04 & 3.72 & 3.88 & 4.54
& 2.53 & 2.11 & 2.40 & 3.08 \\

MedGemma~\cite{sellergren2026medgemma15technicalreport}
& 3.95 & 3.76 & 3.59 & 4.49
& 3.87 & 3.59 & 3.52 & 4.50
& 3.12 & 2.98 & 2.75 & 3.63 \\

Qwen3-VL~\cite{bai2025qwen3vltechnicalreport}
& 4.42 & 4.16 & 4.43 & 4.68
& 4.29 & 3.86 & 4.34 & 4.65
& 3.63 & 3.30 & 3.49 & 4.10 \\
\midrule
\textsc{MMedAgent-Plus} (ours)
& \textbf{4.84} & \textbf{4.85} & \textbf{4.90} & \textbf{4.78}
& \textbf{4.79} & \textbf{4.68} & \textbf{4.87} & \textbf{4.81}
& \textbf{4.92} & \textbf{5.00} & \textbf{5.00} & \textbf{4.77} \\
\bottomrule
\end{tabular}
\end{table}

We evaluate the quality of final responses generated by $\mathcal{M}$ on 600 held-out cases (200 per dialogue setting), each comprising a single tool-return and response round extracted from a held-out dialogue (Section~\ref{sec:eval_data}). These cases are disjoint from those used for tool orchestration evaluation (Section~\ref{sec:orchestration}). To isolate response synthesis from tool-selection errors, we provide $\mathcal{M}$ with the correct specialist outputs and assess its ability to integrate them into coherent, tool-grounded responses. We adopt an LLM-as-a-judge protocol evaluating three criteria: faithfulness, completeness, and conciseness, each scored from 1 to 5, with a combined total of 3 to 15, defined in Section~\ref{sec:evaluation}). For $\mathcal{M}$, faithfulness measures consistency with the provided specialist outputs rather than clinical correctness relative to the original images. For direct-answer baselines, which do not receive specialist outputs, faithfulness is assessed against the original multimodal inputs (Section~\ref{sec:baseline}). The results across the three dialogue settings are summarized in Table~\ref{tab:final_answer_quality}, with modality- and task-level analyses presented in Figure~\ref{fig:overview}b.

\textsc{MMedAgent-Plus} achieves the highest overall response quality scores across all three dialogue settings (Table~\ref{tab:final_answer_quality}). Its largest advantage is observed in the multi-tool single-round setting, where it scores 14.77 out of 15, compared with 10.89 for the strongest direct-answer baseline, Qwen3-VL-8B-Instruct~\cite{bai2025qwen3vltechnicalreport}. The corresponding scores are 14.52 versus 13.27 in single-tool multi-round dialogues and 14.36 versus 12.86 in multi-tool multi-round dialogues. Improvements are most pronounced in faithfulness and completeness, with \textsc{MMedAgent-Plus} achieving perfect scores of 5.00 on both criteria in the multi-tool single-round setting. Conciseness scores are also higher across all three settings, reaching 4.77 compared with 4.10 for Qwen3-VL-8B-Instruct in the multi-tool single-round setting.

Modality-level analysis shows that \textsc{MMedAgent-Plus} consistently outperforms direct-answer baselines across the evaluated imaging modalities, with particularly large differences in retinal imaging (Figure~\ref{fig:overview}b). For fundus images, all baselines score below 50\%, whereas \textsc{MMedAgent-Plus} approaches 100\%. A similarly large margin is observed for retinal OCT, while higher scores are also maintained across CT, MRI, X-ray, and ultrasound.

Task-level analysis shows that \textsc{MMedAgent-Plus} achieves higher response quality scores than direct-answer baselines across most evaluated tasks (Figure~\ref{fig:overview}b), with particularly strong performance in report generation, RAG, and NER. An important distinction arises in grounding-segmentation tasks, where direct-answer baselines provide only textual descriptions without generating segmentation outputs. Consequently, response quality scores capture only the textual component of these tasks and do not fully reflect the differences in task completion between tool-grounded and direct-answer approaches.

Together, these results demonstrate that the strengths of $\mathcal{M}$ extend beyond tool selection to the synthesis of specialist-generated evidence. Given correct tool outputs, $\mathcal{M}$ produces more faithful and complete responses than direct-answer baselines under the evaluation protocol, with the largest margin observed in the multi-tool single-round setting. As this evaluation isolates response synthesis from tool selection and execution, it does not establish whether these advantages translate into end-to-end task performance. We investigate this question in Section~\ref{sec:end-to-end}, where $\mathcal{M}$ must independently select and execute specialist tools before generating its final responses.

\subsection{\textsc{MMedAgent-Plus} preserves specialist-level performance}
\label{sec:end-to-end}

We evaluate the end-to-end performance of \textsc{MMedAgent-Plus} without pre-supplied specialist outputs, requiring the system to independently select and execute tools before generating final responses. The evaluation covers seven representative tasks, each associated with one of the 18 specialist tools, using 50 held-out cases per task (350 in total) drawn from the single-tool multi-round test set (Section~\ref{sec:tools}, Table~\ref{tab:e2e_metrics}). These include QA~/~rationale generation (PMC-LLaMA~\cite{wu2024pmc}), named entity recognition (RaTE-NER~\cite{zhao2024ratescore}), report generation (LLaVA-Rad~\cite{chaves2024towards}), pathology classification (CONCH~\cite{lu2024visual}), multimodal classification (BiomedCLIP~\cite{zhang2023biomedclip}), fundus segmentation (IterNet~\cite{li2020iternet}), and general medical image segmentation (MedSAM~\cite{ma2024segment}). We assess required-tool invocation accuracy and task-specific performance across all seven tasks, with additional LLM-as-a-judge evaluation for the five text-producing tasks.

\begin{figure}[!h]
    \centering
    \includegraphics[width=\linewidth]{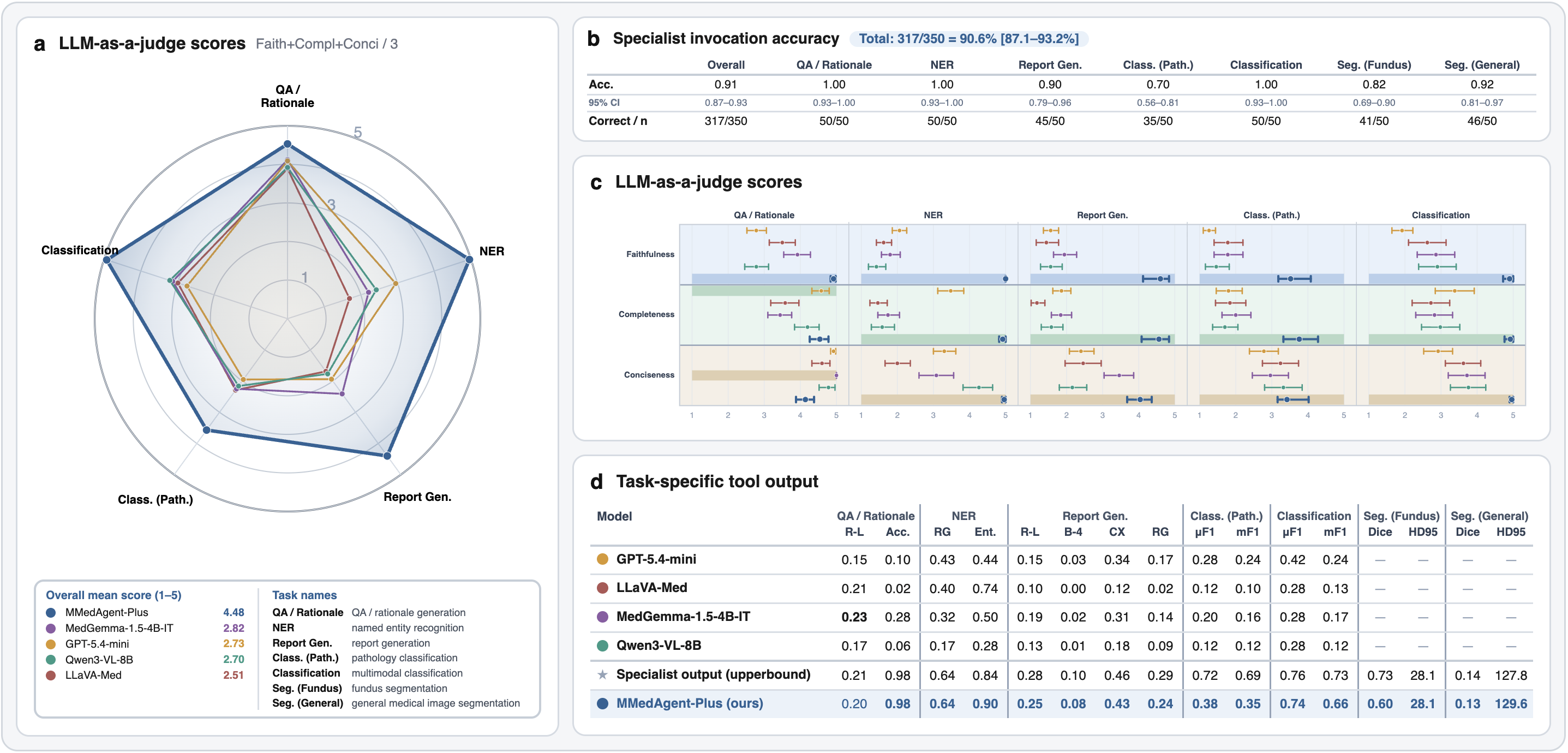}
    \caption{\textbf{End-to-end performance of \textsc{MMedAgent-Plus} across seven representative tasks.} Each task is evaluated on 50 held-out cases using one specialist tool (Table~\ref{tab:e2e_metrics}), giving 350 cases in total.
    \textbf{a,} LLM-as-a-judge scores for five text-producing tasks, showing the mean of faithfulness, completeness, and conciseness (1-5 scale) for each task. The legend reports the overall mean across the five tasks.
    \textbf{b,} Specialist invocation accuracy, reported as the fraction of cases in which $\mathcal{M}$ selects the required specialist, together with 95\% Wilson confidence intervals and the underlying counts.
    \textbf{c,} Task-level scores from \textbf{a} decomposed into faithfulness, completeness, and conciseness. Points indicate means, and error bars represent 95\% bootstrap confidence intervals. Shaded bands highlight the highest mean within each task. Direct-answer baselines are evaluated against the original multimodal inputs because they do not receive specialist outputs.
    \textbf{d,} Task-specific output quality compared with standalone specialists and direct-answer baselines. The \textit{Specialist output} row reports metrics computed directly from specialist outputs, whereas model rows report metrics computed from final responses. Abbreviations: R-L, ROUGE-L; Acc., accuracy; RG, RadGraph F1; Ent., entity-level F1; B-4, BLEU-4; CX, CheXpert F1; $\mu$F1, micro-F1; mF1, macro-F1; HD95, 95th-percentile Hausdorff distance. For segmentation, Dice is averaged over all 50 cases (missing masks score 0), whereas HD95 is averaged only over cases with generated masks (lower is better, 0 indicates a perfect match). Segmentation metrics are not applicable to text-only responses and are indicated by dashes.}
    \label{fig:e2e}
\end{figure}

\subsubsection{Specialist Invocation Accuracy}

Across 350 held-out end-to-end cases, $\mathcal{M}$ correctly invokes the required specialist in 317 cases, achieving 90.6\% invocation accuracy (95\% Wilson CI, 87.1-93.2\%; Figure~\ref{fig:e2e}b).

Perfect invocation accuracy (50/50 cases) is observed for QA~/~rationale generation, , named entity recognition, and multimodal classification. General medical image segmentation reaches 92.0\% (81.2-96.9\%), report generation reaches 90.0\% (78.6-95.7\%), and fundus segmentation reaches 82.0\% (69.2-90.2\%). Pathology classification shows the lowest accuracy at 70.0\% (56.2-80.9\%), consistent with the tool-routing difficulties observed for CONCH in Section~\ref{sec:orchestration}.

\subsubsection{Response Quality under LLM-as-a-judge}

We assess end-to-end response quality using the same three-criterion LLM-as-a-judge protocol described in Section~\ref{sec:final_answer_quality}. \textsc{MMedAgent-Plus} is judged against the specialist outputs it receives, whereas the direct-answer baselines are judged against the original multimodal input. Figure~\ref{fig:e2e}d provides a shared task-specific reference. Figure~\ref{fig:e2e}c reports means and 95\% confidence intervals for faithfulness, completeness, and conciseness on all five baseline-comparable tasks (QA~/~rationale generation, named entity recognition, report generation, pathology classification, and multimodal classification). Figure~\ref{fig:e2e}a summarizes the same five tasks as per-task means of the three criteria.

\textsc{MMedAgent-Plus} achieves the highest mean response quality score across all five evaluated tasks. The largest margins over direct-answer baselines are observed in radiology report generation (4.40 versus 2.41 for MedGemma-1.5-4B-IT), named entity recognition (4.96 versus 2.95), and multimodal classification (4.93 versus 3.21 for Qwen3-VL-8B). Smaller but consistent differences are observed in QA~/~rationale generation (4.53 versus 4.12 for MedGemma-1.5-4B-IT) and pathology classification (3.57 versus 2.29). For radiology report generation, \textsc{MMedAgent-Plus} achieves scores of 4.60 and 4.56, respectively, whereas all direct-answer baselines score below 2.0 on both criteria. Conciseness is generally comparable or higher, except in QA~/~rationale generation, where MedGemma-1.5-4B-IT and GPT-5.4-mini score 5.00 and 4.92, respectively, compared with 4.14 for \textsc{MMedAgent-Plus}.

\subsubsection{Task-Specific Tool Output Quality}

Figure~\ref{fig:e2e}d compares task-specific performance measured directly from specialist outputs with that obtained from the final responses of \textsc{MMedAgent-Plus} and direct-answer VLM baselines. This comparison assesses how well specialist capabilities are preserved during end-to-end execution.

For QA~/~rationale generation, \textsc{MMedAgent-Plus} matches the specialist accuracy of 0.98, compared with 0.28 for the strongest direct-answer baseline. In named entity recognition, it preserves the specialist RadGraph F1 of 0.64 and achieves a higher entity-level F1 (0.90 versus 0.84). For report generation and multimodal classification, \textsc{MMedAgent-Plus} also achieves scores closest to the specialist outputs and outperforms direct-answer baselines across the reported metrics. The exception is ROUGE-L for QA~/~rationale generation, where MedGemma-1.5-4B-IT scores 0.23 compared with 0.20 for \textsc{MMedAgent-Plus}.

Pathology classification exhibits a larger performance gap relative to the specialist, with micro-F1 decreasing from 0.72 to 0.38 across all 50 cases. Among the 35 cases with correct specialist invocation, the corresponding scores are 0.69 and 0.51, suggesting that both routing failures and subsequent response synthesis contribute to the overall performance loss. For the two segmentation tasks, Dice and HD95 are computed directly from masks generated during end-to-end execution. These metrics are unavailable for direct-answer baselines, which produce only textual responses rather than segmentation masks.

%% file: sections/Discussion.tex
\section{Discussion}

\subsection{Tool orchestration extends medical AI beyond isolated specialist models}

\textsc{MMedAgent-Plus} demonstrates the potential of integrating heterogeneous specialist models through a unified vision-language planner. By coordinating 18 specialist tools spanning 12 medical tasks and eight modalities, the system extends beyond isolated predictions toward flexible tool-based medical workflows. Instruction tuning across three dialogue settings enables the planner to select appropriate specialists and integrate their outputs into coherent responses. The results show high tool-call accuracy across settings, reaching 98.1\% tool-call accuracy in multi-tool multi-round setting, 91.0\% in single-tool multi-round and 80.7\% in multi-tool single-round (Figure~\ref{fig:tool_orchestration}a). Its response synthesis consistently achieves higher faithfulness and completeness scores than direct-answer VLM baselines. In the multi-tool single-round setting, for example, $\mathcal{M}$ achieves a total final response score of 14.77, compared with 10.89 for the strongest baseline (Table~\ref{tab:final_answer_quality}). End-to-end evaluations further demonstrate that the system can autonomously invoke specialist tools and retain task-specific performance in several medical tasks (Figure~\ref{fig:e2e}c)). Together, these results show that $\mathcal{M}$ can orchestrate heterogeneous specialist tools and preserve their outputs in the final response, combining specialized medical capabilities within a unified agent framework.

The value of this approach is particularly evident in medical workflows involving multiple interdependent analyses. \textsc{MMedAgent-Plus} is designed to coordinate specialist tools sequentially, allowing intermediate outputs to inform subsequent operations. This capability distinguishes tool-based orchestration from two common approaches. Generalist VLMs~\cite{luo2024biomedgpt, nath2025vila, xu2025lingshu, lu2024multimodal} support diverse medical tasks within unified architectures but typically generate responses directly from the input without explicitly executing specialized operations.
Role-based medical agents, including MMedAgent-RL~\cite{xia2025mmedagent}, TeamMedAgents~\cite{mishra2025teammedagents}, and MedLA~\cite{ma2026medla}, primarily coordinate language-model agents through collaborative reasoning rather than the execution and transfer of non-text medical artifacts. In contrast, \textsc{MMedAgent-Plus} explicitly models tool-call sequences, achieving whole-case accuracies of 95.8\% in multi-tool multi-round dialogues and 77.1\% in multi-tool single-round dialogues (Figure~\ref{fig:tool_orchestration}a).

\textsc{MMedAgent-Plus} also extends the scope of existing tool-augmented medical agents through a broader specialist library. As summarized in Table~\ref{tab:agent_scope}, many previous systems focus on specific imaging modalities or disease domains~\cite{lyu2025wsi, chen2025pathagent, zhang2025radagents, qi2026eviagent, lin2026march, wang2025mrgagents, shen2026evo, hou2025adagent}. MMedAgent~\cite{li2024mmedagent}, on which our system builds, supports five modalities and six tasks, whereas $\mathcal{M}$ covers eight modalities and twelve tasks, including training examples involving multiple specialist tools. This broader coverage is accompanied by high response quality across diverse modalities, including fundus imaging and retinal OCT, where the largest margins over direct-answer baselines are observed (Figure~\ref{fig:overview}b). Averaged across eight modality axes, $\mathcal{M}$ achieves 96.1\%, compared with 78.2\% for the strongest baseline (Qwen3-VL-8B). These findings suggest that heterogeneous specialist integration can support broad medical task coverage while maintaining high tool-grounded response quality.

\begin{table}[ht]
\centering
\caption{\textbf{Task and modality coverage of medical tool-orchestrating agents.} Comparison of the reported modality or clinical scope and supported tasks of representative systems. MMedAgent is the base system extended by \textsc{MMedAgent-Plus}.}
\label{tab:agent_scope}
\footnotesize
\begin{tabular}{l c c}
\toprule
\textbf{System} & \textbf{Modality / clinical scope} & \textbf{Supported tasks} \\
\midrule
WSI-Agents~\cite{lyu2025wsi} & Pathology (WSI) & Multi-task WSI analysis \\
PathAgent~\cite{chen2025pathagent} & Pathology (WSI) & WSI analysis \\
RadAgents~\cite{zhang2025radagents} & Chest X-ray & Reporting, interpretation \\
EviAgent~\cite{qi2026eviagent} & Chest X-ray & Report generation \\
MARCH~\cite{lin2026march} & Chest CT & Report generation \\
MRGAgents~\cite{wang2025mrgagents} & Chest X-ray & Report generation \\
Evo-MedAgent~\cite{shen2026evo} & Chest X-ray & Classification, VQA \\
ADAgent~\cite{hou2025adagent} & Alzheimer's disease (MRI, PET) & Diagnosis, prognosis \\
\midrule
MMedAgent~\cite{li2024mmedagent} (base) & 5 modalities & 6 tasks \\
\textbf{\textsc{MMedAgent-Plus} (ours)} & \textbf{8 modalities} & \textbf{12 tasks} \\
\bottomrule
\end{tabular}
\end{table}

\subsection{Orchestration errors reflect tool ambiguity and implicit workflow requirements}

Tool orchestration performance varies substantially across the three dialogue settings (Figure~\ref{fig:tool_orchestration}). Tool-call accuracy is highest in multi-tool multi-round dialogues (98.1\%), followed by single-tool multi-round (91.0\%) and multi-tool single-round dialogues (80.7\%), with whole-case accuracy following the same trend. These differences suggest that orchestration difficulty depends not only on the number of required tools but also on when information becomes available to the planner. In multi-round interactions, preceding requests and intermediate tool outputs can inform subsequent tool selections. By contrast, multi-tool single-round requests require the planner to determine the complete tool sequence upfront, before any specialist is executed or intermediate results become available.

The multi-tool multi-round setting achieves the highest performance, with 98.1\% tool-call accuracy and 95.8\% whole-case accuracy. In this setting, the planner selects tools incrementally rather than committing to the complete sequence from the initial request. Follow-up turns can therefore be interpreted together with previously executed tools and their returned intermediate outputs. Such contextual information may help narrow the choice of specialist tools by clarifying the imaging modality and current workflow stage. The closely aligned micro- and macro-averaged F1 scores (98.1\% and 97.9\%, respectively) further suggest that strong performance extends across the tool library rather than being dominated by frequently selected tools. Together, these findings highlight the potential value of intermediate context for sequential tool orchestration.

Single-tool multi-round dialogues achieve 91.0\% tool-call and whole-case accuracy, with identical values because each dialogue contains only one tool invocation. Unlike multi-tool multi-round interactions, these dialogues provide no preceding specialist outputs to inform tool selection, requiring the planner to identify the appropriate tool primarily from the current request and input. This can introduce ambiguity when specialists accept similar input modalities or produce similar outputs. Per-tool F1 scores are lower for CONCH (0.50), LLaVA-Med (0.78), and CellSAM (0.80), whereas several other tools approach perfect performance (Figure~\ref{fig:tool_orchestration}b). For example, CONCH performs pathology classification, while LLaVA-Med supports general medical visual question answering. Both of them can process pathology images and generate textual outputs, potentially making their intended use difficult to distinguish from certain requests. Such ambiguity may contribute to the gap between micro- and macro-averaged F1, suggesting that orchestration errors are concentrated in a subset of tools with overlapping capabilities.

Multi-tool single-round dialogues show the lowest orchestration performance, with 80.7\% tool-call accuracy and 77.1\% whole-case accuracy. Micro-averaged recall equals tool-call accuracy (80.7\%) and is lower than micro-averaged precision (87.5\%), suggesting that missed tool invocations are more prevalent than unnecessary selections. This imbalance is less apparent under macro-averaging, where precision (82.7\%) and recall (82.6\%) are nearly identical, reflecting differences in tool frequency and error distribution. LLaVA-Rad illustrates the omission pattern, achieving 100\% precision but only 78.0\% recall across 350 required calls, with missed invocations routed to \texttt{no tool}. In these cases, report generation is an implicit intermediate step that the planner must infer from the request. Failure to recognize this dependency leads to incomplete tool sequences, contributing to the lower whole-case accuracy. These findings highlight the difficulty of planning multi-tool workflows when intermediate operations are not explicitly specified.

Taken together, these findings suggest two recurring challenges in tool orchestration (Figure~\ref{fig:routing}). First, tools with overlapping input modalities and output formats can be difficult to distinguish when the user's intended task is ambiguous, as illustrated by confusion between CONCH and LLaVA-Med. Second, required intermediate operations may be omitted when they are not explicitly specified in the request, leading to incomplete tool sequences. The higher accuracy observed in multi-tool multi-round dialogues suggests that preceding tool calls and intermediate outputs may help resolve such ambiguity by clarifying the current workflow state. Conversely, tools with more distinctive task and modality requirements, including ChatCAD-R, PMC-LLaMA, RaTE-NER, SpecialistVLMs, and CellViT, achieve recall above 99\%, with similarly high recall observed across CT, MRI, ultrasound, retinal imaging, and X-ray tasks.

\subsection{Residual errors arise from specialist selection and response synthesis}

End-to-end evaluation shows that $\mathcal{M}$ successfully invokes the required specialist in 90.6\% of cases and achieves higher LLM-as-a-judge scores than direct-answer baselines across all five evaluated tasks (Figure~\ref{fig:e2e}a-c). The largest margins are observed in tasks where specialist tools provide task-specific evidence, including radiology report generation and multimodal classification, whereas the difference is smaller for open-ended question answering. Task-specific metrics further indicate that specialist performance is largely retained during response synthesis for several tasks, with $\mathcal{M}$ matching the specialist accuracy of 0.98 in QA~/~rationale generation and achieving comparable performance in named entity recognition, report generation, and multimodal classification (Figure~\ref{fig:e2e}d). These findings suggest that the benefits of tool-grounded response generation can extend to autonomous end-to-end execution, although performance remains sensitive to errors in specialist selection and subsequent output integration.

End-to-end performance losses can arise at two stages: specialist selection and post-invocation response synthesis. Pathology classification illustrates both challenges. Its invocation accuracy is the lowest among the evaluated tasks (70.0\%), consistent with the confusion between CONCH and LLaVA-Med observed in the tool orchestration analysis. Even when the correct specialist is invoked, converting its outputs into final responses can introduce further information loss. Across all 50 pathology classification cases, micro-F1 decreases from 0.72 for the specialist to 0.38 for $\mathcal{M}$. Among the 35 cases with correct specialist invocation, the corresponding scores are 0.69 and 0.51, suggesting that both routing failures and post-invocation losses contribute to the overall performance gap. A related trade-off appears in QA~/~rationale generation, where $\mathcal{M}$ achieves higher response quality but lower conciseness scores than MedGemma and GPT-5.4-mini. For segmentation tasks, performance is assessed from generated masks using Dice and HD95, rather than from natural-language responses. Together, these findings highlight the importance of both accurate tool selection and faithful preservation of specialist outputs during end-to-end execution.

\subsection{Extending tool orchestration to adaptive workflows and clinical practice}

Our framework has several limitations. First, end-to-end performance depends on the quality of the underlying specialist tools, whose errors may propagate to final responses or subsequent workflow steps. This limitation is evident in general medical image segmentation, where output quality remains limited despite correct tool invocation (Figure~\ref{fig:e2e}d). Second, the evaluated workflows are less complex than those encountered in clinical practice. Our multi-tool settings involve chains of two to three tools constructed from public medical benchmarks rather than prospectively collected clinical workflows. Our evaluation does not cover substantially longer tool sequences, workflows in which the choice of a later tool depends on the specific output of an earlier step, or cases in which several valid tool sequences may address the same request. The reliability observed in the current experiments should therefore not be assumed to extend directly to these more complex settings. Finally, extending the tool library to new tasks or modalities requires additional instruction tuning. Although compatible specialists can replace existing tools without changing their interfaces, introducing new capabilities requires the planner to learn when and how to invoke them. Prior work on MMedAgent~\cite{li2024mmedagent} suggests that incremental training can incorporate new tools without substantially degrading existing capabilities, but this remains unverified for our larger, more heterogeneous library.

These limitations suggest several directions for future work. More explicit workflow planning, targeted instruction tuning, and tool-selection verification may help address ambiguity between specialists and failures to identify implicit intermediate operations. Future systems should also support longer, adaptive workflows in which tool selection responds to intermediate findings, multiple execution paths are permitted, and workflows terminate once sufficient evidence is obtained. Beyond benchmark-based evaluation, clinician-designed cases and prospective clinical studies are needed to assess orchestration reliability under realistic variation in patient data, task requirements, and workflow complexity.

In conclusion, \textsc{MMedAgent-Plus} demonstrates the feasibility of coordinating heterogeneous specialist medical models through an instruction-tuned vision-language planner. Integrating 18 specialist tools across 12 tasks and 8 modalities, the system achieves high tool-selection accuracy and generates faithful, complete responses grounded in specialist outputs. End-to-end evaluation further demonstrates the preservation of specialist-level performance on selected tasks, while highlighting the effects of routing and post-invocation errors. These findings support tool orchestration as a complementary approach to generalist medical AI, although longer adaptive workflows and prospective clinical validation remain necessary before deployment in clinical practice.

%% file: sections/Methods.tex
\section{Methods}

\subsection{System Overview}

\textsc{MMedAgent-Plus} is a multimodal medical tool-use agent that integrates a vision-language planner $\mathcal{M}$ with a library of specialist medical tools $\mathcal{T} = \{t_1, \ldots, t_K\}$, where $K$ denotes the number of available tools. Rather than incorporating all medical capabilities into a single model, the framework coordinates medical foundation models and task-specific specialists with heterogeneous input and output interfaces. Given a medical request, the planner determines whether specialist tools are needed, selects the appropriate tools, and coordinates their execution to support the requested task or workflow.

At inference, $\mathcal{M}$ receives a multimodal input $x = (q, \mathcal{I})$, where $q$ denotes the user query and $\mathcal{I}$ denotes the set of input images, which may be empty for text-only tasks. The planner also incorporates the dialogue history and any outputs returned by previously executed tools. Based on this context, $\mathcal{M}$ generates structured tool calls when external execution is required. The selected specialists are executed, and their outputs are returned to the planner to inform subsequent tool selections or final response generation. This iterative process enables the system to coordinate specialist operations across dialogue turns and incorporate intermediate results into subsequent steps. The following sections detail the planner architecture and fine-tuning (Section~\ref{sec:backbone}), instruction-tuning data construction (Section~\ref{sec:dataset}), and specialist tool library and interfaces (Section~\ref{sec:tools}).

\subsection{Backbone Model and Interaction Framework}
\label{sec:backbone}

\textsc{MMedAgent-Plus} employs Qwen2.5-VL-7B~\cite{qwen2.5-VL} as its vision-language tool-use planner $\mathcal{M}$. The planner generates tool calls in JSON format, specifying the specialist tool and its input parameters. For example, a MedSAM call takes the form \texttt{\{"API\_name": "MedSAM", "API\_params": \{"boxes": [[x1,y1,x2,y2]]\}\}}. During multi-round interactions, the planner receives the dialogue history and outputs from previously executed tools, including segmentation masks, bounding boxes, and generated reports, to inform subsequent tool selection and response generation.

We adapt $\mathcal{M}$ for medical tool orchestration through supervised instruction tuning on the curated dataset $\mathcal{S}$ (Section~\ref{sec:dataset}). The vision-language projector and language decoder are fine-tuned while the vision encoder remains frozen. Training minimizes the causal language modeling loss in Eq.~\eqref{eq:loss} over generated sequences comprising structured tool calls and natural-language responses.

\begin{equation}
    \mathcal{L} = -\frac{1}{N} \sum_{i=1}^{N} 
    \log P_{\mathcal{M}}\!\left(y_i \mid y_{<i},\, x\right),
    \label{eq:loss}
\end{equation}

where $y_i$ denotes the $i$-th output token and $x$ represents the multimodal context, including the user input, dialogue history, images, and available tool outputs. This formulation allows $\mathcal{M}$ learn tool selection, tool sequencing, and response generation through instruction tuning.

\subsection{Dataset Curation}
\label{sec:dataset}

We construct a multimodal instruction-tuning corpus $\mathcal{S}$ to train the planner $\mathcal{M}$ for specialist tool selection and workflow orchestration (Table~\ref{tab:dataset_overview}). The corpus comprises three dialogue settings: \textit{single-tool multi-round}, which trains individual specialist invocation across dialogue turns; \textit{multi-tool multi-round}, which supports sequential tool selection based on preceding dialogue and available tool outputs; and \textit{multi-tool single-round}, which requires the planner to determine a complete sequence of specialist tools from a single request. Each sample comprises a user query, optional medical images and preceding tool outputs, and a target sequence containing structured tool calls, natural-language responses, or both. Training and evaluation data are separated according to the protocol described in Section~\ref{sec:implementation}.

\subsubsection{Data Sources}

We curate training samples from 32 publicly available medical datasets and benchmarks spanning text, X-ray, CT, MRI, ultrasound, retinal imaging, and pathology. The corpus covers a broad range of medical tasks, including question answering, named entity recognition, retrieval-augmented generation, visual question answering, classification, detection, segmentation, registration, reconstruction, super-resolution, and report generation.

Text-based tasks draw on datasets associated with PMC-LLaMA~\cite{wu2024pmc}, PubMedQA, and RaTE-NER~\cite{zhao2024ratescore}. Visual understanding samples are derived from resources associated with LLaVA-Med~\cite{li2023llava}, BiomedCLIP~\cite{zhang2023biomedclip}, CONCH~\cite{lu2024visual}, and DSMIL~\cite{li2021dual}. Segmentation samples are curated from benchmarks used in MedSAM~\cite{ma2024segment} for general medical imaging, UltraSAM~\cite{meyer2024ultrasam} for ultrasound, IterNet~\cite{li2020iternet} for retinal fundus imaging, and CellViT~\cite{horst2023cellvit} and CellSAM~\cite{marks2025cellsam} for pathology. Image transformation data cover registration and reconstruction or super-resolution, using resources associated with UniGradICON~\cite{tian2024unigradicon} and HealthGPT~\cite{lin2025healthgpt}, respectively. Report generation samples are drawn from datasets associated with LLaVA-Rad~\cite{chaves2024towards} for radiology and SpecialistVLMs~\cite{holland2024specialized} for retinal OCT. The complete list of data sources and sample counts is provided in Table~\ref{tab:dataset_details}.

\subsubsection{Instruction Dataset Construction}
The instruction-tuning corpus $\mathcal{S}$ comprises 208{,}928 samples organized into three dialogue settings. The instruction format, including user inputs, structured tool calls, returned tool outputs, and final responses, is detailed in Supplementary Section~\ref{sup:instruction_schema}.

$\mathcal{S}_1$ contains 97{,}055 single-tool multi-round dialogues. Each dialogue is associated with one specialist tool $t_k \in \mathcal{T}$ and trains $\mathcal{M}$ to determine when the tool should be invoked during an ongoing conversation. Tool-use turns are constructed from the corresponding task-specific datasets, with general conversational turns randomly inserted before or after the tool-use turn. This design exposes the planner to contexts both requiring and not requiring specialist invocation, while preserving dialogue continuity across multiple turns.

$\mathcal{S}_2$ contains 90{,}000 multi-tool multi-round dialogues, each involving two to four tool invocations distributed across multiple turns. Dialogues are constructed from seven abstract task-chain templates combined with four conversation scenarios: a single image reused throughout; two images provided initially for tasks such as registration; a second image introduced in a later turn; and a switch to an unrelated image during the dialogue. Each abstract task is mapped to a specialist through a tool-modality routing table, allowing the same template to yield different tool sequences across modalities (Supplementary Table~\ref{tab:chains_s2}). The resulting dialogues follow two interaction patterns. In \textit{dependent workflows}, successive requests concern the same medical study, with later tools operating on findings returned by earlier specialists. Intermediate outputs, such as segmentation masks, reports, and extracted entities, are retained in the dialogue context for subsequent decisions. In \textit{independent workflows}, each turn introduces a separate request with its own image or text, requiring the planner to select tools according to the current request rather than reuse previous selections. Candidate tool chains are constrained by modality compatibility in both patterns and by input-output interface compatibility in dependent workflows. All generated $\mathcal{S}_2$ dialogues undergo three levels of automatic validation. \textit{Structural validation} checks dialogue-input consistency, including image-placeholder counts and tool-specific image requirements. \textit{Modality validation} verifies that each selected tool supports its assigned input modality according to the compatibility matrix. \textit{Interface validation} checks whether outputs from earlier tools can serve as valid inputs to downstream operations in dependent workflows. Dialogues failing any applicable check are excluded. The chain templates, interaction patterns, and corresponding tool combinations are detailed in Supplementary Table~\ref{tab:chains_s2}.

$\mathcal{S}_3$ contains 21{,}873 multi-tool single-round instructions, each requiring $\mathcal{M}$ to select and order a complete sequence of two to three specialist tools from a single request. Unlike $\mathcal{S}_2$, the planner must determine the entire sequence before any specialist output becomes available. Chain construction follows the same modality and interface compatibility constraints as $\mathcal{S}_2$. It combines multimodal classification, question answering, segmentation, and report generation, including both two-tool and three-tool sequences. Intermediate tool outputs sampled from $\mathcal{S}_1$ are incorporated during dataset construction to provide realistic inputs for downstream operations, but are not available to the planner when it initially selects the complete tool sequence.

\subsubsection{Dialogue Generation and Data Quality Control}
\label{sec:synthesis}

New instruction-tuning dialogues are constructed through a template-based procedure without per-sample language model generation. For each task, predefined banks of user-request phrasings and assistant-response templates are created in a single offline pass using GPT-4~\cite{achiam2023gpt} and subsequently fixed (Supplementary Table~\ref{tab:template_banks}). Single-tool samples are generated by pairing a randomly selected request template with a case from the corresponding source benchmark. Target responses combine structured tool calls, generated according to the tool registry schema, with response templates populated using the reference outputs of the source cases. Multi-tool dialogues are assembled from these samples according to the predefined chain templates and conversation scenarios, with transitions between turns drawn from a fixed phrase bank. A fixed random seed is used for sampling, and the construction scripts and template inventories are provided to support reproducibility.

The instruction-tuning corpus combines data from three sources. For the twelve newly introduced tools, dialogues are synthesized using the template-based procedure described above. For the six tools inherited from MMedAgent~\cite{li2024mmedagent}, we retain the original instruction data, which were generated using GPT-4 and source-dataset annotations according to the procedure described in that work. For template-based synthesis, medical inputs and reference outputs, including images, questions, labels, reports, entity annotations, and segmentation masks, are taken from the source benchmarks rather than generated by GPT-4. The language model is used only to construct the natural-language templates, while structured tool calls follow the predefined tool registry. Generated multi-tool dialogues are further checked for structural consistency, modality compatibility, and input-output interface compatibility as described above. Tool-specific data provenance and sample counts are reported in Supplementary Table~\ref{tab:dataset_details}.

\begin{table}[ht]
\centering
\caption{Composition of the instruction-tuning corpus across three dialogue settings. Single-tool multi-round samples are grouped by specialist tool, whereas multi-tool settings are reported by their total sample counts. Tool-chain configurations are detailed in Supplementary Table~\ref{tab:chains_s3}-~\ref{tab:chains_s2}.}

\label{tab:dataset_overview}
\small
\setlength{\tabcolsep}{6pt}
\renewcommand{\arraystretch}{1.15}
\begin{tabular}{l l r}
\hline\hline
\textbf{Category} & \textbf{Tool} & \textbf{\#Samples} \\
\hline
\multirow{20}{*}{\textbf{Single-tool Multi-round}} 
 & PMC-LLaMA~\cite{wu2024pmc} & 9,000 \\
 & RaTE-NER~\cite{zhao2024ratescore} & 4,500 \\
 & ChatCAD-R~\cite{wang2024interactive} & 2,528 \\
 & LLaVA-Med~\cite{li2023llava} & 16,687 \\
 & BiomedCLIP~\cite{zhang2023biomedclip} & 7,923 \\
 & Grounding DINO~\cite{liu2024grounding} & 4,207 \\
 & Grounding DINO~\cite{liu2024grounding} + MedSAM~\cite{ma2024segment} & 2,303 \\
 & MedSAM~\cite{ma2024segment} & 5,765 \\
 & UltraSAM~\cite{meyer2024ultrasam} & 5,000 \\
 & IterNet~\cite{li2020iternet} & 4,500 \\
 & UNIGradICON~\cite{tian2024unigradicon} & 5,000 \\
 & HealthGPT~\cite{lin2025healthgpt} (For Reconstruction) & 6,147 \\
 & HealthGPT~\cite{lin2025healthgpt} (For Super Resolution) & 4,593 \\
 & LLaVA-Rad~\cite{chaves2024towards} & 4,500 \\
 & SpecialistVLMs~\cite{holland2024specialized} & 3,602 \\
 & CONCH~\cite{lu2024visual} & 3,600 \\
 & DSMIL-TCGA~\cite{li2021dual} & 1,800 \\
 & DSMIL-C16~\cite{li2021dual} & 1,800 \\
 & CellViT~\cite{horst2023cellvit} & 1,800 \\
 & CellSAM~\cite{marks2025cellsam} & 1,800 \\
 & \textbf{Subtotal} & \textbf{97,055} \\
\hline
\textbf{Multi-tool Single-round} & & 21,873 \\
\textbf{Multi-tool Multi-round}  & & 90,000 \\
\hline
\textbf{Total} &  & \textbf{208,928} \\
\hline\hline
\end{tabular}
\end{table}

\subsection{Tool Integration}
\label{sec:tools}

The tool library $\mathcal{T}$ comprises 18 specialist medical models grouped into four functional categories: (i) text understanding and generation, including question answering, named entity recognition, and retrieval-augmented generation; (ii) visual understanding, including visual question answering and image classification; (iii) image processing, including object detection, segmentation, grounding-segmentation, image registration, reconstruction, and super-resolution; and (iv) medical report generation, covering radiology and retinal OCT reporting. Collectively, these tools span 12 medical tasks and 8 input modalities: text, X-ray, CT, MRI, ultrasound, retinal fundus imaging, retinal OCT, and pathology.

The library integrates medical foundation models and task-specific specialists selected for their reported performance and public availability in the corresponding tasks and modalities. Each tool is registered through a standardized invocation interface specifying its supported input modalities, required parameters, and output types. This common interface allows the planner $\mathcal{M}$ to invoke specialists with heterogeneous architectures and input-output formats. It also supports multi-step workflows in which outputs from upstream tools can be passed to downstream tools when their interfaces are compatible. The complete inventory of specialist models, supported tasks and modalities, and associated data sources is provided in Supplementary Table~\ref{tab:dataset_details}.

\subsubsection{Text Understanding and Generation}

\textbf{QA \& Rationale Generation.}
We use PMC-LLaMA \cite{wu2024pmc} as the tool of QA and rationale generation. To incorporate domain knowledge, PMC-LLaMA undergoes initial pretraining on the MedC-K medical corpus, then is fine-tuned with instructions using the MedC-I domain-specific instruction dataset. This two-stage training approach ensures that PMC-LLaMA performs effectively across a variety of medical reasoning tasks.

\textbf{Named Entity Recognition.}
We employ RaTE-NER \cite{zhao2024ratescore}, a named entity recognition (NER) model designed for understanding radiology reports. RaTE-NER is trained on a newly introduced dataset comprising radiology reports annotated for key medical entities, including anatomical structures, findings, diagnoses, and negations. The model is designed to reflect human reasoning by aligning with RaTEscore, a framework that captures clinically relevant relationships between entities and report quality.

\textbf{Retrieval-Augmented Generation.}
We integrate ChatCAD-R \cite{wang2024interactive}, a retrieval-augmented generation framework designed to combine free-text radiology knowledge with structured databases. ChatCAD-R retrieves relevant entries from natural language text format. It merge the strengths of LLMs' medical domain knowledge and logical reasoning with the vision understanding capability of existing medical-image CAD models.

\subsubsection{Visual Understanding}

\textbf{Visual Question Answering (All).}
We adopt LLaVA-Med \cite{li2023llava} as the backbone tool for medical VQA. LLaVA-Med is trained on multimodal pairs curated from PMC articles, covering MRI, CT, histology, gross images, and X-rays. Its visual encoder is aligned with biomedical language corpora, allowing it to handle both conversational queries and detailed image-centric questions. LLaVA-Med shows excellent multimodal conversational capability and can follow open-ended instruction to assist with inquiries about a biomedical image.

\textbf{Classification (All).}
We employ BiomedCLIP \cite{zhang2023biomedclip}, a biomedical foundation model trained on 15 million scientific image–text pairs, as the classification tool. BiomedCLIP achieves state-of-the-art performance in zero-shot and fine-grained classification across diverse imaging modalities. It classifies input images based on similarity between visual and textual embeddings, enabling robust cross-modal understanding.

\textbf{Classification (Pathology).}
We utilize CONCH \cite{lu2024visual}, a specialized visual-language foundation model trained on over 1.17 million diverse histopathology image-caption pairs, for zero-shot pathology classification. We adapt CONCH to answer diagnostic queries by matching image features to a predefined dictionary of textual descriptions. It leverages variations of pathological class names (e.g., tumor subtypes, tissue types, or morphological features) inserted into standard templates to create ensembled text embeddings for each category. Images are then evaluated based on the highest cosine similarity in the shared image-text embedding space.

\subsubsection{Image Processing}

\textbf{Tumor Detection and Subtyping (Pathology).}
We employ DSMIL \cite{li2021dual}, a deep multiple instance learning network, for localized tumor detection and classification in whole slide images. To support distinct diagnostic tasks, we adapt the model into two specialized tools: DSMIL-C16 and DSMIL-TCGA. DSMIL-C16 grounds on the Camelyon16 dataset to serve as a general tumor classifier, distinguishing malignant from benign histology. DSMIL-TCGA grounds on the TCGA dataset to perform lung cancer malignancy classification and subtyping, such as differentiating adenocarcinoma from squamous cell carcinoma.

\textbf{Grounding (All).}
We use Grounding DINO \cite{liu2024grounding} as our object detection tool. It merges Transformer-based detector DINO with grounded pre-training, fine-tuned for medical applications to localize anatomical structures, lesions, and cells across MRI, CT, X-ray, and histology. By transforming segmentation masks into bounding boxes during fine-tuning, the model learns to generalize across modalities and disease categories, achieving accurate detection and localization for downstream tasks.

\textbf{Segmentation (All).}
For general segmentation, we adopt MedSAM \cite{ma2024segment}, a domain-adapted version of the Segment Anything Model (SAM) \cite{kirillov2023segment} fine-tuned on diverse medical datasets spanning MRI, CT, X-ray, histology, and gross images. It supports box- and point-based prompts, enabling interactive and delivering accurate and efficient segmentation across modalities.

\textbf{Grounding-Segmentation (All).}
For grounding-segmentation (G-Seg), we chain Grounding DINO with MedSAM. Grounding DINO first localizes the region of interest based on text prompts, and MedSAM refines these into pixel-level segmentation masks. This combination supports clinically realistic workflows in which a physician specifies an object of interest and the agent automatically outputs precise boundaries.

\textbf{Segmentation (Ultrasound).} We use UltraSAM \cite{meyer2024ultrasam}, a Segment Anything–based visual foundation model for ultrasound image segmentation. UltraSAM was trained on US-43d, the largest public ultrasound dataset, and addressed anatomical complexity and limited annotations. It supports point- and box-based prompts, serving as a flexible backbone for various tasks.

\textbf{Segmentation (Fundus).}
We use IterNet \cite{li2020iternet} for retinal vessel segmentation tasks. IterNet extends the U-Net architecture, extracts fine-grained vascular details by iteratively refining the output of previous segmentations. The model comprises multiple mini-U-Net blocks, allowing it to achieve depths of up to four times that of standard U-Net.

\textbf{WSI Cell Segmentation (Pathology).}
We employ CellViT \cite{horst2023cellvit} for automated cell nuclei instance segmentation in gigapixel whole-slide images (WSIs). The model utilizes a U-Net-shaped encoder-decoder architecture where a ViT serves as the image encoder. CellViT is trained and evaluated on the PanNuke dataset. It automatically preprocesses the WSI by extracting manageable image patches, and subsequently performs precise, large-scale cell segmentation and detection across the entire slide. Our framework deploys the pre-trained SAM-H backbone variant.

\textbf{2D Cell Segmentation (Pathology).}
We employ CellSAM \cite{marks2025cellsam} for automated cell instance segmentation in 2D images. Built upon SAM foundation, CellSAM integrates a specialized object detector, CellFinder, which utilizes an Anchor DETR framework to automatically generate bounding box prompts for SAM's mask decoder. The model is trained on a diverse corpus of cellular imaging data-incorporating datasets.

\textbf{Registration (All).}
We use uniGradICON \cite{tian2024unigradicon} as a radiology registration tool. The model fuses the performance of deep learning with the adaptability of traditional methods. It enables zero-shot registration on untrained tasks and offers strong generalization across diverse datasets, anatomical regions, imaging modalities, and acquisition protocols.

\textbf{Image Reconstruction \& Super Resolution (All).}
We use HealthGPT \cite{lin2025healthgpt} for image reconstruction and super resolution tasks. This model is designed to perform both comprehension and generation tasks in complex medical scenarios. HealthGPT built upon a multi-LoRA PEFT architecture, employs a task-gated decoupling strategy to mitigate data conflict and enhance multi-task learning under limited supervision.

\subsubsection{Medical Report Generation}

\textbf{Report Generation (All).}
We use LLaVA-Rad \cite{chaves2024towards} for generating radiology reports. It offers fast inference and efficient fine-tuning, which aligns well with visual abnormalities and report content. Its modular design bridges multimodal performance gaps while maintaining low computational costs, facilitating practical clinical integration.

\textbf{Report Generation (Retina OCT).}
We use SpecialistVLMs \cite{holland2024specialized} for OCT report generation. This model was trained on domain-specific OCT images along with expert annotations, allowing for accurate identification of retinal structures and pathologies.

\subsection{Implementation Details}
\label{sec:implementation}

\subsubsection{Model Training}

The planner $\mathcal{M}$ is initialized from Qwen2.5-VL-7B~\cite{qwen2.5-VL} and fine-tuned on the instruction-tuning corpus $\mathcal{S}$ (Section~\ref{sec:dataset}) using the objective in Eq.~\ref{eq:loss}. The vision encoder is kept frozen, while the vision-language projector and language decoder are updated during training.

We use AdamW~\cite{loshchilov2019decoupled} with a learning rate of $1 \times 10^{-5}$, weight decay of $0.01$, and a maximum gradient norm of $1.0$. The learning rate follows a cosine decay schedule with linear warmup over the first 10\% of optimization steps. Training is performed for one epoch using bfloat16 precision and gradient checkpointing on four NVIDIA A100-SXM4-80GB GPUs. A per-device batch size of 1 and gradient accumulation over 64 steps yield an effective global batch size of 256.

\subsubsection{Deployment}
\label{sec:deployment}

\textsc{MMedAgent-Plus} is deployed as a set of containerized services coordinated by a central gateway. The planner $\mathcal{M}$ and each specialist tool $t_k \in \mathcal{T}$ run as separate services, with specialist tools exposed through standardized invocation interfaces. For each user request, the gateway forwards the multimodal input and dialogue context to the planner, parses the generated structured tool calls, and dispatches them to the specialist services. Tool outputs are returned to the planner for subsequent tool selection or final response generation. Multi-tool workflows are executed through the gateway, which coordinates the invocation of successive specialists according to the tool sequence specified by the planner. This modular architecture allows specialist implementations to be updated or replaced independently, provided that their interfaces remain compatible.

\subsubsection{Evaluation Protocol}
\label{sec:evaluation}

We evaluate \textsc{MMedAgent-Plus} across three aspects: tool orchestration accuracy, final response quality, and end-to-end task performance. The evaluations assess the correctness of specialist selection and sequencing, the quality of generated responses, and task-specific performance relative to direct-answer baselines and standalone specialist models.

\paragraph{Tool Orchestration.}
We assess tool orchestration using five complementary metrics at the individual tool-call and dialogue levels.

\textbf{Tool Call Accuracy.}
Tool call accuracy measures the proportion of correctly selected tools across individual invocations:

\begin{equation}
    \text{Tool Call Accuracy} = \frac{1}{N} \sum_{i=1}^{N} 
    \mathbb{1}\bigl[\hat{t}_i = t_i\bigr],
    \label{eq:tool_acc}
\end{equation}

where $t_i$ and $\hat{t}_i$ denote the reference and predicted tool identities for the $i$-th invocation, respectively, and $N$ is the total number of reference tool calls across the evaluation cases.

\textbf{Whole-Case Accuracy.}
Whole-case accuracy measures the proportion of dialogues in which the predicted tool sequence exactly matches the reference sequence, including both tool identities and execution order:

\begin{equation}
    \text{Whole-Case Accuracy}
    =
    \frac{1}{M}
    \sum_{j=1}^{M}
    \mathbb{1}
    \bigl[
    \hat{\mathbf{t}}_j = \mathbf{t}_j
    \bigr],
    \label{eq:case_acc}
\end{equation}

where $M$ denotes the number of evaluation dialogues, $\mathbf{t}j = (t{j,1}, \ldots, t_{j,n_j})$ is the ordered reference tool sequence for dialogue $j$, and $\hat{\mathbf{t}}_j$ is the corresponding predicted sequence. A dialogue is counted as correct only if all required tools are selected in the correct order, with no missing, additional, or incorrectly ordered calls. This metric provides a stricter assessment of orchestration correctness than tool-call accuracy in multi-tool settings.

\textbf{Precision, Recall, and F1.}
We compute precision, recall, and F1 to characterize tool selection errors independently of invocation order. Each specialist tool $t_k \in \mathcal{T}$ is treated as a binary label indicating its presence in a dialogue. A true positive ($TP_k$) occurs when $t_k$ is both required and predicted, a false positive ($FP_k$) when it is predicted but not required, and a false negative ($FN_k$) when it is required but omitted. These counts are aggregated across evaluation dialogues for each tool.

Tool-specific precision, recall, and F1 are defined as:

\begin{equation}
P_k = \frac{TP_k}{TP_k + FP_k}, \qquad
R_k = \frac{TP_k}{TP_k + FN_k}, \qquad
F1_k = \frac{2P_kR_k}{P_k + R_k}.
\end{equation}

Macro-averaged metrics assign equal weight to each specialist tool:

\begin{equation}
P_{\mathrm{macro}} = \frac{1}{K}\sum_{k=1}^{K}P_k, \qquad
R_{\mathrm{macro}} = \frac{1}{K}\sum_{k=1}^{K}R_k, \qquad
F1_{\mathrm{macro}} = \frac{1}{K}\sum_{k=1}^{K}F1_k.
\end{equation}

Micro-averaged metrics aggregate true positives, false positives, and false negatives across all tools before computing the scores:

\begin{equation}
P_{\mathrm{micro}} =
\frac{\sum_{k=1}^{K}TP_k}
{\sum_{k=1}^{K}(TP_k+FP_k)}, \qquad
R_{\mathrm{micro}} =
\frac{\sum_{k=1}^{K}TP_k}
{\sum_{k=1}^{K}(TP_k+FN_k)},
\end{equation}

\begin{equation}
F1_{\mathrm{micro}} =
\frac{2P_{\mathrm{micro}}R_{\mathrm{micro}}}
{P_{\mathrm{micro}}+R_{\mathrm{micro}}}.
\end{equation}

Precision reflects the frequency of unnecessary tool selections, whereas recall reflects the frequency of omitted tools. Macro averaging weights specialists equally, while micro averaging summarizes performance across tool-presence decisions. Unlike whole-case accuracy, these metrics do not account for invocation order or repeated calls to the same tool within a dialogue.

\textbf{Per-Task and Per-Modality Accuracy.}
To assess variation in orchestration performance, we report tool-call accuracy stratified by medical task and input modality. Task categories include segmentation, report generation, visual question answering, named entity recognition, etc. Modality groups include CT, MRI, X-ray, ultrasound, retinal fundus, OCT, pathology, and text. These analyses identify tasks and modalities associated with higher rates of tool-selection errors.

\textbf{Error Pattern Analysis.}
We further characterize orchestration errors by examining row-normalized recall and residual incorrect assignments at the tool and task levels (Figure~\ref{fig:routing}). Row-normalized recall measures the proportion of reference calls correctly assigned to each tool or task, while residual assignments identify alternative predictions when the correct tool is not selected. These errors include substitutions between specialists with overlapping capabilities and omissions of required tool calls, represented by \texttt{no tool}. Modality-level residuals are also examined to identify errors associated with mismatches between reference and predicted input modalities (Section~\ref{sec:orchestration}).

Together, these evaluations provide a complementary assessment of tool orchestration reliability. Tool-call accuracy and precision, recall, and F1 quantify selection performance at the invocation level, while whole-case accuracy assesses whether the complete tool sequence is correctly selected and ordered. Task- and modality-stratified analyses reveal where errors are concentrated, providing insight into the limitations of tool selection across heterogeneous medical tasks.

\paragraph{Task-Specific Tool Output Quality.}

We assess end-to-end task performance using seven representative specialist tools spanning text-based and image-based medical tasks: PMC-LLaMA for question answering and rationale generation, RaTE-NER for named entity recognition, CONCH for pathology classification, BiomedCLIP for multimodal classification, IterNet for retinal fundus segmentation, MedSAM for general medical image segmentation, and LLaVA-Rad for radiology report generation. Each tool is evaluated using task-specific metrics appropriate to its output format, including textual responses, entity annotations, class predictions, segmentation masks, and radiology reports (Table~\ref{tab:e2e_metrics}). These metrics are used to compare the outputs of \textsc{MMedAgent-Plus} with those of the corresponding standalone specialists.

For radiology report generation, we combine lexical similarity metrics with clinical measures to capture complementary aspects of report quality~\cite{liu2019clinically}. ROUGE-L~\cite{lin2004rouge} and BLEU-4~\cite{papineni2002bleu} quantify textual overlap with reference reports. CheXpert F1~\cite{irvin2019chexpert} measures agreement on extracted clinical observations, while RadGraph F1~\cite{jain2021radgraph} evaluates agreement on clinical entities and their relations. The latter metrics provide a more clinically focused assessment than lexical overlap.

\begin{table}[h]
\centering
\caption{Task-specific evaluation metrics used in end-to-end assessment. 
Tasks span both text and image modalities; metrics are selected to reflect 
clinically relevant aspects of output quality. QA, question answering; 
NER, named entity recognition; MRI, magnetic resonance imaging; 
CT, computed tomography; US, ultrasound; HD95, 95th-percentile Hausdorff distance.}
\label{tab:e2e_metrics}
\setlength{\tabcolsep}{2pt}
\begin{tabular}{llll}
\toprule
\textbf{Task} & \textbf{Tool} & \textbf{Modality} & \textbf{Metric(s)} \\
\midrule
QA / Rationale Generation & PMC-LLaMA~\cite{wu2024pmc}  & Text       & ROUGE-L, Accuracy \\
Named Entity Recognition   & RaTE-NER~\cite{zhao2024ratescore}   & Text       & RadGraph F1, Entity-level F1 \\
Classification (Pathology)           & CONCH~\cite{lu2024visual}      & Pathology  & Micro F1, Macro F1 \\
Classification            & BiomedCLIP~\cite{zhang2023biomedclip} & X-Ray, MRI, CT, US & Micro F1, Macro F1 \\
Segmentation (Fundus)     & IterNet~\cite{li2020iternet}    & Retina     & Dice, HD95 \\
Segmentation (General)    & MedSAM~\cite{ma2024segment}     & X-Ray, MRI, CT, US & Dice, HD95 \\
Report Generation         & LLaVA-Rad~\cite{chaves2024towards}  & X-Ray, MRI, CT, US & ROUGE-L, BLEU-4, \\
                          &            &            & CheXpert F1, RadGraph F1 \\
\bottomrule
\end{tabular}
    \end{table}

\paragraph{Final Answer Quality.}

Accurate tool selection does not guarantee that specialist outputs are faithfully incorporated into the final response. We therefore assess response quality using GPT-5.4-mini~\cite{openai2026gpt54mini} as an LLM judge. Given the dialogue history and returned specialist outputs, \textsc{MMedAgent-Plus} generates a natural-language response that is evaluated on three criteria, each scored on a five-point integer scale.

\begin{itemize}

    \item \textbf{Faithfulness} assesses whether the response accurately represents the returned specialist outputs without introducing unsupported claims or contradictions. This criterion is evaluated against the tool outputs alone, without reference to the ground-truth answer.

    \item \textbf{Completeness} assesses whether the response includes the relevant information from the specialist outputs needed to address the user request. The reference answer is used to identify task-relevant information rather than as a target for lexical matching.

    \item \textbf{Conciseness} assesses whether the response avoids irrelevant or redundant content while retaining the information necessary to address the user request. The reference answer serves only as a guide to the expected scope of the response.
    
\end{itemize}

We report the mean score for each criterion across evaluation dialogues. The three criterion scores are also summed to obtain a combined score ranging from 3 to 15, which serves as a secondary summary metric. For task- and modality-level visualizations, the combined score is normalized by the maximum possible score of 15 and expressed as a percentage. To reduce interference between criteria, each criterion is evaluated in a separate judge call using a dedicated rubric and three scored few-shot examples.

Two evaluation rules accommodate different output formats and workflow structures. For tools producing non-text outputs, including MedSAM, IterNet, HealthGPT, and UniGradICON, the generated image, mask, or file is treated as the primary output. A brief acknowledgment of successful output generation is considered sufficient when the user request does not require a textual description. For multi-tool workflows, response quality is assessed primarily against the terminal tool output. Intermediate outputs are not required to be separately described unless explicitly requested by the user.

We additionally identify two types of response-generation errors. A \textit{wrong-context response} occurs when $\mathcal{M}$ produces a tool-call-style output instead of synthesizing a final response after tool execution. Such responses are assigned the minimum score of 1 for each criterion without LLM-judge evaluation. A \textit{redundant tool call} occurs when an additional tool invocation accompanies an otherwise substantive final response. These responses are scored normally and separately recorded for error analysis.

For each evaluation instance, the judge receives the tool name, original user request, terminal tool output, generated response, reference answer, and indicators for non-text outputs and multi-tool workflows. Intermediate reasoning traces and training annotations are excluded. Each criterion is assessed in a separate judge call using a dedicated rubric and three scored few-shot examples to minimize interference between criteria. The judge returns an integer score and a brief justification. Judge responses are cached using input hashes to ensure consistent reuse of previously obtained evaluations.

\paragraph{Baseline Comparison.}
\label{sec:baseline}

We compare \textsc{MMedAgent-Plus} with four direct-answer multimodal models: Qwen3-VL-8B~\cite{bai2025qwen3vltechnicalreport}, LLaVA-Med~\cite{li2023llava}, GPT-5.4-mini~\cite{openai2026gpt54mini}, and MedGemma-1.5-4B-IT~\cite{sellergren2026medgemma15technicalreport}. These baselines span general-purpose and medical-domain vision-language models. The comparison assesses the final response quality of specialist-assisted generation relative to direct generation from the original multimodal input. Answer generation and response evaluation are conducted separately.

\textbf{Answer generation.}
Each baseline receives the original user request and associated medical images, without access to specialist outputs, intermediate workflow results, or reference answers. The models are instructed to respond directly and concisely using the provided inputs. Images are supplied in the format required by each model. LLaVA-Med processes only the first image in multi-image cases, whereas Qwen3-VL-8B and MedGemma-1.5-4B-IT accept multiple images within a single request. Locally hosted models use deterministic decoding with a maximum of 1,024 new tokens, matching the generation limit used for \textsc{MMedAgent-Plus}. API-based models use the most deterministic decoding configuration supported by their interfaces.

\textbf{Evaluation.}
Baseline responses are evaluated using GPT-5.4-mini with the same three criteria of faithfulness, completeness, and conciseness described above. Detailed evaluation rubrics and score visualizations are provided in Supplementary Section~\ref{sec:supp_eval}. Because direct-answer baselines do not receive specialist outputs, faithfulness is assessed against the original multimodal input rather than returned tool results. Reference answers are excluded from faithfulness evaluation but are used to identify task-relevant information and expected response scope for completeness and conciseness, respectively. All responses are scored on the same five-point scales using criterion-specific rubrics and three scored few-shot examples. We report the mean score for each criterion and the combined score.

This comparison evaluates specialist-assisted response generation against direct-answer models operating without tool outputs. Although the evaluation uses the same requests and scoring criteria, differences in model architectures, training, and available input evidence mean that the observed performance gaps cannot be attributed solely to specialist tool access.

\paragraph{End-to-end Task Evaluation.}

To assess end-to-end task execution without providing reference tool calls or oracle tool outputs, we construct an evaluation set from the held-out single-tool multi-round corpus described in Section~\ref{sec:dataset}. From a pool of 9{,}128 held-out dialogues, we select 50 cases for each of seven specialist tools, yielding 350 cases in total. The selected tools are PMC-LLaMA, RaTE-NER, LLaVA-Rad, CONCH, BiomedCLIP, IterNet, and MedSAM, covering question answering and rationale generation, named entity recognition, radiology report generation, pathology classification, multimodal classification, retinal vessel segmentation, and general medical image segmentation. Cases are sampled using a fixed random seed of 42, except for BiomedCLIP, for which a predefined subset of 50 cases is used.

We first assess specialist invocation across all 350 cases. Each dialogue is replayed turn by turn, and the tool selected by \textsc{MMedAgent-Plus} is compared with the reference tool. Overall and per-tool invocation accuracy are calculated using the tool-call accuracy metric defined in Section~\ref{sec:evaluation}.

We then evaluate final response quality for the 250 cases involving PMC-LLaMA, RaTE-NER, CONCH, BiomedCLIP, and LLaVA-Rad. IterNet and MedSAM are excluded from direct-answer baseline comparisons because their primary outputs are segmentation masks, which the evaluated baseline VLMs do not generate. For each case, baselines receive the user request immediately preceding the reference tool invocation and its associated images. Baseline responses are generated according to Section~\ref{sec:baseline} and assessed using the LLM-as-a-judge protocol described above.

Finally, we assess end-to-end task performance across all seven tools using task-specific metrics, including Dice and HD95 for segmentation, ROUGE and BLEU for generated text, and micro- and macro-averaged F1 for structured predictions (Table~\ref{tab:e2e_metrics}). For PMC-LLaMA, RaTE-NER, and LLaVA-Rad, the same metrics are also computed from direct-answer baseline responses on the corresponding held-out cases. These evaluations complement response-quality scores by measuring how well specialist task performance is preserved during end-to-end execution.

%% file: sections/Data_Availability.tex
\section{Data availability}

All datasets used in this study are derived from publicly available medical datasets and benchmarks, with the original sources listed in Supplementary Table~\ref{tab:dataset_details}. The instruction annotations and the training and test splits are available from the MMedAgent organization on Hugging Face at \url{https://huggingface.co/MMedAgent}. The underlying medical images are not owned by the authors and are not redistributed. They should be obtained from their original repositories under the corresponding licenses. Instructions for downloading the source data and reconstructing the full dataset are provided at \url{https://github.com/RoxanneWAAANG/MMedAgent-Bench}. Further documentation is available on the project website at \url{https://mmedagent.org}.

%% file: sections/Code_Availability.tex
\section{Code availability}

The code for \textsc{MMedAgent-Plus} is released under the Apache-2.0 license in two repositories. The training code for the planner, including the fine-tuning scripts and the configuration used in this work, is available at \url{https://github.com/RoxanneWAAANG/MMedAgent-Plus}. The evaluation code, including the tool-calling evaluation and metric scripts and the LLM-as-a-judge evaluation with its prompts and rubrics, is available at \url{https://github.com/RoxanneWAAANG/MMedAgent-Bench}, together with example instruction-generation scripts and template inventories. The fine-tuned model weights are available from the MMedAgent organization on Hugging Face at \url{https://huggingface.co/MMedAgent}. An interactive demo and the project website are available at \url{https://mmedagent.org}. The deployment code, comprising the gateway and the containerized tool services, together with the task-specific metric implementations, the remaining instruction-generation scripts for the tools introduced in this work, and the evaluation sample indices required to reproduce the reported experiments, will be added to these repositories upon publication. Instruction data for the tools inherited from MMedAgent~\cite{li2024mmedagent} was produced by that work's pipeline and is available from its original release.

%% file: sections/Supplementary.tex
\section{Supplementary}

\setcounter{subsection}{0}
\renewcommand{\thesubsection}{S\arabic{subsection}}
\setcounter{table}{0}
\renewcommand{\thetable}{S\arabic{table}}
\setcounter{figure}{0}
\renewcommand{\thefigure}{S\arabic{figure}}
\setcounter{equation}{0}
\renewcommand{\theequation}{S\arabic{equation}}

\subsection{System and Tools}
\label{sup:system_and_tools}

This section details the specialist tool library $\mathcal{T}$ introduced in Section~\ref{sec:tools}. We provide the model checkpoints and licenses (Table~\ref{tab:tool_cards}), the tool-modality compatibility matrix used for dataset construction (Table~\ref{tab:modality_matrix}), and the invocation and interface compatibility rules governing tool execution and intermediate output transfer (Section~\ref{sup:interface_rules}).

\subsubsection{Tool Cards}
\label{sup:tool_cards}

Table~\ref{tab:tool_cards} summarizes the supported task, deployment checkpoint, and license for each of the 18 specialist tools. Tools are deployed as independent containerized services exposed through a common invocation interface. The grounding-segmentation tool (G-Seg) is implemented as a composite workflow that sequentially invokes Grounding DINO and MedSAM through the gateway. Each tool invocation is subject to a wall-clock timeout. Calls exceeding this limit return a structured error to the planner $\mathcal{M}$ without terminating the dialogue (Section~\ref{sup:interface_rules}).

\begin{table}[!htb]
\centering
\caption{\textbf{Tool cards for the 18 specialist tools in $\mathcal{T}$.}
For each tool, we report its supported task, deployment checkpoint, and declared license, as documented in the original code repository or model card. ‘Not specified’ indicates that no explicit license was identified in the corresponding release and does not imply permission for unrestricted use. Tools marked with a dagger ($\dagger$) are inherited from MMedAgent~\cite{li2024mmedagent} and retain their original configurations.}
\label{tab:tool_cards}

\scriptsize
\setlength{\tabcolsep}{3pt}
\renewcommand{\arraystretch}{1.25}

\begin{tabularx}{\textwidth}{
    P{2.7cm}
    P{2.4cm}
    Y
    P{1.8cm}
}
\hline\hline
\textbf{Tool} &
\textbf{Task} &
\textbf{Checkpoint} &
\textbf{License} \\
\hline

\multicolumn{4}{l}{\textit{Text understanding and generation}} \\

PMC-LLaMA~\cite{wu2024pmc}
& QA \& rationale generation
& \texttt{axiong/PMC\_LLaMA\_13B}
& OpenRAIL \\

RaTE-NER~\cite{zhao2024ratescore}
& Named entity recognition
& \texttt{Angelakeke/RaTE-NER-Deberta}; token classification with span aggregation
& MIT \\

ChatCAD-R$^\dagger$~\cite{wang2024interactive}
& Retrieval-augmented generation
& Released ChatCAD checkpoint:
  \url{https://drive.google.com/drive/folders/1Wcfspge2K0db-Fz2K9vjjFGMKX32WCvc}
& Apache-2.0 \\

\hline
\multicolumn{4}{l}{\textit{Visual understanding}} \\

LLaVA-Med$^\dagger$~\cite{li2023llava}
& Visual question answering
& \texttt{microsoft/llava-med-v1.5-mistral-7b}.
  The original LLaVA-Med v1.0.0 checkpoint is no longer accessible from the original release repository.
& Microsoft Research License \\

BiomedCLIP$^\dagger$~\cite{zhang2023biomedclip}
& Classification (all)
& \texttt{microsoft/BiomedCLIP-PubMedBERT\_256-vit\_base\_patch16\_224}
& MIT \\

CONCH~\cite{lu2024visual}
& Classification (pathology)
& ---
& --- \\

\hline
\multicolumn{4}{l}{\textit{Image processing}} \\

DSMIL-C16~\cite{li2021dual}
& Tumor detection (pathology)
& ---
& --- \\

DSMIL-TCGA~\cite{li2021dual}
& Tumor subtyping (pathology)
& ---
& --- \\

Grounding DINO$^\dagger$~\cite{liu2024grounding}
& Grounding (all)
& Grounding DINO fine-tuned on medical data by MMedAgent~\cite{li2024mmedagent};
  \url{https://drive.google.com/drive/folders/1eK27gz0tkbcp-9hx2fI9J_Wj2zHv14pL}.
  Box threshold $0.3$; text threshold $0.25$.
& Apache-2.0 \\

MedSAM$^\dagger$~\cite{ma2024segment}
& Segmentation (all)
& Released MedSAM checkpoint:
  \url{https://drive.google.com/drive/folders/1ETWmi4AiniJeWOt6HAsYgTjYv_fkgzoN}.
  Box or point prompts.
& Apache-2.0 \\

G-Seg$^\dagger$
& Grounding-segmentation (all)
& Grounding DINO followed by MedSAM, composed in the gateway
& Apache-2.0 \\

UltraSAM~\cite{meyer2024ultrasam}
& Segmentation (ultrasound)
& \texttt{UltraSam.pth} with the default released segmentation configuration
& CC BY-NC-SA 4.0 \\

IterNet~\cite{li2020iternet}
& Segmentation (fundus)
& Released IterNet weights; minimum kernel $32$, dropout $0.1$, and three refinement iterations
& MIT \\

CellViT~\cite{horst2023cellvit}
& WSI cell segmentation (pathology)
& ---
& --- \\

CellSAM~\cite{marks2025cellsam}
& 2D cell segmentation (pathology)
& ---
& --- \\

UniGradICON~\cite{tian2024unigradicon}
& Registration (all)
& Released UniGradICON weights; inputs resampled to $256 \times 256$
& Apache-2.0 \\

HealthGPT~\cite{lin2025healthgpt}
& Reconstruction \& super-resolution
& \texttt{microsoft/Phi-4} language backbone with
  \texttt{openai/clip-vit-large-patch14-336} vision tower and the released H-LoRA adapter;
  FP16, rank $32$, $\alpha=64$, four adapters, and $8192$ VQ indices
& Apache-2.0 \\

\hline
\multicolumn{4}{l}{\textit{Medical report generation}} \\

LLaVA-Rad~\cite{chaves2024towards}
& Report generation (all)
& \texttt{microsoft/llava-rad} over the \texttt{lmsys/vicuna-7b-v1.5} base
& Microsoft Research License \\

SpecialistVLMs~\cite{holland2024specialized}
& Report generation (retina OCT)
& Released RetinaVLM weights; inputs resampled to $455 \times 455$
& Not specified \\

\hline\hline
\end{tabularx}
\end{table}

\subsubsection{Tool-Modality Compatibility}
\label{sup:modality_matrix}

Table~\ref{tab:modality_matrix} presents the tool-modality compatibility matrix used during instruction dataset construction (Section~\ref{sec:dataset}). A checkmark indicates that a tool supports the corresponding input modality and that the combination is represented in the training corpus $\mathcal{S}$. An empty cell indicates that the tool is not available for that modality within the framework. During dataset construction, candidate tool calls with unsupported modality assignments are excluded. Several specialists are restricted to specific modalities, including UltraSAM for ultrasound, IterNet for retinal fundus imaging, SpecialistVLMs for retinal OCT, and pathology-specific tools. These constraints ensure that tool selection accounts for both task requirements and input modality.

\begin{table}[!h]
\centering
\caption{\textbf{Tool-modality compatibility matrix.} Checkmarks indicate supported tool-modality combinations included in the instruction-tuning corpus. Text-based tools operate on textual inputs, while image-based tools may also accept textual prompts
(Section~\ref{sup:interface_rules}).}
\label{tab:modality_matrix}
\small
\setlength{\tabcolsep}{5pt}
\renewcommand{\arraystretch}{1.15}
\begin{tabular}{l cccccccc}
\hline\hline
\textbf{Tool} & \textbf{Text} & \textbf{X-ray} & \textbf{CT} & \textbf{MRI} & \textbf{US} & \textbf{Fundus} & \textbf{OCT} & \textbf{Pathology} \\
\hline
PMC-LLaMA~\cite{wu2024pmc}        & \checkmark & & & & & & & \\
RaTE-NER~\cite{zhao2024ratescore}         & \checkmark & & & & & & & \\
ChatCAD-R~\cite{wang2024interactive}        & \checkmark & & & & & & & \\
\hline
LLaVA-Med~\cite{li2023llava}        & & \checkmark & \checkmark & \checkmark & & & & \checkmark \\
BiomedCLIP~\cite{zhang2023biomedclip}       & & \checkmark & \checkmark & \checkmark & & & & \checkmark \\
CONCH~\cite{lu2024visual}            & & & & & & & & \checkmark \\
\hline
DSMIL-C16~\cite{li2021dual}        & & & & & & & & \checkmark \\
DSMIL-TCGA~\cite{li2021dual}       & & & & & & & & \checkmark \\
Grounding DINO~\cite{liu2024grounding}   & & \checkmark & \checkmark & \checkmark & & & & \checkmark \\
MedSAM~\cite{ma2024segment}           & & \checkmark & \checkmark & \checkmark & & & & \checkmark \\
G-Seg            & & \checkmark & \checkmark & \checkmark & & & & \checkmark \\
UltraSAM~\cite{meyer2024ultrasam}         & & & & & \checkmark & & & \\
IterNet~\cite{li2020iternet}          & & & & & & \checkmark & & \\
CellViT~\cite{horst2023cellvit}          & & & & & & & & \checkmark \\
CellSAM~\cite{marks2025cellsam}          & & & & & & & & \checkmark \\
UniGradICON~\cite{tian2024unigradicon}      & & & \checkmark & \checkmark & & & & \\
HealthGPT~\cite{lin2025healthgpt}        & & \checkmark & \checkmark & \checkmark & \checkmark & & & \\
\hline
LLaVA-Rad~\cite{chaves2024towards}        & & \checkmark & \checkmark & \checkmark & \checkmark & & & \\
SpecialistVLMs~\cite{holland2024specialized}   & & & & & & & \checkmark & \\
\hline\hline
\end{tabular}
\end{table}

\subsubsection{Interface Compatibility Rules}
\label{sup:interface_rules}

\paragraph{Tool call format.}
The planner $\mathcal{M}$ requests tool execution by generating a structured action block containing one or more tool calls, each specifying a tool name and named parameters. Multiple calls can be specified within a single action block, allowing the planner to define a complete tool sequence before execution in the multi-tool single-round setting $\mathcal{S}_3$. A response without an action block is treated as the final response and terminates the workflow.

\paragraph{Tool name resolution.}
Tool names are resolved case-insensitively against a central registry. Each tool is registered under its canonical name together with a set of aliases, so that variants such as \texttt{Grounding\_DINO} and \texttt{grounding-dino} both resolve to the same implementation. This makes execution robust to surface variation in the generated tool name without requiring $\mathcal{M}$ to reproduce a single exact string.

\paragraph{Parameter binding.}
Each tool defines canonical parameter names and accepted aliases, which are mapped to the corresponding tool arguments during execution. Text inputs accept
\texttt{text}, \texttt{input\_text}, \texttt{query}, \texttt{input}, \texttt{context}, and \texttt{prompt}; image inputs accept \texttt{image\_path}, \texttt{image}, \texttt{input\_image}, \texttt{img\_path}, and \texttt{input}; registration accepts \texttt{fixed}, \texttt{target}, or \texttt{reference} for the fixed image and \texttt{moving}, \texttt{source}, or \texttt{input} for the moving image. Parameters that are not supplied fall back to the tool defaults reported in Table~\ref{tab:tool_cards}, so a call needs to name only the parameters it overrides.

\paragraph{Image binding.}
User-provided images are handled separately from the language channel. Tool calls reference images through placeholders, which are resolved to the corresponding images at execution time. When an image-based tool is invoked without an explicit image parameter, the current image is bound automatically. For image registration, the first image is assigned as the fixed image and the second as the moving image, consistent with the structural validation requirement that registration requests contain paired CT or MRI images (Section~\ref{sec:dataset}). This mechanism also allows tools to access images provided in earlier dialogue turns.

\paragraph{Output types and error handling.}
Each tool invocation returns a structured record containing a success flag, the resolved tool name, and either an output payload or an error message. Output formats are tool-specific. Generative tools return text, classification tools return predicted labels, grounding tools return bounding boxes with confidence scores, matched phrases, and image dimensions, segmentation tools return run-length-encoded masks and mask counts, and registration tools return warped images and deformation fields. Execution failures, including unavailable services, timeouts, and missing required parameters, are returned as structured observations rather than unhandled exceptions. This allows $\mathcal{M}$ to attempt recovery through a revised tool call or report the failure in its final response.

\paragraph{Interface compatibility.}
Tool chains are validated by checking whether the output type of an upstream tool is compatible with the input requirements of the subsequent tool. Three classes of compatibility are supported. The first is \emph{bounding boxes to masks}, in which Grounding DINO provides bounding boxes used as prompts by MedSAM through the composite G-Seg tool. The second is \emph{image to image or text}, in which an image generated by HealthGPT or UniGradICON is passed to a downstream image-analysis tool. The third is \emph{text to text}, in which generated reports are provided to RaTE-NER for entity extraction or to other text-based tools as contextual input. These compatibility rules are applied during the construction of $\mathcal{S}_2$ and $\mathcal{S}_3$.

\paragraph{Iteration budget.}
Execution alternates between planner generation and tool execution for a maximum of $T_{\max}$ interaction rounds. Workflows that do not produce a final response within this limit are terminated and labeled as truncated. Truncated workflows are counted as failures in the tool orchestration evaluation (Section~\ref{sec:evaluation}).

\subsection{Dataset}
\label{sup:dataset}

This section details the instruction-tuning corpus $\mathcal{S}$ described in Section~\ref{sec:dataset}. We provide the source datasets and sample counts for each specialist tool (Section~\ref{sup:data_sources}), the tool-chain configurations used in the multi-tool settings (Section~\ref{sup:tool_chains}), the instruction record schema (Section~\ref{sup:instruction_schema}), and the training and evaluation split protocol (Section~\ref{sup:split_protocol}).

\subsubsection{Data Sources}
\label{sup:data_sources}

Table~\ref{tab:dataset_details} summarizes the source datasets used to construct the single-tool subset $\mathcal{S}_1$, together with their associated specialist tools, input modalities, and resulting instruction sample counts. Each tool-specific subset is constructed from datasets that match the tool's supported tasks and input modalities, as defined by the tool-modality compatibility matrix Table~\ref{tab:modality_matrix}.

\begin{table}[!h]
\centering
\caption{\textbf{Source datasets for instruction-tuning data construction.} The table lists the source datasets, associated specialist tools, input modalities, and number of instruction samples used to construct $\mathcal{S}_1$. Dataset links are provided in footnotes.}
\label{tab:dataset_details}
\small
\setlength{\tabcolsep}{5pt}
\renewcommand{\arraystretch}{1.15}
\begin{tabular}{p{3.5cm} p{3cm} p{3.2cm} p{3.5cm} p{1.4cm}}
\hline\hline
\textbf{Task} & \textbf{Tool} & \textbf{Modality} & \textbf{Data Source} & \textbf{\#Samples} \\
\hline
QA / Rationale Generation & \href{https://arxiv.org/pdf/2304.14454}{PMC-LLaMA} & Text & \href{https://huggingface.co/datasets/axiong/pmc_llama_instructions}{PMC-LLaMA}, \href{https://huggingface.co/datasets/qiaojin/PubMedQA}{PubMedQA}, \href{https://huggingface.co/datasets/GBaker/MedQA-USMLE-4-options}{MedQA-USMLE} & 9,000 \\
Named Entity Recognition & \href{https://aclanthology.org/2024.emnlp-main.836.pdf}{RaTE-NER} & Text & \href{https://huggingface.co/datasets/Angelakeke/RaTE-NER}{RaTE-NER}, \href{https://github.com/vgupta123/sumpubmed/tree/master}{SumPubMed} & 4,500 \\
Retrieval-Augmented Generation & \href{https://ieeexplore.ieee.org/document/10522762}{ChatCAD-R} & Text & \href{https://www.merckmanuals.com/}{Merck Manual} & 2,528 \\
Visual Question Answering & \href{https://proceedings.neurips.cc/paper_files/paper/2023/file/5abcdf8ecdcacba028c6662789194572-Paper-Datasets_and_Benchmarks.pdf}{LLaVA-Med} & MRI, CT, X-ray, Histology, Gross & \href{https://github.com/microsoft/LLaVA-Med}{PMC article
60K-IM} & 16,687 \\
Classification & \href{https://arxiv.org/abs/2307.12914}{CONCH} & Pathology & \href{https://zenodo.org/records/1214456?utm_source=chatgpt.com}{CRC100k} & 3,600 \\
Classification & \href{https://arxiv.org/pdf/2303.00915}{BiomedCLIP} & MRI, CT, X-ray, Histology, Gross & \href{https://github.com/microsoft/LLaVA-Med}{PMC article
60K-IM} & 7,923 \\
Classification & \href{https://arxiv.org/abs/2011.08939}{DSMIL-TCGA} & Pathology (Lung Tumor) & \href{https://portal.gdc.cancer.gov/?utm_source=chatgpt.com}{TCGA} & 1,800 \\
Classification & \href{https://arxiv.org/abs/2011.08939}{DSMIL-C16} & Pathology (Breast Tumor) & \href{https://camelyon16.grand-challenge.org/Data/?utm_source=chatgpt.com}{Camelyon16} & 1,800 \\
Grounding                  & \href{https://arxiv.org/pdf/2303.05499}{Grounding DINO} & MRI, CT, X-ray, Histology & \href{https://github.com/HiLab-git/WORD}{WORD}, \href{https://flare.grand-challenge.org/FLARE21}{FLARE2021}, \href{http://braintumorsegmentation.org/}{BRATS}, \href{https://data.lhncbc.nlm.nih.gov/public/Tuberculosis-Chest-X-ray-Datasets/Montgomery-County-CXR-Set/MontgomerySet/index.html}{Montgomery County X-ray Set}, \href{https://physionet.org/content/vindr-cxr/1.0.0/}{VinDr-CXR}, \href{https://neurips22-cellseg.grand-challenge.org/}{CellSeg} & 4,207 \\
G-Seg                      & Grounding DINO + MedSAM & MRI, CT, X-ray, Histology &  & 2,303 \\
Segmentation               & \href{https://www.nature.com/articles/s41467-024-44824-z.pdf}{MedSAM} & MRI, CT, X-ray, Histology, Gross & \href{https://github.com/HiLab-git/WORD}{WORD}, \href{https://flare.grand-challenge.org/FLARE21}{FLARE2021}, \href{http://braintumorsegmentation.org/}{BRATS}, \href{https://data.lhncbc.nlm.nih.gov/public/Tuberculosis-Chest-X-ray-Datasets/Montgomery-County-CXR-Set/MontgomerySet/index.html}{Montgomery County X-ray Set}, \href{https://physionet.org/content/vindr-cxr/1.0.0/}{VinDr-CXR}, \href{https://neurips22-cellseg.grand-challenge.org/}{CellSeg} & 5,765 \\
Segmentation & \href{https://arxiv.org/pdf/2411.16222}{UltraSAM} & US & \href{https://www.kaggle.com/datasets/ignaciorlando/ussimandsegm}{US-SimSeg}, \href{https://zenodo.org/records/7272660}{Zenodo-US}, \href{https://www.cancerimagingarchive.net/collection/breast-lesions-usg/}{TCIA-USG} & 5,000 \\
Segmentation & \href{https://www.liangzhili.com/paper/li-2020-iternet.pdf}{IterNet} & Retina (Fundus) & \href{https://cecas.clemson.edu/~ahoover/stare/}{STARE}, \href{https://www.kaggle.com/datasets/deathtrooper/multichannel-glaucoma-benchmark-dataset}{Glaucoma-Benchmark}, \href{https://www.kaggle.com/datasets/ascanipek/eyepacs-aptos-messidor-diabetic-retinopathy}{Diabetic-Retinopathy} & 4,500 \\
Segmentation & \href{https://arxiv.org/abs/2306.15350}{CellViT} & Pathology (WSI Cell) & & 1,800 \\
Segmentation & \href{https://arxiv.org/abs/2311.11004}{CellSAM} & Pathology (2D Cell) & \href{https://zenodo.org/records/1214456?utm_source=chatgpt.com}{CRC100k} & 1,800 \\
Registration & \href{https://arxiv.org/abs/2403.05780}{UniGradICON} & CT, MRI & \href{https://med.emory.edu/departments/radiation-oncology/research-laboratories/deformable-image-registration/downloads-and-reference-data/index.html}{EmoryReg}, \href{https://learn2reg.grand-challenge.org/Datasets/}{Learn2Reg}, \href{https://db.humanconnectome.org/data/projects/HCP_1200}{HCP} & 5,000 \\
Reconstruction & \href{https://arxiv.org/abs/2502.09838}{HealthGPT} & X-ray, MRI, CT, US & \href{https://www.kaggle.com/datasets/nih-chest-xrays/data/data}{NIH ChestX-ray}, \href{https://fastmri.med.nyu.edu/}{FastMRI} & 6,147 \\
Super Resolution & \href{https://arxiv.org/abs/2502.09838}{HealthGPT} & X-ray, MRI, CT, US & \href{https://www.kaggle.com/datasets/nih-chest-xrays/data/data}{NIH ChestX-ray} & 4,593 \\
Report Generation & \href{https://www.nature.com/articles/s41467-025-58344-x}{LLaVA-Rad} & X-ray, MRI, CT, US & \href{https://physionet.org/content/llava-rad-mimic-cxr-annotation/1.0.0/}{LLaVA-Rad CXR}, \href{https://stanfordaimi.azurewebsites.net/datasets}{Stanford-AIMI}, \href{https://openi.nlm.nih.gov/faq}{OpenI} & 4,500 \\
Report Generation & \href{https://arxiv.org/abs/2407.08410}{SpecialistVLMs} & Retina (OCT) & \href{https://github.com/Jhhuangkay/DeepOpht-Medical-Report-Generation-for-Retinal-Images-via-Deep-Models-and-Visual-Explanation?tab=readme-ov-file}{DeepOpht}(non-public) & 3,602 \\
\hline
\textbf{Total} & & & & \textbf{97,055} \\
\hline\hline
\end{tabular}
\end{table}

\subsubsection{Request and Response Templates}
\label{sup:templates}

Instruction dialogues for newly integrated tools are constructed using predefined natural-language templates rather than per-sample language model generation. For each task, banks of user-request phrasings and assistant-response templates are generated in a single offline pass using GPT-4~\cite{achiam2023gpt} and subsequently fixed. Single-tool samples are instantiated by pairing a randomly selected request phrasing with a case from the corresponding source benchmark. The target assistant turn combines a structured tool call with a response template populated using the reference output of that case. Multi-tool dialogues are assembled from these samples according to predefined workflow templates, with transitions drawn from a fixed bank of 96 connective phrases, 28 greetings, and 43 follow-up turns. The size of each task-specific template bank is reported in Table~\ref{tab:template_banks}.

All medical content in the corpus is taken from the source benchmarks. Images, questions, labels, reports, entity spans, and segmentation masks are those released with the corresponding datasets, and all tool calls are emitted from the tool registry schema of Section~\ref{sup:tool_cards}. GPT-4 contributes only the natural-language scaffolding of requests and responses, and no clinical content is model-generated.

\begin{table}[!h]
\centering
\caption{\textbf{Size of the request and response template inventories.} For each task we report the number of distinct user-request phrasings and assistant-response templates from which samples are drawn. Inventories are frozen before corpus construction and are released with the generation scripts.}
\label{tab:template_banks}
\small
\setlength{\tabcolsep}{6pt}
\renewcommand{\arraystretch}{1.15}
\begin{tabular}{l l r r}
\hline\hline
\textbf{Task} & \textbf{Tool} & \textbf{Request} & \textbf{Response} \\
\hline
QA \& rationale generation & PMC-LLaMA~\cite{wu2024pmc} & 48 & 40 \\
Named entity recognition & RaTE-NER~\cite{zhao2024ratescore} & 40 & 40 \\
Report generation (radiology) & LLaVA-Rad~\cite{chaves2024towards} & 81 & 40 \\
Report generation (retina OCT) & SpecialistVLMs~\cite{holland2024specialized} & 90 & 41 \\
Segmentation (ultrasound) & UltraSAM~\cite{meyer2024ultrasam} & 49 & 41 \\
Segmentation (fundus) & IterNet~\cite{li2020iternet} & 65 & --- \\
Registration & UniGradICON~\cite{tian2024unigradicon} & 100 & 25 \\
Reconstruction & HealthGPT~\cite{lin2025healthgpt} & 30 & 20 \\
Super-resolution & HealthGPT~\cite{lin2025healthgpt} & 42 & 41 \\
\hline\hline
\end{tabular}
\end{table}

\subsubsection{Data Provenance}
\label{sup:provenance}

The instruction-tuning corpus combines newly constructed data with instruction data inherited from MMedAgent~\cite{li2024mmedagent}. Data for the newly integrated specialist tools are generated using the template-based procedure described in Section~\ref{sup:templates}. For the inherited tools, comprising five specialist models and the composite grounding-segmentation endpoint, we retain the original instruction data generated using GPT-4 and source-dataset annotations following the procedure described in MMedAgent. In addition to the automated validation procedures described in Section~\ref{sec:dataset}, the authors manually inspect 50 to 100 generated samples per specialist tool during dataset development to assess the consistency of user requests, selected tools, and target responses with their intended tasks and input modalities.

\subsubsection{Tool Chains}
\label{sup:tool_chains}

The two multi-tool settings differ in how tool sequences are planned and executed. In the multi-tool single-round setting $\mathcal{S}_3$, the planner $\mathcal{M}$ selects and orders the complete tool sequence from a single request before any specialist output becomes available. In the multi-tool multi-round setting $\mathcal{S}_2$, tools are selected across dialogue turns, allowing subsequent decisions to incorporate preceding dialogue context and available tool outputs. Candidate chains in both settings are constructed according to the modality and interface compatibility rules described in Section~\ref{sup:interface_rules}.

\paragraph{Multi-tool single-round chains.}
Table~\ref{tab:chains_s3} presents the predefined tool chains used in $\mathcal{S}_3$. These chains comprise two or three specialist operations combining medical image analysis, question answering, and report generation. The complete tool sequence is specified before execution, with intermediate outputs used to support downstream operations where the tool interfaces permit.

\begin{table}[!h]
\centering
\caption{\textbf{Tool chains in the multi-tool single-round subset $\mathcal{S}_3$.} Each sequence is selected and ordered from a single user request before any specialist tool is executed.}
\label{tab:chains_s3}
\small
\setlength{\tabcolsep}{6pt}
\renewcommand{\arraystretch}{1.2}
\begin{tabular}{l l}
\hline\hline
\textbf{Tool chain} & \textbf{Workflow} \\
\hline
BiomedCLIP $\rightarrow$ PMC-LLaMA $\rightarrow$ LLaVA-Rad & Identify the study, interpret it, then report \\
LLaVA-Rad $\rightarrow$ PMC-LLaMA & Generate a report, then answer a question about it \\
MedSAM $\rightarrow$ LLaVA-Rad & Segment a structure, then report on it \\
UltraSAM $\rightarrow$ LLaVA-Rad & Segment an ultrasound study, then report on it \\
IterNet $\rightarrow$ LLaVA-Rad & Extract retinal vessels, then report on them \\
\hline\hline
\end{tabular}
\end{table}

\paragraph{Multi-tool multi-round chains.}
Dialogues in $\mathcal{S}_2$ contain two to four tool invocations distributed across multiple turns. The subset comprises two complementary interaction patterns, distinguished by whether subsequent requests depend on earlier tool outputs.

The first pattern consists of 33{,}749 dependent dialogues, each containing two tool invocations. An image-based specialist first analyzes the medical input, and a text-based tool subsequently processes the returned findings. The upstream tool is selected from 15 visual understanding and image-processing specialists, while the downstream tool is selected from RaTE-NER, PMC-LLaMA, or ChatCAD-R, yielding 45 distinct tool chains. This pattern trains $\mathcal{M}$ to incorporate intermediate specialist outputs into subsequent tool-selection decisions.

The second pattern comprises 38{,}230 dialogues in which each round introduces a new request with its own image or text, so the invocations are independent rather than sequentially dependent. These dialogues contain two, three, or four invocations (13{,}177, 12{,}813, and 12{,}240 samples respectively) and span a much larger combination space, covering 342, 5{,}195, and 11{,}460 distinct chains. Because the tools are sampled from the set of modality-compatible candidates rather than from a curated workflow list, this pattern exposes $\mathcal{M}$ to combinations that no predefined chain inventory would contain, and trains it to re-select a tool for each request instead of continuing to apply the tool used in the previous round.

Representative chains from both patterns are provided in Table~\ref{tab:chains_s2}.

\begin{table}[!h]
\centering
\caption{\textbf{Representative tool chains in the multi-tool multi-round subset $\mathcal{S}_2$.} The first block shows dependent workflows in which a downstream text-based tool processes the output of an upstream specialist. The remaining blocks show independent tool sequences in which each turn introduces a separate request. These sequences illustrate the diversity of tool combinations rather than dependencies between successive invocations.}
\label{tab:chains_s2}
\footnotesize
\setlength{\tabcolsep}{4pt}
\renewcommand{\arraystretch}{1.2}
\begin{tabularx}{\textwidth}{@{}>{\hsize=1.18\hsize\raggedright\arraybackslash}X >{\hsize=0.82\hsize\raggedright\arraybackslash}X@{}}
\hline\hline
\textbf{Tool chain} & \textbf{Workflow} \\
\hline
\multicolumn{2}{l}{\textit{Successive requests on the same study (two invocations; 45 distinct chains)}} \\
BiomedCLIP $\rightarrow$ PMC-LLaMA & Classify the study, then answer a question about it \\
BiomedCLIP $\rightarrow$ RaTE-NER & Classify the study, then extract typed entities \\
BiomedCLIP $\rightarrow$ ChatCAD-R & Classify the study, then retrieve supporting knowledge \\
MedSAM $\rightarrow$ PMC-LLaMA & Segment a structure, then answer a question about it \\
LLaVA-Med $\rightarrow$ PMC-LLaMA & Answer a visual question, then reason over the answer \\
LLaVA-Med $\rightarrow$ ChatCAD-R & Answer a visual question, then retrieve supporting knowledge \\
LLaVA-Rad $\rightarrow$ RaTE-NER & Generate a report, then extract typed entities \\
LLaVA-Rad $\rightarrow$ ChatCAD-R & Generate a report, then retrieve supporting knowledge \\
HealthGPT $\rightarrow$ LLava-Rad & Enhance an image, then generate a report \\
G-Seg $\rightarrow$ PMC-LLaMA & Localize and segment, then answer a question about it \\
\hline
\multicolumn{2}{l}{\textit{Independent requests, two invocations (342 distinct chains)}} \\
UltraSAM $\rightarrow$ HealthGPT & Ultrasound segmentation, then image enhancement \\
LLaVA-Med $\rightarrow$ Grounding DINO & Visual question answering, then localization \\
BiomedCLIP $\rightarrow$ CellViT & Image classification, then WSI cell segmentation \\
MedSAM $\rightarrow$ DSMIL-C16 & General segmentation, then tumor detection \\
SpecialistVLMs $\rightarrow$ Grounding DINO & OCT reporting, then localization \\
\hline
\multicolumn{2}{l}{\textit{Independent requests, three invocations (5{,}195 distinct chains)}} \\
BiomedCLIP $\rightarrow$ UniGradICON $\rightarrow$ IterNet & Classification, registration, vessel segmentation \\
CONCH $\rightarrow$ UltraSAM $\rightarrow$ LLaVA-Rad & Pathology classification, ultrasound segmentation, reporting \\
DSMIL-TCGA $\rightarrow$ HealthGPT $\rightarrow$ Grounding DINO & Tumor subtyping, enhancement, localization \\
G-Seg $\rightarrow$ CONCH $\rightarrow$ UniGradICON & Grounding-segmentation, pathology classification, registration \\
PMC-LLaMA $\rightarrow$ RaTE-NER $\rightarrow$ DSMIL-TCGA & Question answering, entity extraction, tumor subtyping \\
\hline
\multicolumn{2}{l}{\textit{Independent requests, four invocations (11{,}460 distinct chains)}} \\
DSMIL-C16 $\rightarrow$ LLaVA-Med $\rightarrow$ UltraSAM $\rightarrow$ LLaVA-Rad & Tumor detection, VQA, US segmentation, reporting \\
UniGradICON $\rightarrow$ CellSAM $\rightarrow$ MedSAM $\rightarrow$ LLaVA-Rad & Registration, cell and general segmentation, reporting \\
LLaVA-Med $\rightarrow$ LLaVA-Rad $\rightarrow$ PMC-LLaMA $\rightarrow$ CellViT & VQA, reporting, question answering, cell segmentation \\
G-Seg $\rightarrow$ DSMIL-C16 $\rightarrow$ UniGradICON $\rightarrow$ DSMIL-TCGA & Grounding-segmentation, tumor detection, registration, subtyping \\
LLaVA-Med $\rightarrow$ HealthGPT $\rightarrow$ SpecialistVLMs $\rightarrow$ UniGradICON & VQA, enhancement, OCT reporting, registration \\
\hline\hline
\end{tabularx}
\end{table}

\subsubsection{Instruction schema}
\label{sup:instruction_schema}

Each training or evaluation sample is stored as a JSONL record containing an ordered dialogue between the user and the planner $\mathcal{M}$, together with any associated specialist outputs. During training, tool outputs are inserted into the dialogue as subsequent messages following the corresponding tool calls. During inference, these outputs are obtained by executing the selected specialist tools and returned to the planner as additional context.

Dialogues follow the ShareGPT(\url{https://sharegpt.com/}) and LLaVA~\cite{liu2023visual} conversation format used by the vision-language training pipeline. Planner messages are labeled \texttt{"from": "gpt"}, while both user messages and specialist outputs are labeled \texttt{"from": "human"}. These labels represent serialization conventions rather than the actual source of each message. User messages contain clinical requests and optional image placeholders, while tool-return messages contain structured or textual outputs prefixed by \texttt{<ToolName> output: ...}. Specialist tools do not have separate role labels in the stored dialogue.

\begin{center}
\small
\begin{tabular}{lll}
\hline
JSON tag & System role & Message content \\
\hline
\texttt{from: "human"} & User or tool return & User query or \texttt{<Tool> output: ...} \\
\texttt{from: "gpt"} & Planner $\mathcal{M}$ & Tool selection or final answer \\
\hline
\end{tabular}
\end{center}

The complete record structure is illustrated in Table~\ref{tab:schema_record}. The \texttt{conversations} field stores the ordered dialogue turns, while the \texttt{image} field contains paths corresponding to \texttt{<image>} placeholders in the messages. The \texttt{meta} field is included only for multi-tool samples.

\begin{table}[!h]
\caption{\textbf{JSONL record layout for instruction samples.}}
\label{tab:schema_record}
\begin{lstlisting}[
  basicstyle=\ttfamily\footnotesize,
  breaklines=true,
  breakindent=1.5em,
  columns=fullflexible,
  keepspaces=true,
  frame=single,
  framerule=0.4pt,
  framesep=5pt,
  xleftmargin=0pt,
  xrightmargin=0pt,
  aboveskip=2pt,
  belowskip=2pt,
  linewidth=\linewidth
]
{
  "image": <path | [path, ...]>,
  "conversations": [ ... ],
  "meta": {                 // multi-tool samples only
    "tools": ["ToolA", "ToolB"],
    "modality": "CT"        // S_3 only
  }
}
\end{lstlisting}
\end{table}

A typical tool-use cycle begins with a user request (\texttt{human}), followed by a planner response (\texttt{gpt}) specifying one or more specialist tools. The corresponding tool outputs are then inserted as \texttt{human}-tagged messages prefixed with \texttt{<ToolName> output: ...}. The planner subsequently generates a final response or continues tool selection when additional steps are required. This message structure supports both single-round tool planning and multi-round interactions.

The three dialogue settings described in Section~\ref{sec:dataset} use this schema with different interaction structures. In $\mathcal{S}_1$ (single-tool multi-round), each dialogue contains one specialist invocation and may include additional conversational turns, with the \texttt{meta} field omitted. In $\mathcal{S}_2$ (multi-tool multi-round), multiple tool-use cycles are distributed across dialogue turns, allowing later tool selections to incorporate preceding context and returned specialist outputs. The \texttt{meta.tools} field records the tools in execution order. In $\mathcal{S}_3$ (multi-tool single-round), the planner specifies a complete sequence of two to three tools from a single user request before any specialist output becomes available. The resulting tool outputs are concatenated into a single \texttt{human}-tagged message, while \texttt{meta} records the tool sequence and input modality.

Table~\ref{tab:schema_example} illustrates the record structure using an $\mathcal{S}_1$ example. Inline comments are included for readability and are not part of the stored JSONL record. In $\mathcal{S}_2$, the structure extends to multiple tool-use cycles, each associated with its corresponding inputs and specialist outputs. In $\mathcal{S}_3$, the planner instead generates the complete tool sequence in a single response, and the tool returns are combined into one message. For example, a sequence may comprise BiomedCLIP $\rightarrow$ PMC-LLaMA $\rightarrow$ LLaVA-Rad. Training and evaluation samples follow the same record schema, with their membership determined by the corresponding data split.

\begin{table}[!h]
\caption{\textbf{Example $\mathcal{S}_1$ JSONL record} (BiomedCLIP classification).}
\label{tab:schema_example}
\begin{lstlisting}[
  basicstyle=\ttfamily\footnotesize,
  breaklines=true,
  breakindent=1.5em,
  columns=fullflexible,
  keepspaces=true,
  frame=single,
  framerule=0.4pt,
  framesep=5pt,
  xleftmargin=0pt,
  xrightmargin=0pt,
  aboveskip=2pt,
  belowskip=2pt,
  linewidth=\linewidth
]
{
  "image": "tool_image/train/33431707_fig3.jpg",
  "conversations": [
    // 1. User request
    {"from": "human",
     "value":
       "Can you identify the histopathology type?\n<image>"},
    // 2. Planner selects a specialist
    {"from": "gpt",
     "value": "I will use the BiomedCLIP model."},
    // 3. Specialist return (tagged "human")
    {"from": "human",
     "value":
       "BiomedCLIP output: immunohistochemistry histopathology\n\nAnswer my first question: ..."},
    // 4. Planner synthesizes the final answer
    {"from": "gpt",
     "value":
       "Based on BiomedCLIP, this is an immunohistochemistry histopathology image."}
  ]
}
\end{lstlisting}
\end{table}

\begin{figure}[htbp]
\centering
\includegraphics[width=\linewidth]{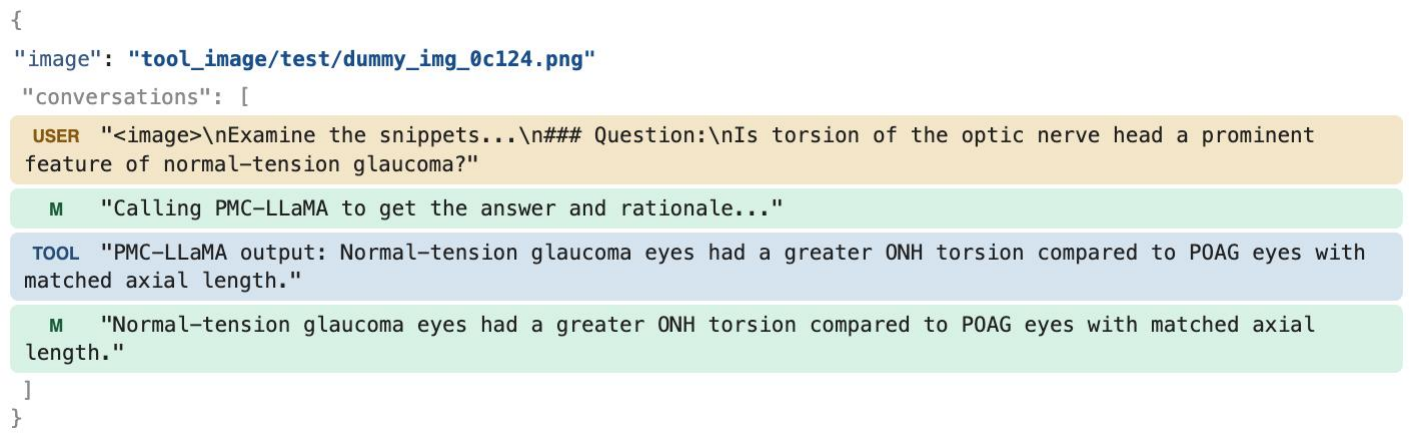}
\caption{PMC-LLaMA (QA~/~rationale generation).}
\label{fig:schema_pmc_llama}
\end{figure}

\begin{figure}[htbp]
\centering
\includegraphics[width=\linewidth]{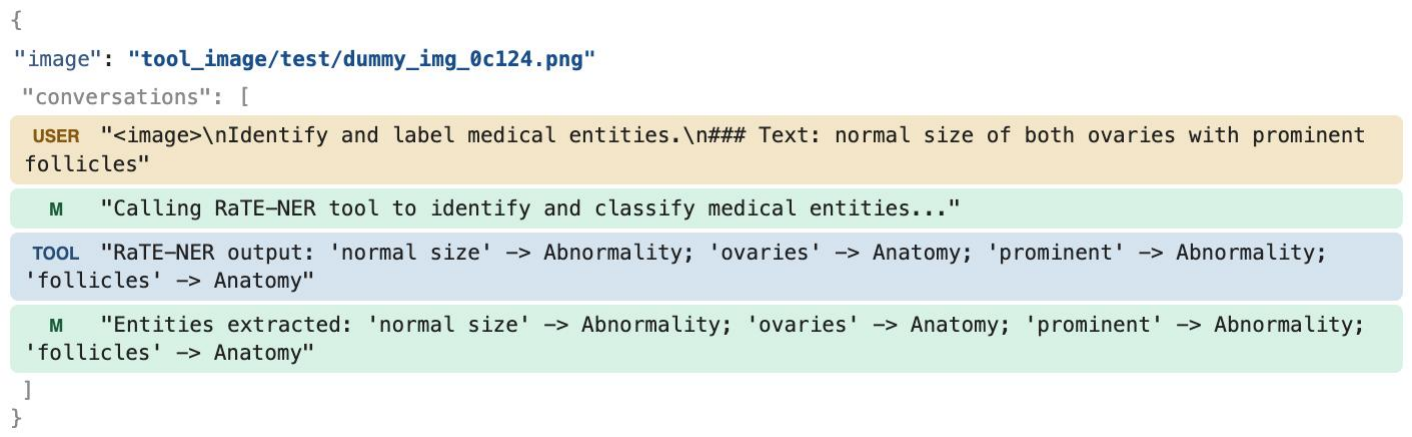}
\caption{RaTE-NER (named entity recognition).}
\label{fig:schema_rate_ner}
\end{figure}

\begin{figure}[htbp]
\centering
\includegraphics[width=\linewidth]{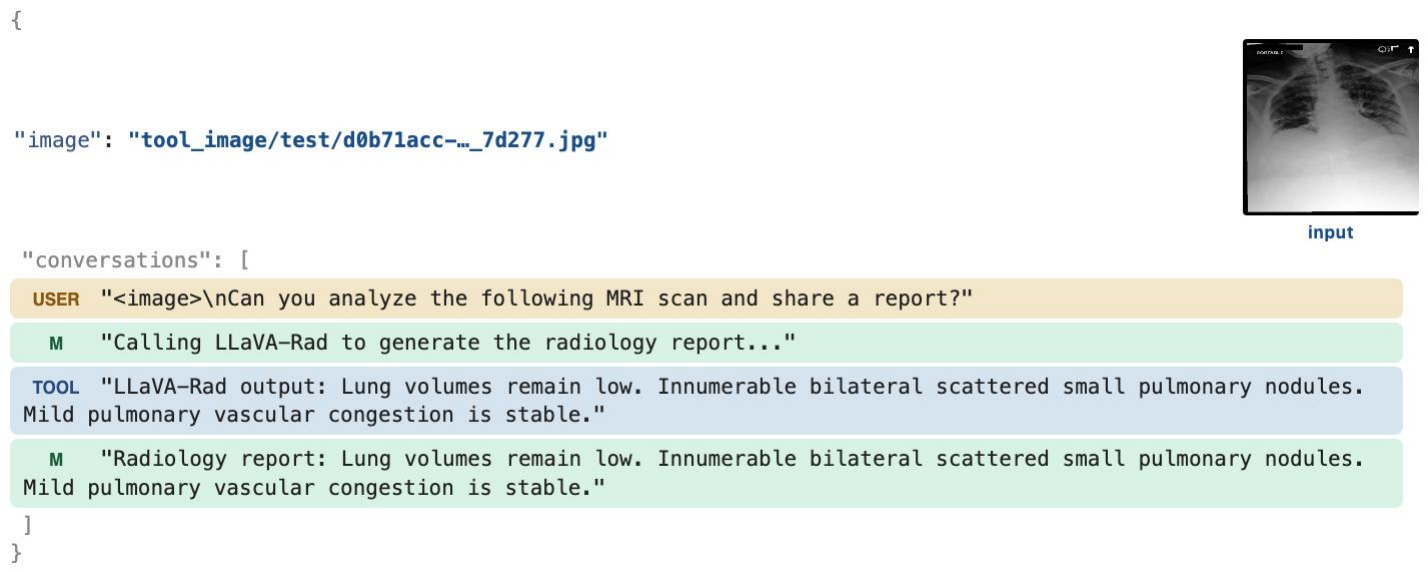}
\caption{LLaVA-Rad (radiology report generation).}
\label{fig:schema_llava_rad}
\end{figure}

\begin{figure}[htbp]
\centering
\includegraphics[width=\linewidth]{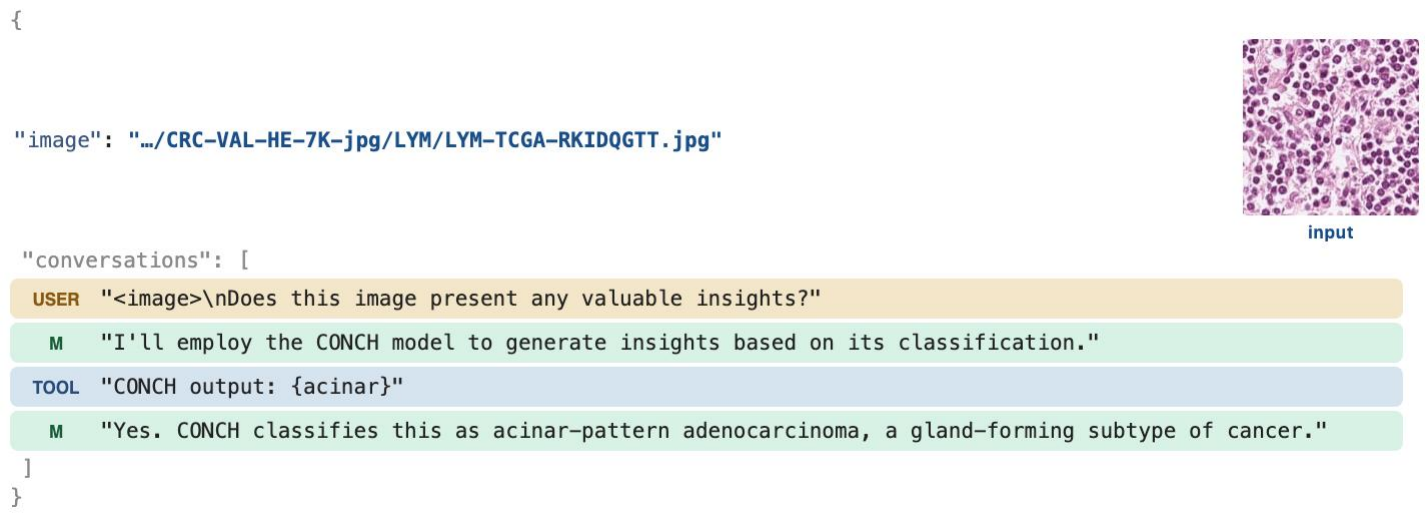}
\caption{CONCH (pathology classification).}
\label{fig:schema_conch}
\end{figure}

\begin{figure}[htbp]
\centering
\includegraphics[width=\linewidth]{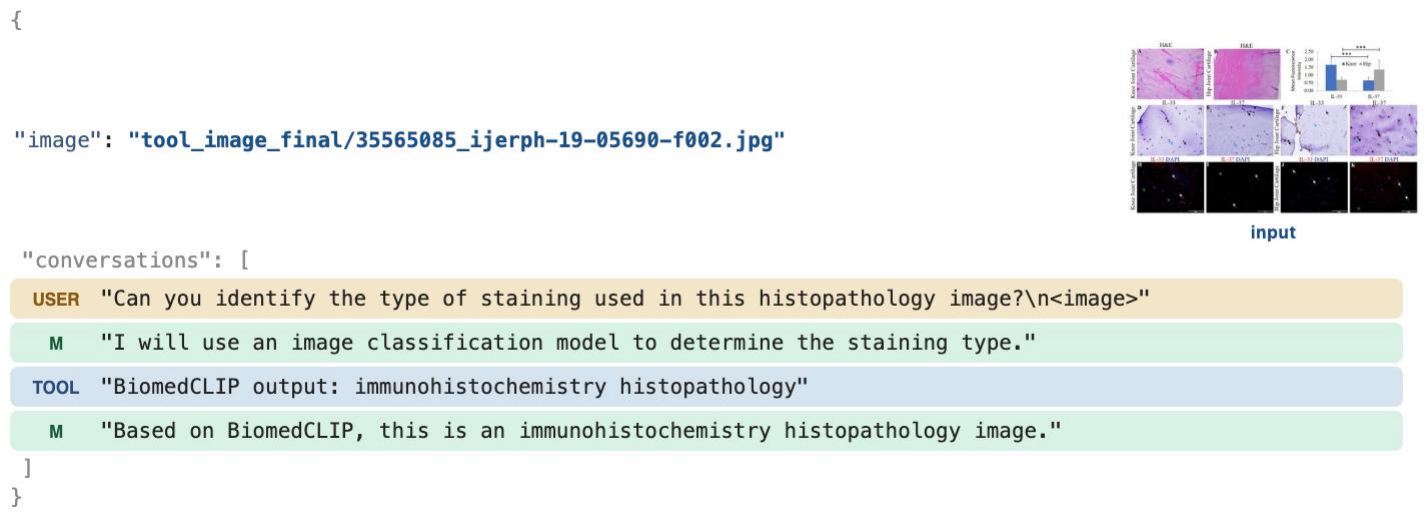}
\caption{BiomedCLIP (multimodal classification).}
\label{fig:schema_biomedclip}
\end{figure}

\begin{figure}[htbp]
\centering
\includegraphics[width=\linewidth]{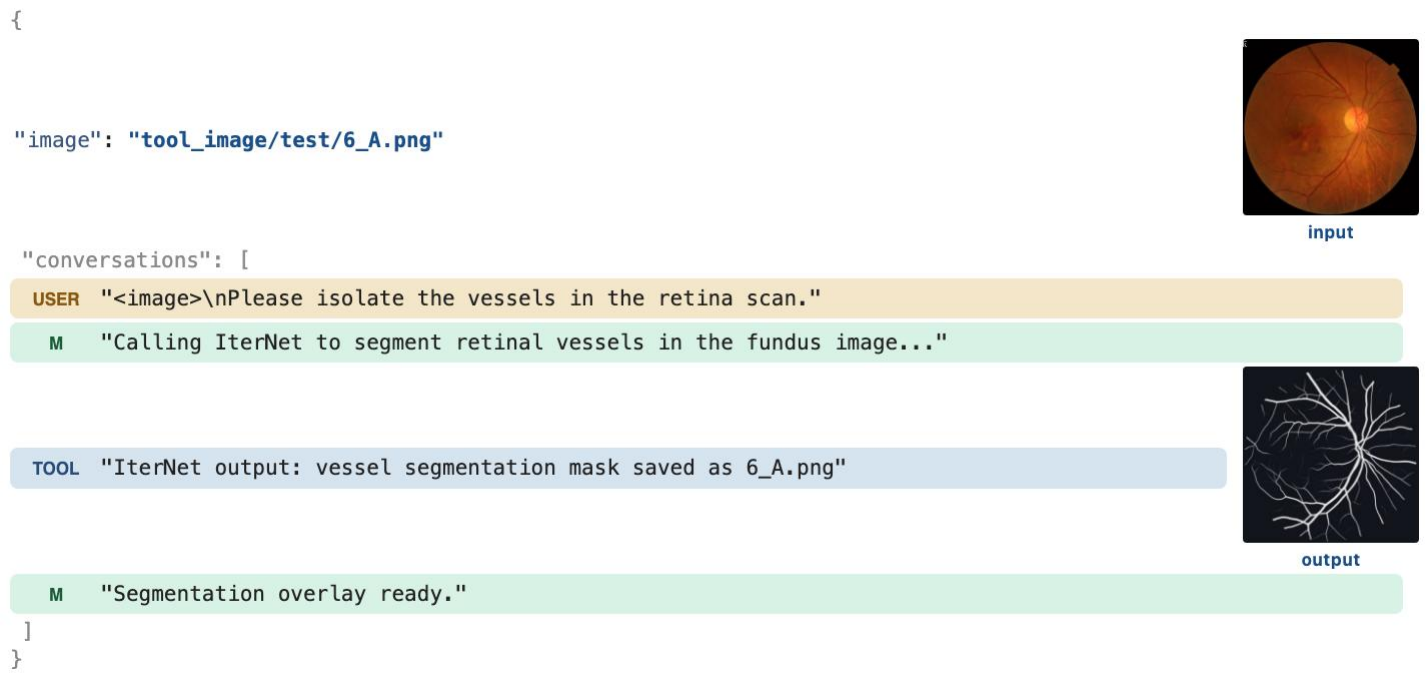}
\caption{IterNet (fundus vessel segmentation).}
\label{fig:schema_iternet}
\end{figure}

\begin{figure}[htbp]
\centering
\includegraphics[width=\linewidth]{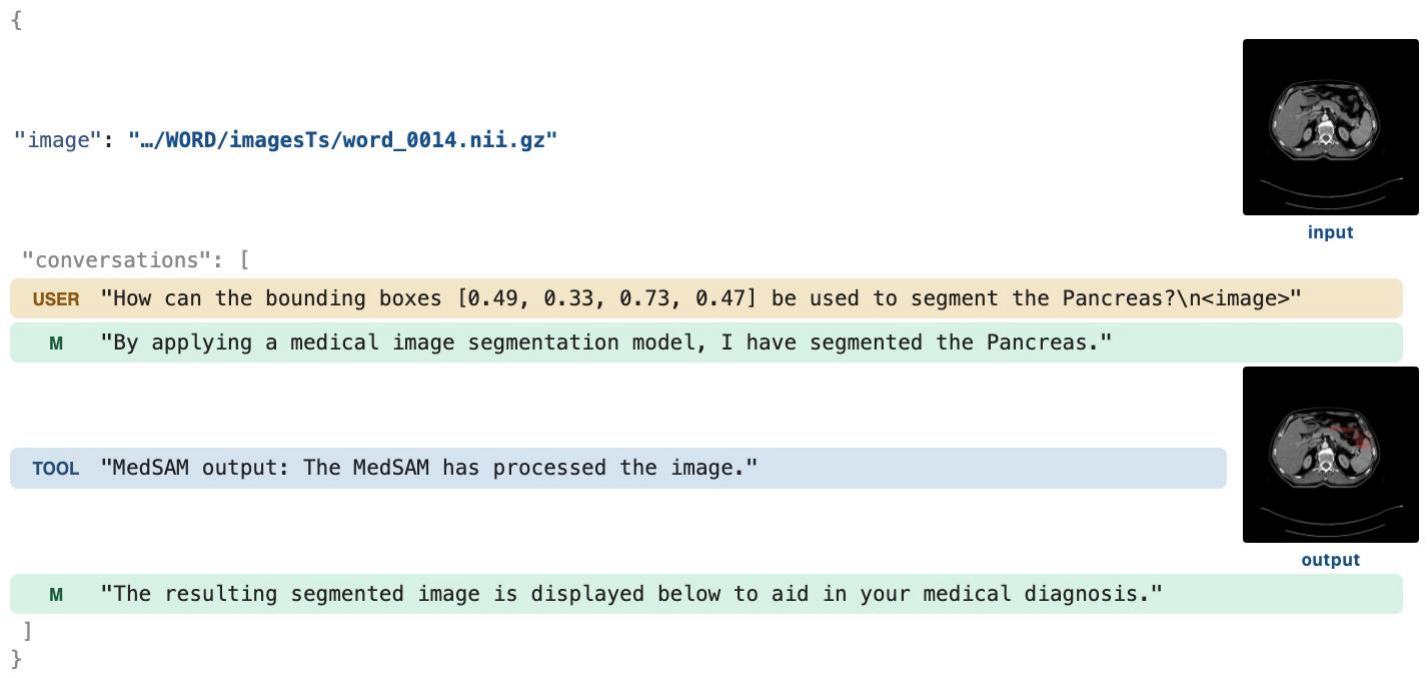}
\caption{MedSAM (general medical image segmentation).}
\label{fig:schema_medsam}
\end{figure}

\begin{figure}[htbp]
\centering
\includegraphics[width=\linewidth]{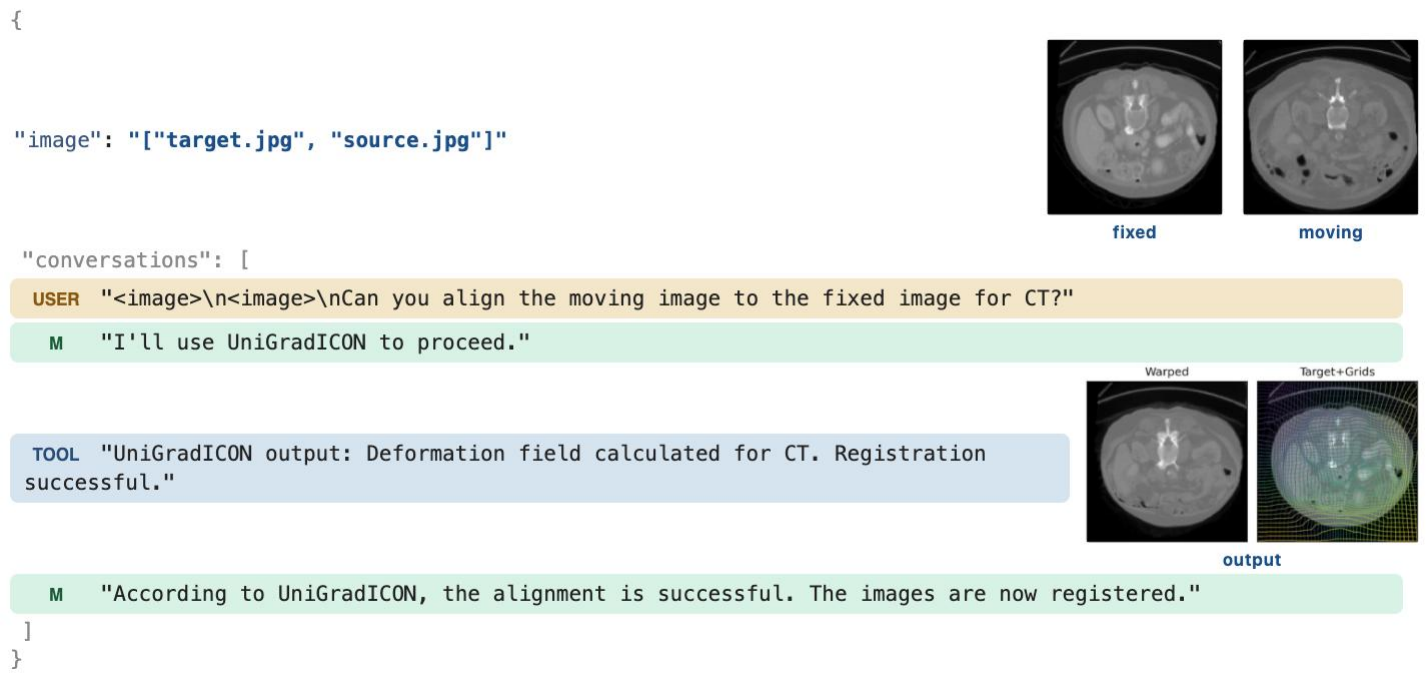}
\caption{UniGradICON (image registration; multiple input images).}
\label{fig:schema_unigradicon}
\end{figure}

Figures~\ref{fig:schema_pmc_llama}-\ref{fig:schema_unigradicon} illustrate representative held-out JSONL records for the seven specialist tools evaluated end-to-end (Section~\ref{sec:end-to-end}), along with a UniGradICON example in which the \texttt{image} field contains multiple image paths. User requests, planner responses, and specialist outputs are highlighted in gold, green, and blue, respectively. Image thumbnails are displayed alongside their corresponding JSON fields. For text-only tools (PMC-LLaMA and RaTE-NER), the \texttt{image} field contains a non-clinical placeholder path, and no thumbnail is displayed.

\subsubsection{Dataset Split Protocol}
\label{sup:split_protocol}

Training and evaluation data are partitioned at the dialogue level, with approximately 10\% of each dialogue setting reserved for held-out evaluation. The resulting held-out pools contain 9{,}128 single-tool multi-round dialogues, 10{,}000 multi-tool multi-round dialogues, and 2{,}430 multi-tool single-round dialogues. Splitting each setting separately maintains its representation in the evaluation data, while dialogue-level partitioning prevents complete dialogues from appearing in both training and held-out sets.

Evaluation cases are sampled from these held-out pools rather than using the entire pools (Section~\ref{sec:evaluation}). Within each setting, cases are selected by uniform random sampling without additional task- or tool-level stratification. Samples exceeding the 8{,}192-token context limit after rendering are excluded to ensure compatibility with the training context window.

We additionally examine image and request overlap between the training and held-out partitions of the multi-tool multi-round subset. The partitions contain 71{,}979 and 7{,}985 dialogues, respectively, comprising 42{,}958 and 11{,}352 distinct images. No source image or image-request pair is shared between these partitions. In contrast, request phrasing overlaps by construction because user requests are sampled from fixed task-specific template inventories (Section~\ref{sup:templates}). Among the 3{,}657 distinct requests in the held-out partition, 67.0\% also appear verbatim in training. Thus, the verified separation in this subset evaluates orchestration on unseen source images, rather than testing generalization to entirely unseen linguistic formulations.

\subsection{Training Details}
\label{sup:training}

This section details the training configuration of the planner $\mathcal{M}$, summarized in Section~\ref{sec:implementation}. The complete set of hyperparameters is provided in Table~\ref{tab:training_config}.

\paragraph{Initialization and trainable parameters.}
The planner $\mathcal{M}$ is initialized from the pretrained Qwen2.5-VL-7B-Instruct checkpoint~\cite{qwen2.5-VL}. The vision encoder remains frozen (Section~\ref{sec:backbone}), while the vision-language projector and language decoder, including the output projection layer, are fully fine-tuned. No parameter-efficient adaptation methods, such as low-rank adaptation, are applied.

\paragraph{Training data and objective.}
Training uses the combined instruction-tuning corpus $\mathcal{S} = \mathcal{S}_1 \cup \mathcal{S}_2 \cup \mathcal{S}_3$ (Section~\ref{sec:dataset}), with each sample included once and no subset reweighting or subsampling. Samples are formatted using the Qwen2.5-VL chat template with a fixed system prompt. The causal language modeling loss (Eq.~\eqref{eq:loss}) is computed only over assistant-generated tokens, including structured tool calls and natural-language responses. System messages, user turns, and returned tool outputs are excluded from the loss through token masking. Sequences are limited to 8{,}192 tokens. To improve training efficiency, samples are packed into flattened sequences, with variable-length attention preventing interactions between tokens from different samples.

\paragraph{Image preprocessing.}
Input images are resized by the Qwen2.5-VL image processor. The total pixel count lies between 256 and 12{,}288 pixels, with aspect ratios preserved and image resolution constrained by the configured minimum and maximum pixel counts.

\paragraph{Optimization.}
We optimize the trainable parameters using AdamW~\cite{loshchilov2019decoupled} with a peak learning rate of $1 \times 10^{-5}$, weight decay of $0.01$, and gradient clipping at a maximum norm of $1.0$. The learning rate follows a cosine decay schedule with linear warmup over the first 10\% of optimization steps. Training is performed for one epoch on four GPUs with a per-device batch size of 1 and gradient accumulation over 64 steps, yielding an effective global batch size of 256 and approximately 816 optimization steps. No validation set is used for checkpoint selection, and the final checkpoint is used for all evaluations.

\paragraph{Compute.}
Training is conducted on four NVIDIA A100-SXM4-80GB GPUs using DeepSpeed ZeRO Stage 3 with CPU offloading of model parameters and optimizer states. We enable bfloat16 mixed precision, FlashAttention-2, and gradient checkpointing to reduce memory requirements. Training requires approximately 75 hours of wall-clock time, corresponding to 300 GPU hours.

\begin{table}[!h]
\centering
\caption{\textbf{Training configuration of \textsc{MMedAgent-Plus}.}}
\label{tab:training_config}
\small
\setlength{\tabcolsep}{6pt}
\renewcommand{\arraystretch}{1.15}
\begin{tabular}{l l}
\hline\hline
\textbf{Setting} & \textbf{Value} \\
\hline
\multicolumn{2}{l}{\textit{Model}} \\
Initialization & Qwen2.5-VL-7B-Instruct \\
Vision encoder & Frozen \\
Vision-language projector & Trained \\
Language decoder & Trained (full parameters) \\
\hline
\multicolumn{2}{l}{\textit{Data}} \\
Training data & $\mathcal{S}_1 \cup \mathcal{S}_2 \cup \mathcal{S}_3$, $|\mathcal{S}| = 208{,}928$ \\
Loss masking & System and user turns masked \\
Maximum sequence length & 8{,}192 tokens \\
Image pixels (min / max) & 256 / 12{,}288 \\
Sequence flattening & Enabled \\
\hline
\multicolumn{2}{l}{\textit{Optimization}} \\
Optimizer & AdamW \\
Peak learning rate & $1 \times 10^{-5}$ \\
Learning rate schedule & Cosine, 10\% linear warmup \\
Weight decay & 0.01 \\
Gradient clipping (max norm) & 1.0 \\
Epochs & 1 \\
Per-device batch size & 1 \\
Gradient accumulation steps & 64 \\
Effective batch size & 256 \\
Total optimization steps & 816 \\
\hline
\multicolumn{2}{l}{\textit{Compute}} \\
Hardware & 4 $\times$ NVIDIA A100-SXM4-80GB \\
Distributed strategy & DeepSpeed ZeRO-3, CPU offload of parameters and optimizer states \\
Precision & bfloat16 \\
Attention & FlashAttention-2 \\
Gradient checkpointing & Enabled \\
Training time & 75 h wall-clock (300 GPU hours) \\
\hline
\multicolumn{2}{l}{\textit{Reproducibility}} \\
Random seed & 42 \\
Data seed & 42 \\
\hline\hline
\end{tabular}
\end{table}

\subsection{Evaluation Protocol and Baselines}

\label{sec:supp_eval}

This section provides implementation details for the evaluation protocol described in Section~\ref{sec:evaluation}, including the LLM-as-a-judge configuration, scoring rubrics, task-specific metrics, and baseline settings.

\subsubsection{LLM-as-a-judge}

Final response quality is evaluated using GPT-5.4-mini~\cite{openai2026gpt54mini} (\texttt{gpt-5.4-mini-2026-03-17}). Faithfulness, completeness, and conciseness are assessed in separate judge calls using criterion-specific prompts to avoid combining the three judgments within a single evaluation. Each call returns a structured JSON response containing a brief rationale and an integer score from 1 to 5. Judge calls use deterministic decoding settings, and responses are cached by input hash to ensure consistent reuse of previously obtained evaluations.

For \textsc{MMedAgent-Plus}, the judge receives the tool name, original user request, terminal specialist output, generated final response, and indicators for non-text outputs and multi-tool workflows. The reference answer is additionally provided for completeness and conciseness evaluation but excluded from faithfulness evaluation. The judge is instructed to assess how accurately and effectively the planner communicates the specialist output, rather than independently evaluating the clinical correctness of that output. Faithfulness assesses consistency with the specialist output, while completeness and conciseness use the reference answer only to identify task-relevant information and the expected response scope.

\begin{figure}[!h]
\centering
\includegraphics[width=\linewidth]{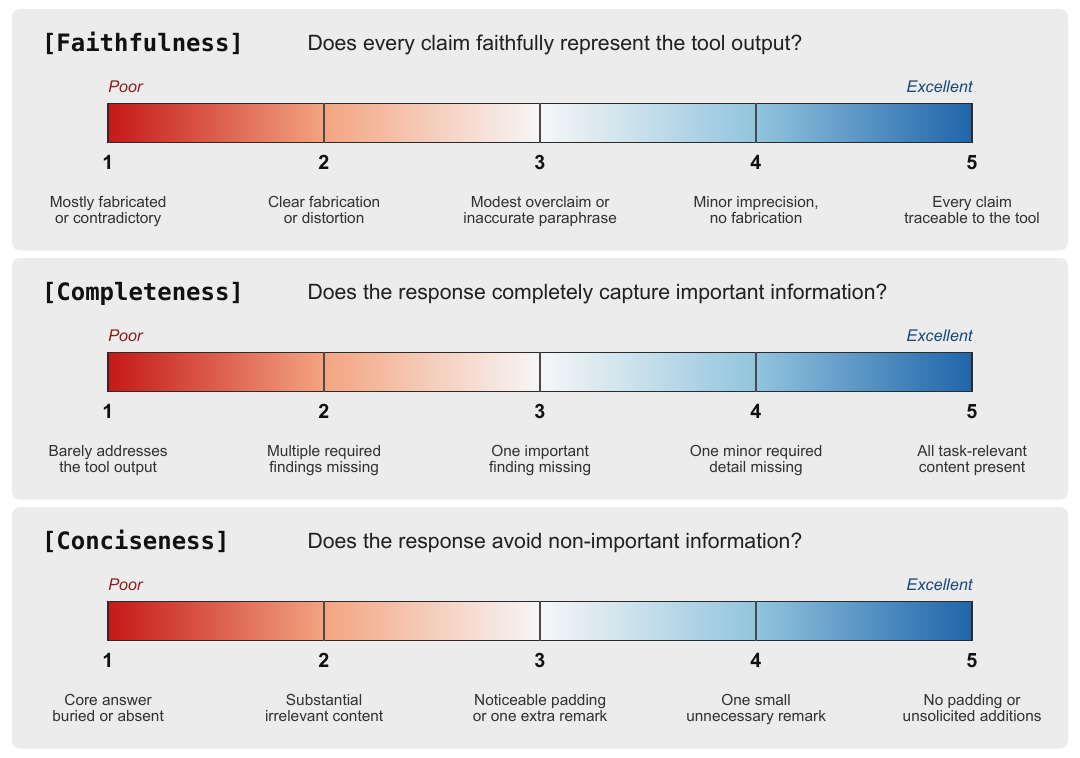}
\caption{\textbf{LLM-as-a-judge scoring rubric.} Each criterion is scored independently. Colour encodes quality from 1 (red) to 5 (blue).}
\label{fig:judge_rubric}
\end{figure}

The five-point scoring rubrics are illustrated in Figure~\ref{fig:judge_rubric}.

Each criterion-specific prompt includes three scored few-shot examples corresponding to scores of 5, 3, and 1. For faithfulness, the examples illustrate accurate restatement of a CONCH classification, a LLaVA-Rad response containing an exaggerated severity assessment and an unsupported finding, and a CONCH response incorrectly reporting a benign result as malignant. Completeness examples use a LLaVA-Rad report to illustrate full coverage, omission of cardiomegaly, and an incomplete response stating only that the lungs appear clear. Conciseness examples contrast a direct one-sentence CONCH classification with a response containing an unsolicited referral disclaimer and an excessively verbose response that obscures the classification result.

Special evaluation rules apply to non-text outputs and malformed final responses. For MedSAM, IterNet, HealthGPT, and UniGradICON, a brief acknowledgment that the requested output was produced is considered sufficient for the maximum score on all three criteria when no textual description is required. Responses that remain in tool-call format after specialist execution are classified as wrong-context responses and assigned the minimum score of 1 for each criterion without judge evaluation. The complete judge prompts and few-shot examples are provided with the evaluation code.

Direct-answer baselines are evaluated using the same judge model, scoring scales, and criterion-specific rubrics. Because these models do not receive specialist outputs, faithfulness is assessed against the original multimodal input rather than a returned tool result (Section~\ref{sec:baseline}).

\subsubsection{Task-specific metrics}

Task-specific evaluation metrics reported in Table~\ref{tab:e2e_metrics} are computed as follows.

ROUGE-L is calculated as the F1 score of the longest common subsequence after lowercasing and whitespace tokenization, without stemming. BLEU-4 is computed at the corpus level using add-one smoothing and the standard brevity penalty. Both metrics compare generated rationales or reports against their corresponding references.

CheXpert F1 is computed using a rule-based approximation of the 14 CheXbert observation labels. An observation is marked as present when a corresponding condition keyword appears in the text without a preceding negation cue within a 40-character window. Negation cues include \textit{no}, \textit{not}, \textit{without}, \textit{absence of}, \textit{free of}, \textit{denies}, \textit{negative for}. Micro-averaged F1 is then calculated over the resulting binary observation labels. RadGraph F1 is \texttt{F1RadGraphv2} from \texttt{rrg\_eval} with \texttt{reward\_level=partial}~\cite{jain2021radgraph}. For RaTE-NER, entity-level F1 is based on exact matching of entity strings and types. RadGraph F1 is additionally computed by comparing reconstructed entity strings with the corresponding source sentences.

Classification performance is measured using micro- and macro-averaged F1 from \texttt{scikit-learn}. Predicted and reference labels are normalized through case folding and removal of surrounding braces or quotation marks before comparison. For PMC-LLaMA, answer accuracy is computed by exact matching of extracted yes/no or multiple-choice answers after the same normalization.

Dice is $2|A\cap B|/(|A|+|B|)$ on binary masks thresholded at 127; two empty masks score 1. HD95 is the 95th percentile of the symmetric surface-to-surface Euclidean distance, obtained from the distance transform of each mask after a one-pixel erosion to isolate the boundary. If one mask is empty and the other is not, HD95 is set to the image diagonal. In the end-to-end setting, Dice averages over all 50 cases per segmentation tool and scores a missing mask as 0; HD95 averages only over cases that produce a mask, because 0 is a perfect match on that metric (Section~\ref{sec:end-to-end}).

\subsubsection{Baseline models and prompts}

\begin{center}
\small
\begin{tabular}{ll}
\hline
Model & Checkpoint \\
\hline
Qwen3-VL-8B-Instruct~\cite{bai2025qwen3vltechnicalreport} & \texttt{Qwen/Qwen3-VL-8B-Instruct} \\
LLaVA-Med v1.5~\cite{li2023llava} & \texttt{microsoft/llava-med-v1.5-mistral-7b} \\
GPT-5.4-mini~\cite{openai2026gpt54mini} & \texttt{gpt-5.4-mini-2026-03-17} \\
MedGemma-1.5-4B-IT~\cite{sellergren2026medgemma15technicalreport} & \texttt{google/medgemma-4b-it} \\
\hline
\end{tabular}
\end{center}

Locally hosted models use greedy decoding (\texttt{do\_sample=False}) with a maximum of 1{,}024 new tokens, matching the generation limit used for $\mathcal{M}$. GPT-5.4-mini uses the most deterministic decoding configuration supported by its API. For requests containing multiple images, LLaVA-Med receives only the first image, whereas Qwen3-VL-8B-Instruct and MedGemma-1.5-4B-IT receive all available images.

Direct-answer baselines receive the user request and associated images without access to specialist outputs, subsequent dialogue history, or reference answers. All baselines use the same system instruction:

\begin{verbatim}
You are a medical AI assistant.
Answer the user's query directly and concisely.
If an image is provided, use it to inform your answer.
Do not mention tool calls.
\end{verbatim}

The user request corresponds to the turn immediately preceding the reference tool invocation, with \texttt{<image>} placeholders aligned to the associated image files. This evaluation compares specialist-assisted response generation with direct-answer models operating without tool outputs. Qwen3-VL-8B-Instruct provides an additional comparison within the Qwen vision-language model family, using a newer and larger backbone than the Qwen2.5-VL-7B model underlying $\mathcal{M}$. However, differences in model architecture, instruction tuning, and available evidence prevent the observed performance gaps from being attributed solely to specialist tool access.

\subsection{Qualitative Examples}

\begin{figure}[!h]
\centering
\includegraphics[width=0.92\linewidth]{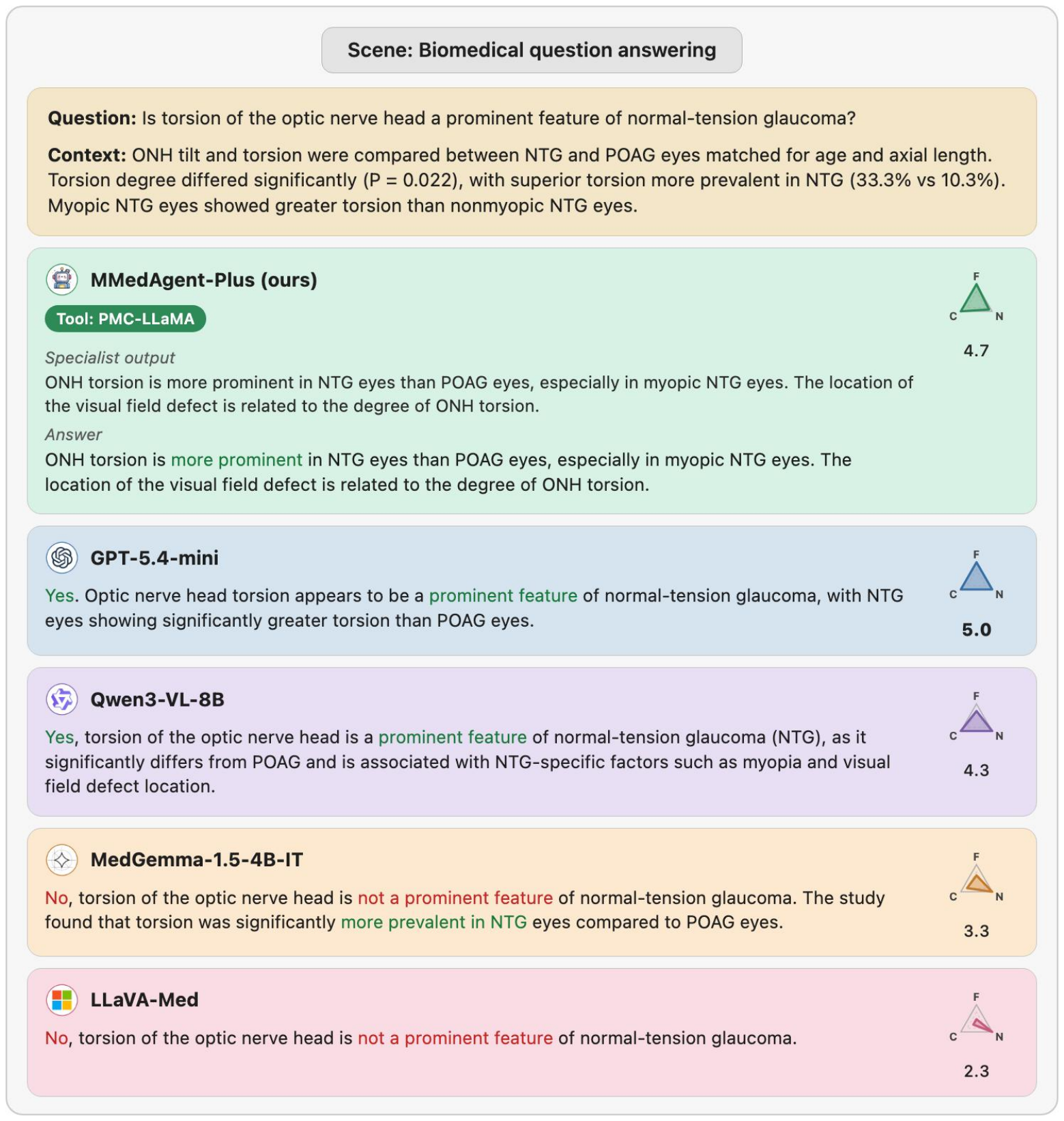}
\caption{\textbf{Biomedical question answering.} \textsc{MMedAgent-Plus} invokes PMC-LLaMA and correctly concludes that optic nerve head torsion is more prominent in normal-tension glaucoma, whereas MedGemma-1.5-4B-IT and LLaVA-Med contradict the cited evidence.}
\label{fig:qual_qa}
\end{figure}

\begin{figure}[!h]
\centering
\includegraphics[width=0.92\linewidth]{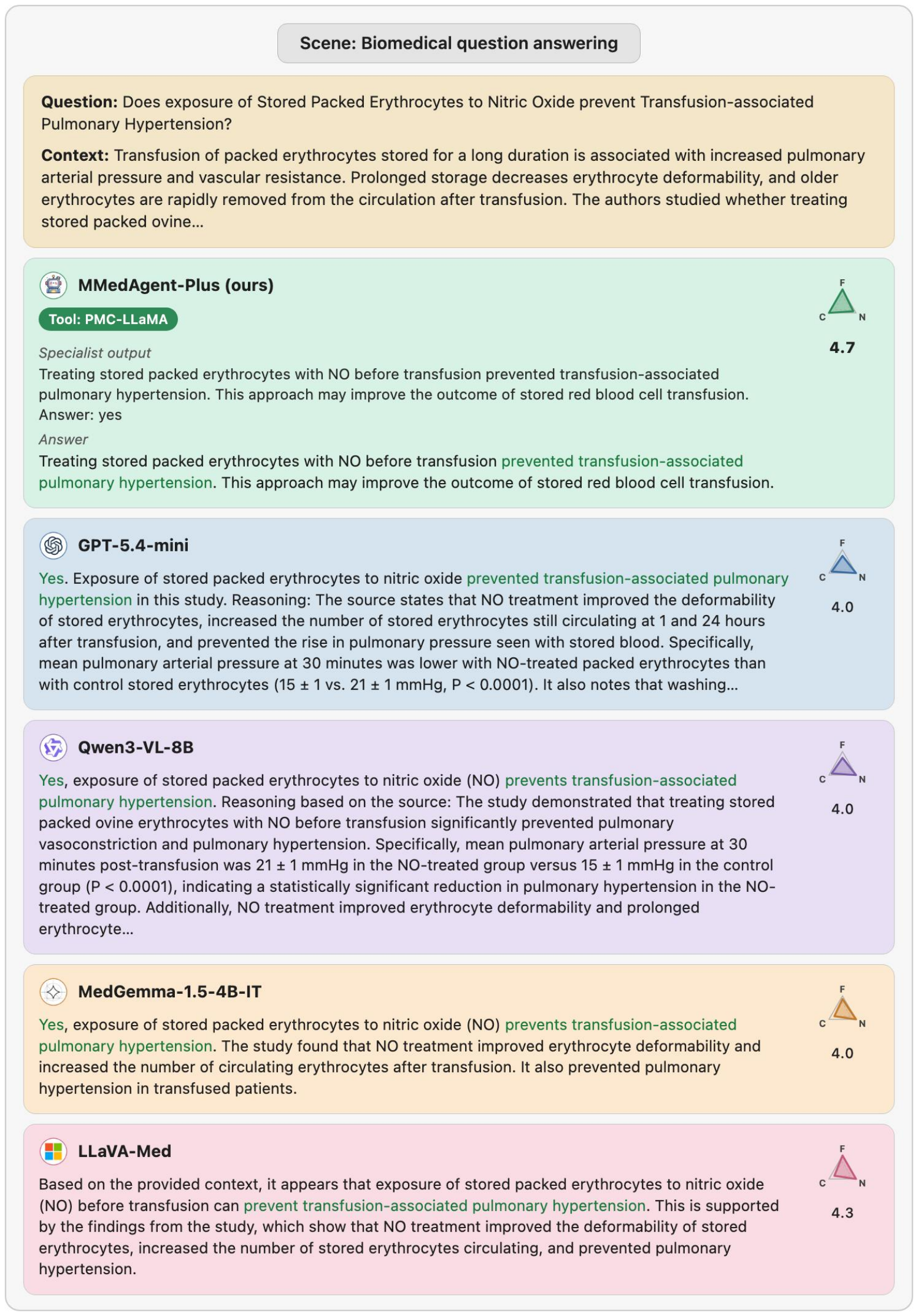}
\caption{\textbf{Biomedical question answering.} For a question on nitric-oxide treatment of stored erythrocytes, \textsc{MMedAgent-Plus} returns a concise PMC-LLaMA-grounded answer, while the direct-answer baselines produce verbose responses that fail our reference matching despite reaching the correct yes/no conclusion.}
\label{fig:qual_qa_2}
\end{figure}

\begin{figure}[!h]
\centering
\includegraphics[width=0.92\linewidth]{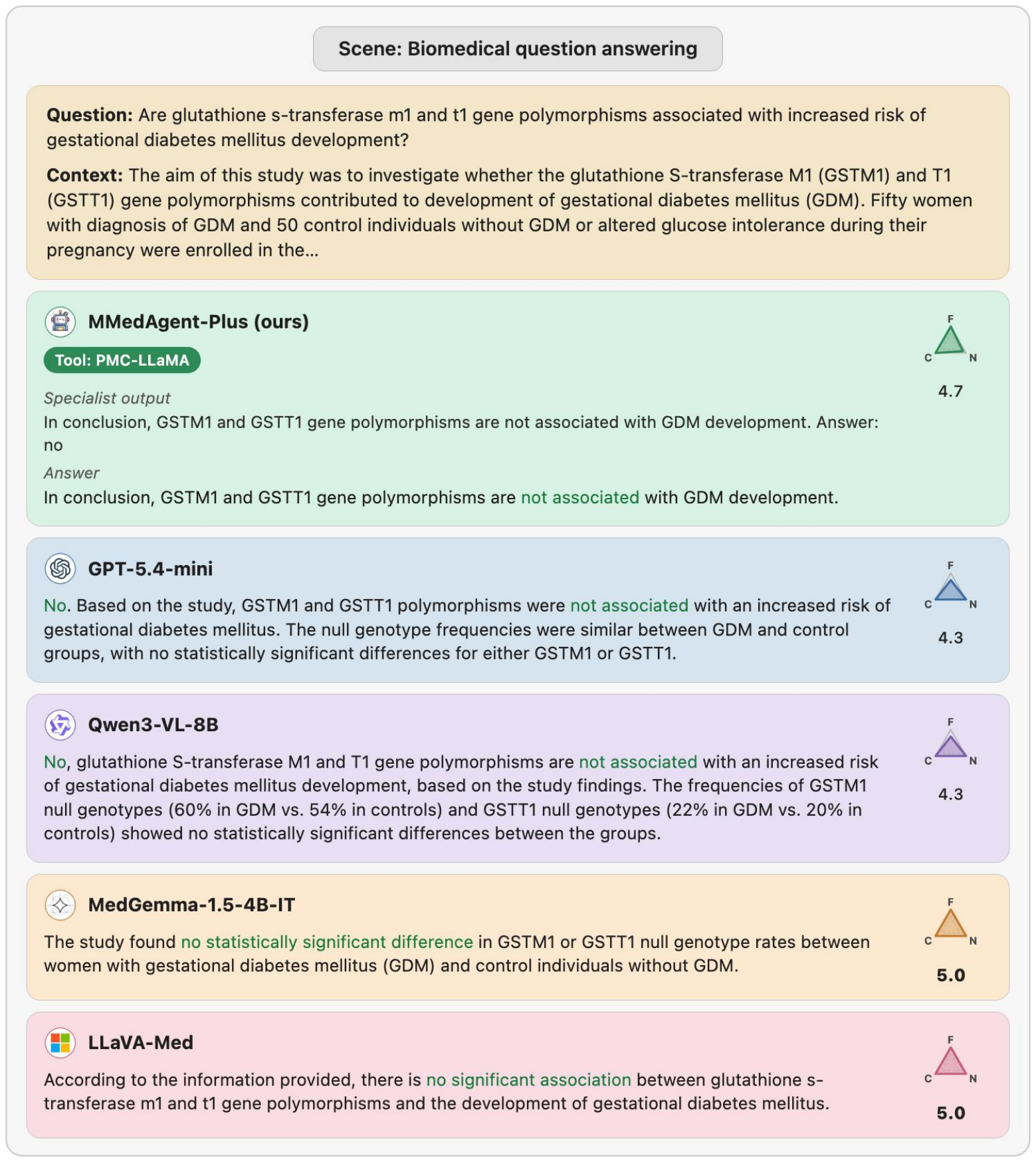}
\caption{\textbf{Biomedical question answering.} \textsc{MMedAgent-Plus} correctly reports that GSTM1 and GSTT1 polymorphisms are not associated with gestational diabetes mellitus; MedGemma-1.5-4B-IT and LLaVA-Med omit the explicit negative association required by the reference answer.}
\label{fig:qual_qa_3}
\end{figure}

\begin{figure}[!h]
\centering
\includegraphics[width=0.92\linewidth]{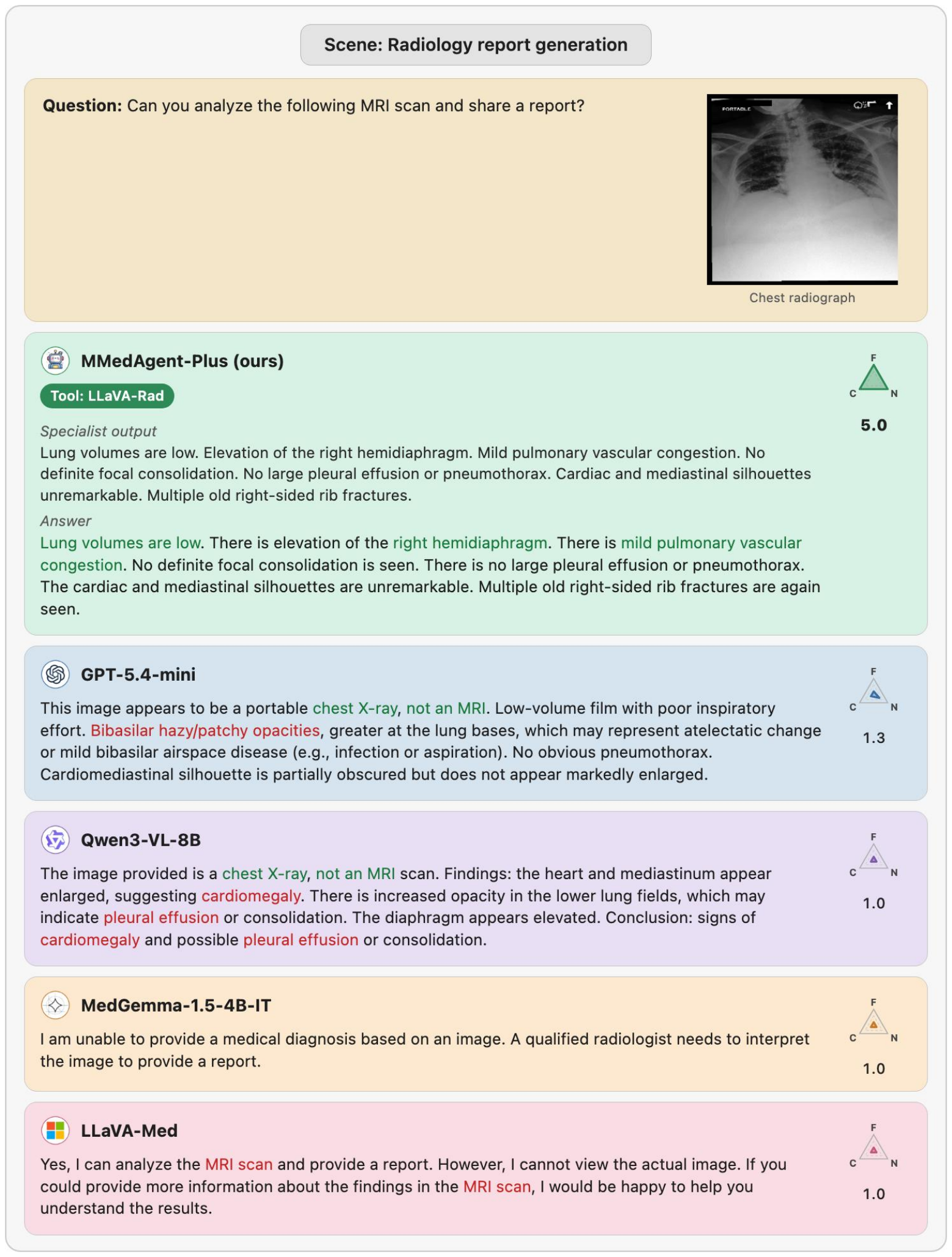}
\caption{\textbf{Radiology report generation.} Despite the user mislabeling a chest radiograph as an MRI, \textsc{MMedAgent-Plus} invokes LLaVA-Rad and produces an accurate structured report; direct-answer baselines misidentify the modality or decline to interpret the image.}
\label{fig:qual_report}
\end{figure}

\begin{figure}[!h]
\centering
\includegraphics[width=0.92\linewidth]{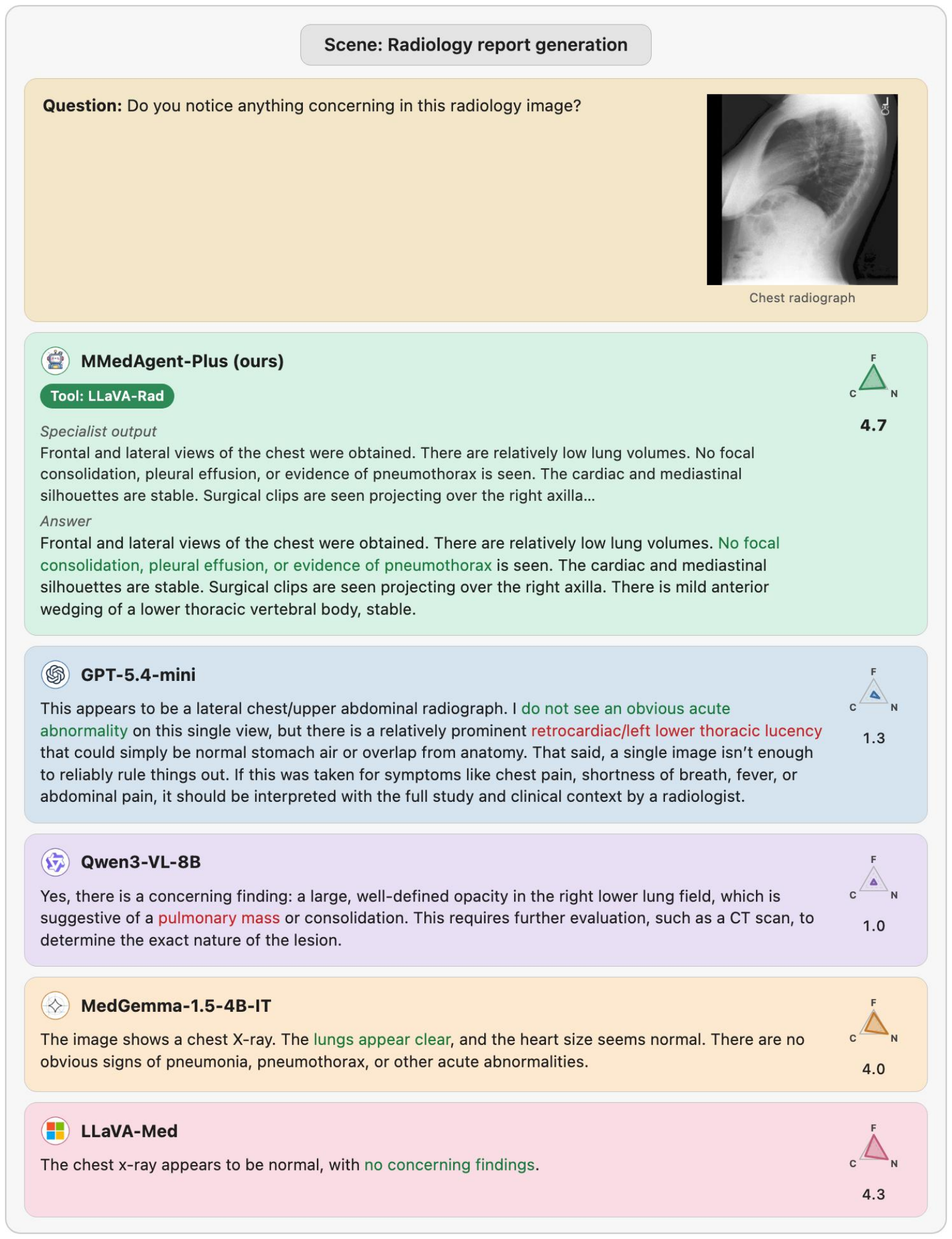}
\caption{\textbf{Radiology report generation.} On a lateral chest radiograph with low lung volumes, \textsc{MMedAgent-Plus} synthesizes a faithful LLaVA-Rad report, whereas Qwen3-VL-8B hallucinates a pulmonary mass and LLaVA-Med incorrectly calls the study normal.}
\label{fig:qual_report_2}
\end{figure}

\begin{figure}[!h]
\centering
\includegraphics[width=0.92\linewidth]{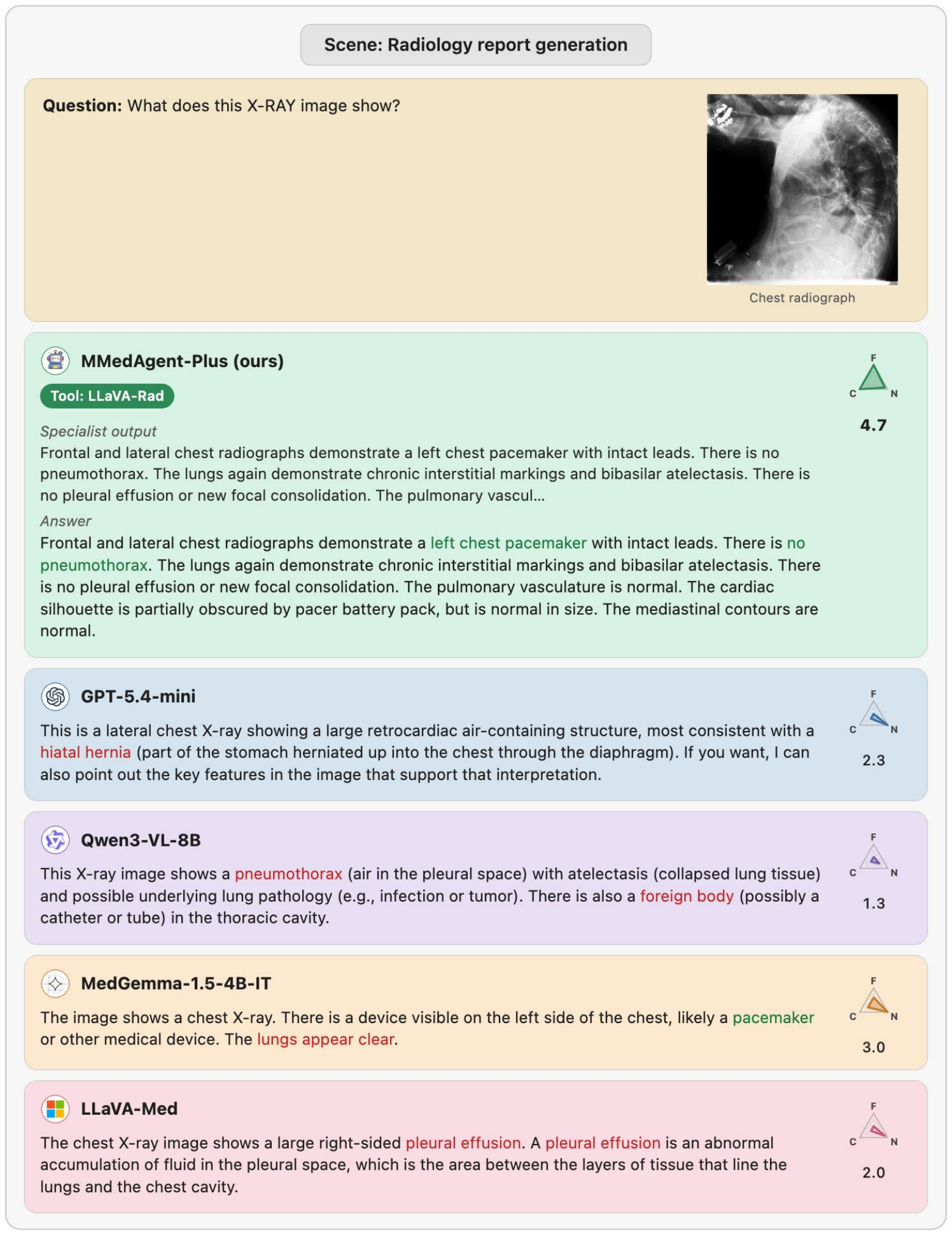}
\caption{\textbf{Radiology report generation.} For a pacemaker chest radiograph with chronic interstitial markings, \textsc{MMedAgent-Plus} preserves LLaVA-Rad findings on device leads and lung parenchyma; baselines misread the study as hiatal hernia, pneumothorax, or pleural effusion.}
\label{fig:qual_report_3}
\end{figure}

\begin{figure}[!h]
\centering
\includegraphics[width=0.92\linewidth]{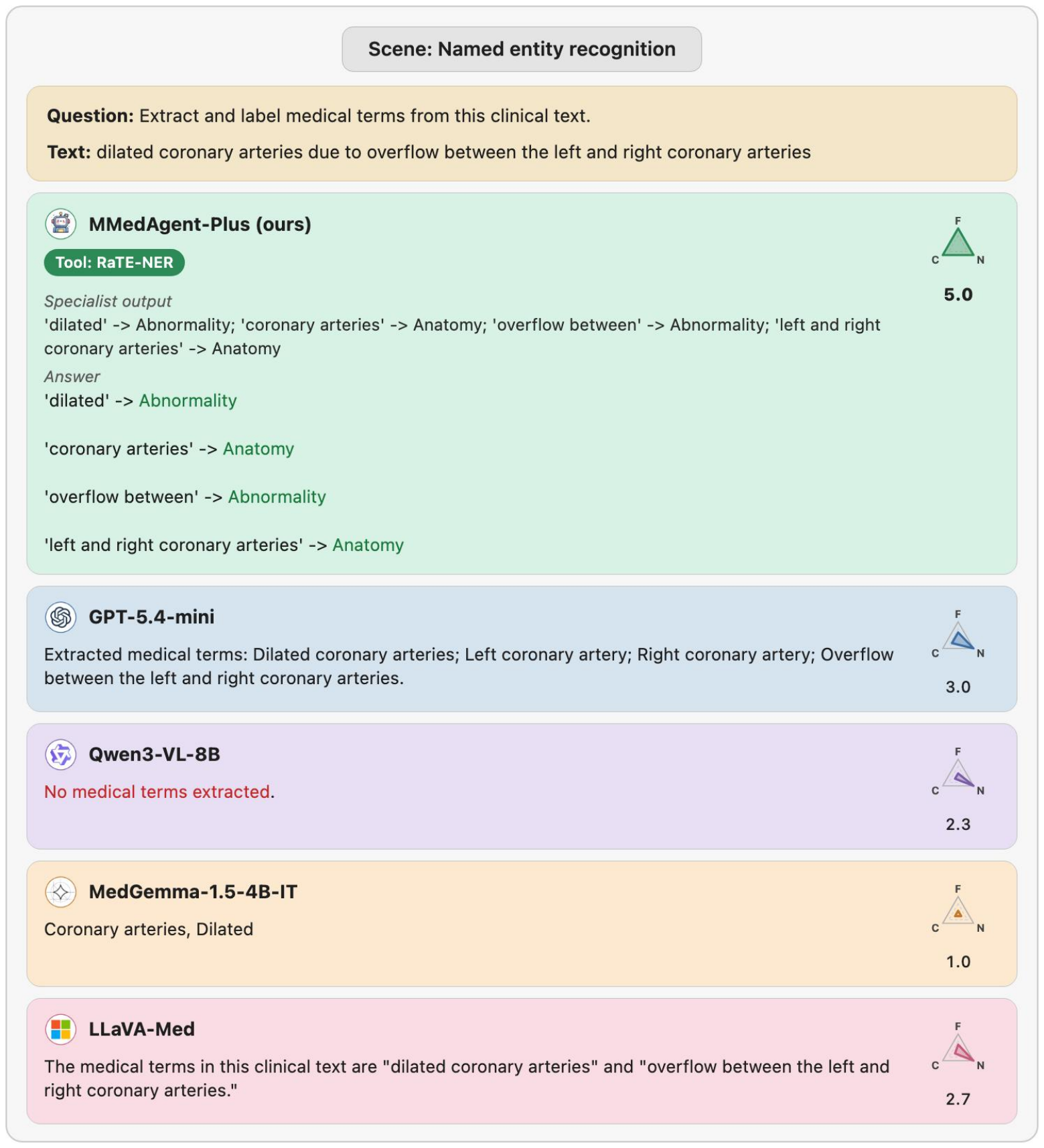}
\caption{\textbf{Named entity recognition.} \textsc{MMedAgent-Plus} reproduces RaTE-NER typed spans for coronary-artery anatomy and abnormality phrases; direct-answer baselines return unstructured term lists without the required entity-type labels.}
\label{fig:qual_ner}
\end{figure}

\begin{figure}[!h]
\centering
\includegraphics[width=0.92\linewidth]{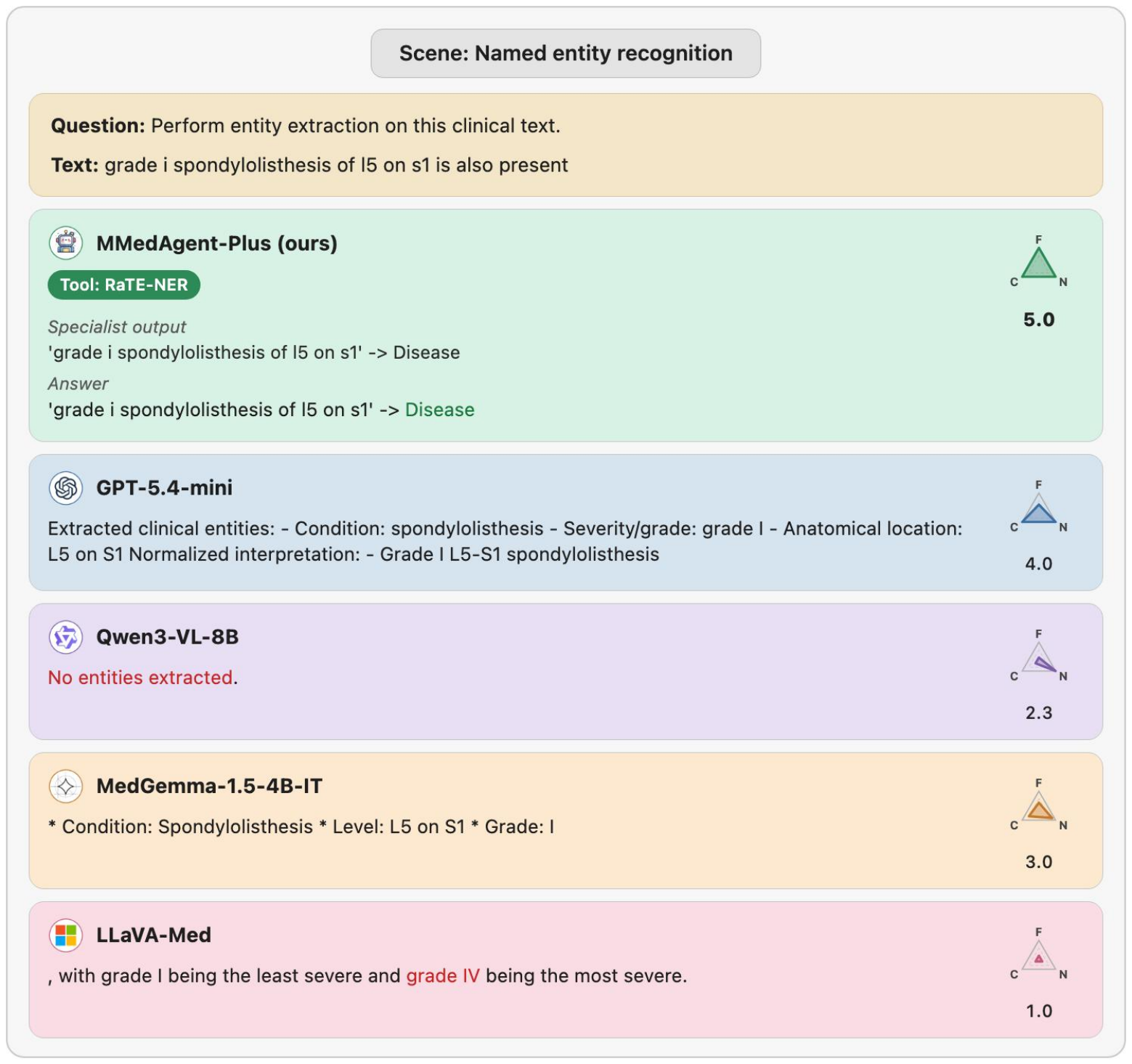}
\caption{\textbf{Named entity recognition.} \textsc{MMedAgent-Plus} identifies \textit{grade I L5 - S1 spondylolisthesis} as a single Disease entity per RaTE-NER, while baselines split or paraphrase the span without matching the specialist schema.}
\label{fig:qual_ner_2}
\end{figure}

\begin{figure}[!h]
\centering
\includegraphics[width=0.92\linewidth]{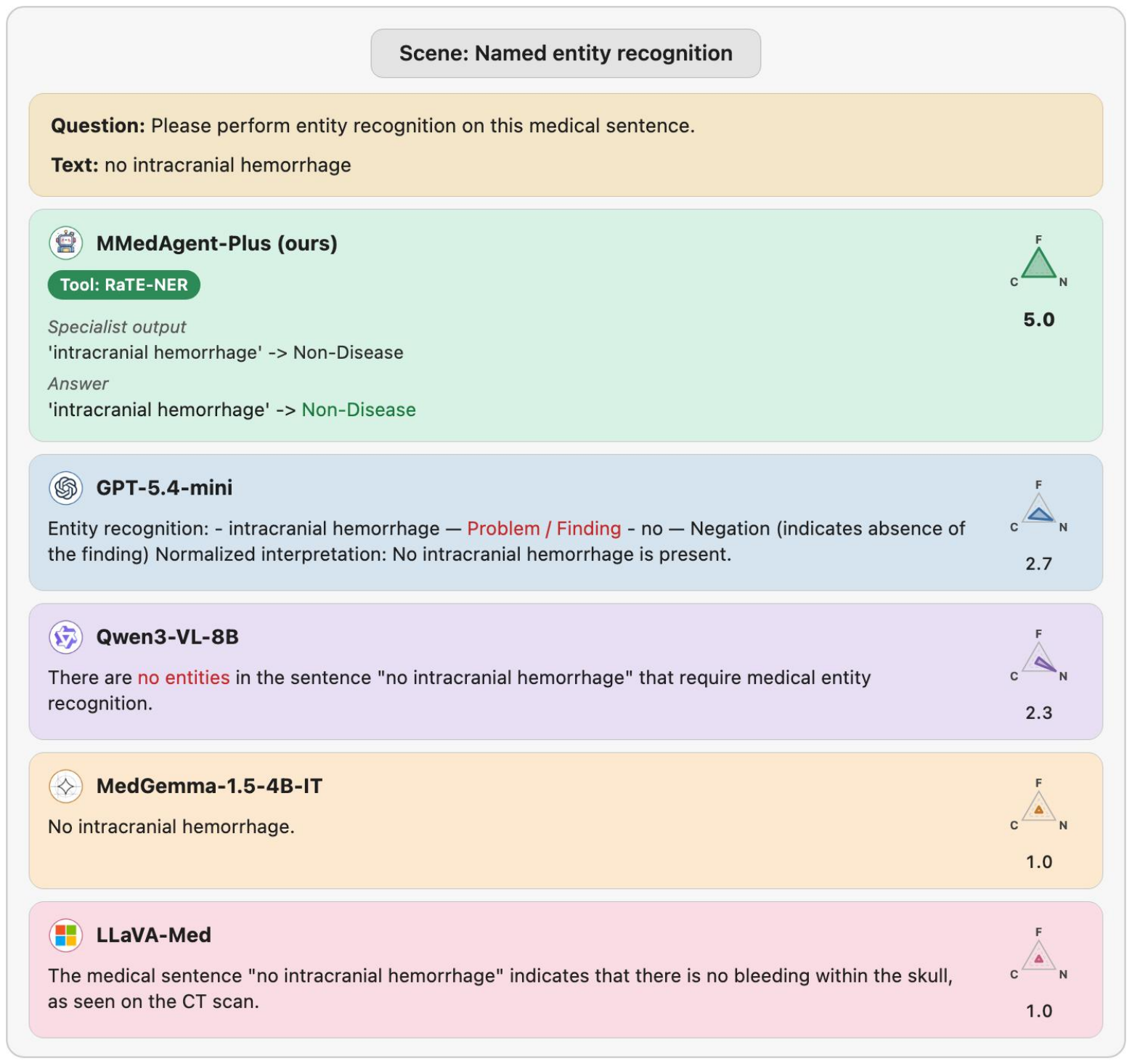}
\caption{\textbf{Named entity recognition.} On the negated finding \textit{no intracranial hemorrhage}, \textsc{MMedAgent-Plus} labels \textit{intracranial hemorrhage} as Non-Disease following RaTE-NER; baselines either skip entity extraction or produce free-text interpretations.}
\label{fig:qual_ner_3}
\end{figure}

\begin{figure}[!h]
\centering
\includegraphics[width=0.92\linewidth]{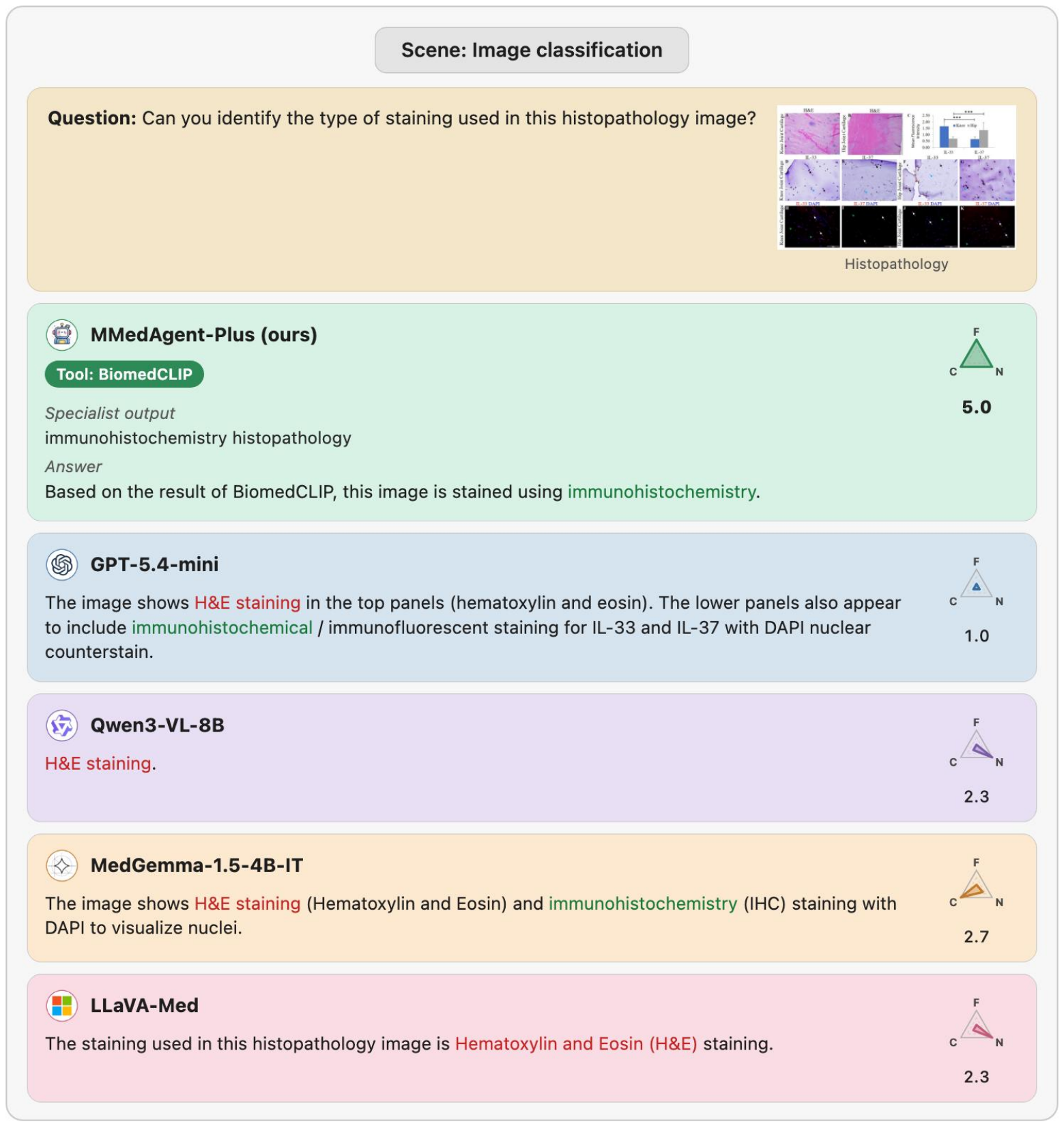}
\caption{\textbf{Image classification.} \textsc{MMedAgent-Plus} invokes BiomedCLIP and correctly identifies immunohistochemistry staining in a histopathology panel; direct-answer baselines default to H\&E or mixed-stain descriptions.}
\label{fig:qual_cls}
\end{figure}

\begin{figure}[!h]
\centering
\includegraphics[width=0.92\linewidth]{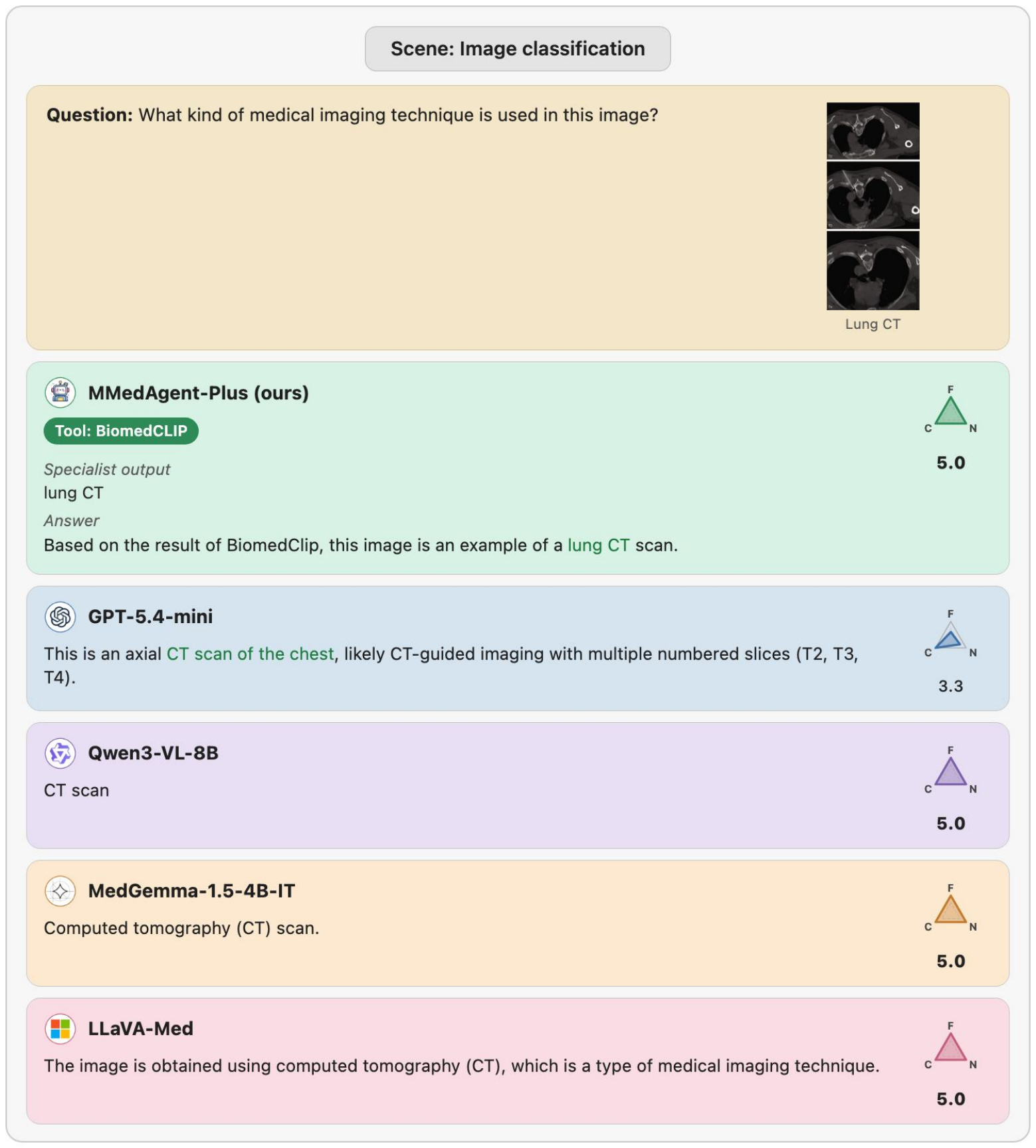}
\caption{\textbf{Image classification.} \textsc{MMedAgent-Plus} classifies the scan as lung CT via BiomedCLIP; baselines give generic ``CT scan'' answers that omit the organ-specific label required by the reference.}
\label{fig:qual_cls_2}
\end{figure}

\begin{figure}[!h]
\centering
\includegraphics[width=0.92\linewidth]{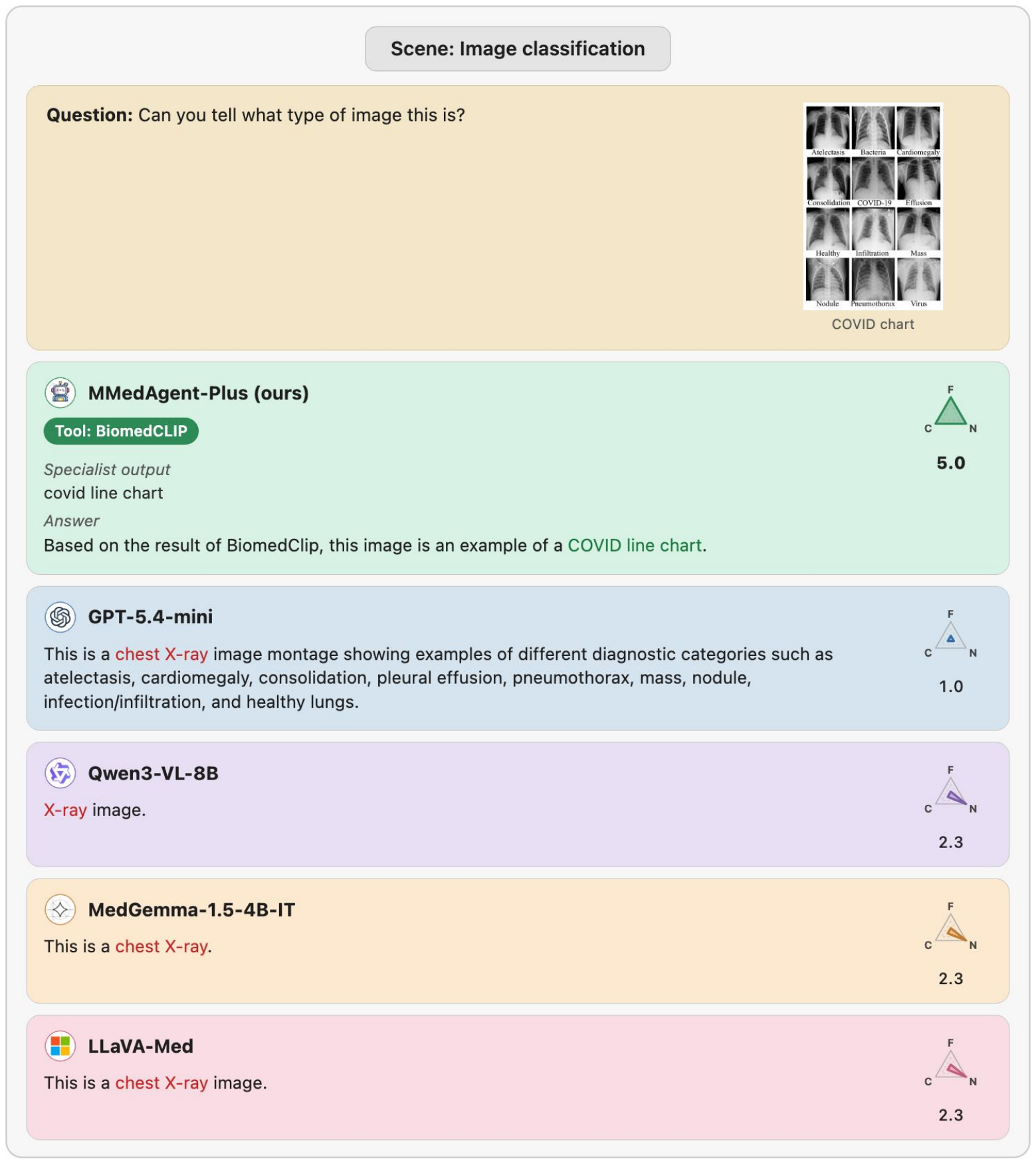}
\caption{\textbf{Image classification.} \textsc{MMedAgent-Plus} recognizes a COVID line chart through BiomedCLIP, whereas all direct-answer baselines misclassify the figure as a chest radiograph montage.}
\label{fig:qual_cls_3}
\end{figure}

\begin{figure}[!h]
\centering
\includegraphics[width=0.92\linewidth]{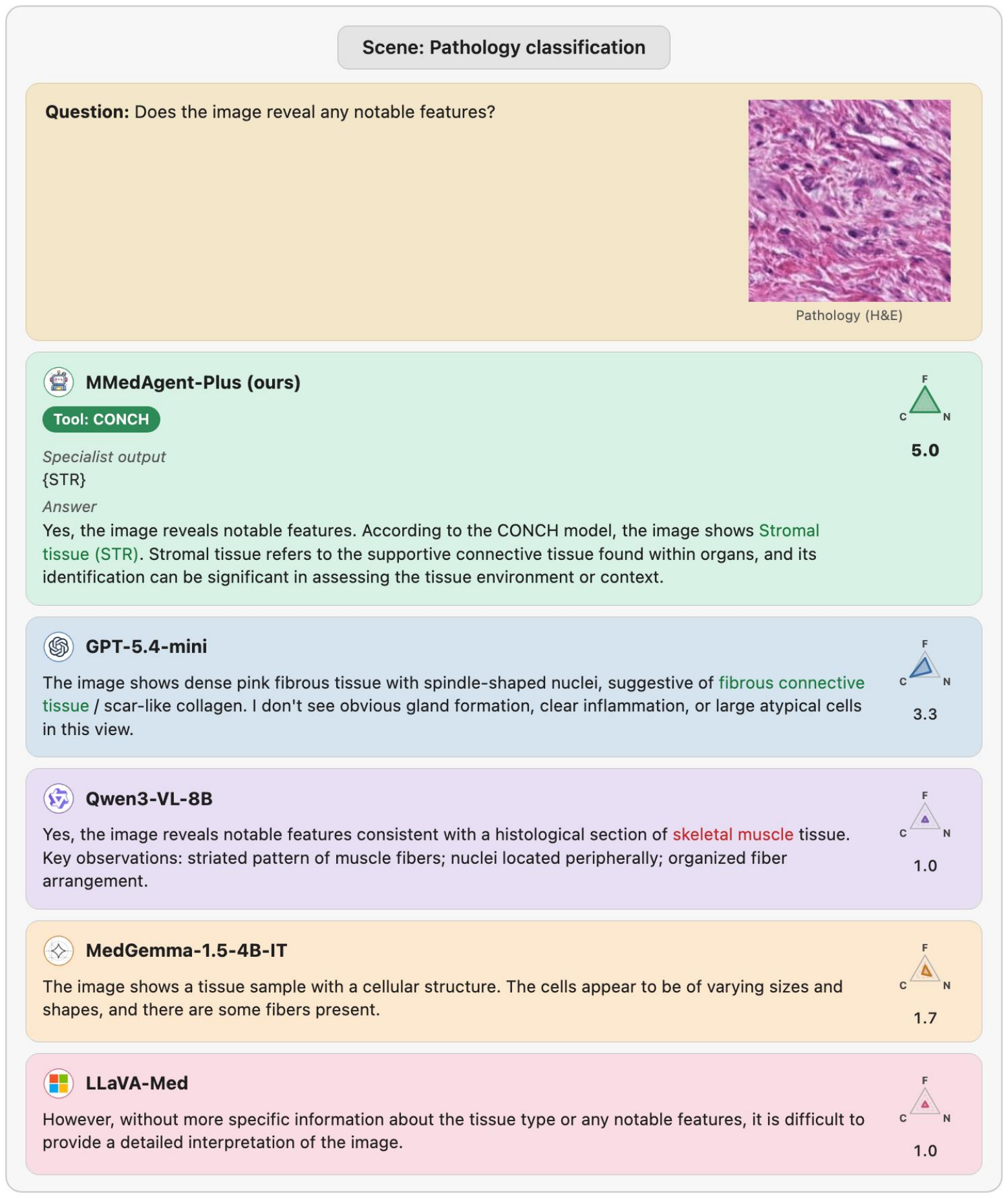}
\caption{\textbf{Pathology classification.} \textsc{MMedAgent-Plus} reports CONCH stromal-tissue (STR) predictions with clinical context; baselines provide generic histology descriptions without the tissue-class label.}
\label{fig:qual_path}
\end{figure}

\begin{figure}[!h]
\centering
\includegraphics[width=0.92\linewidth]{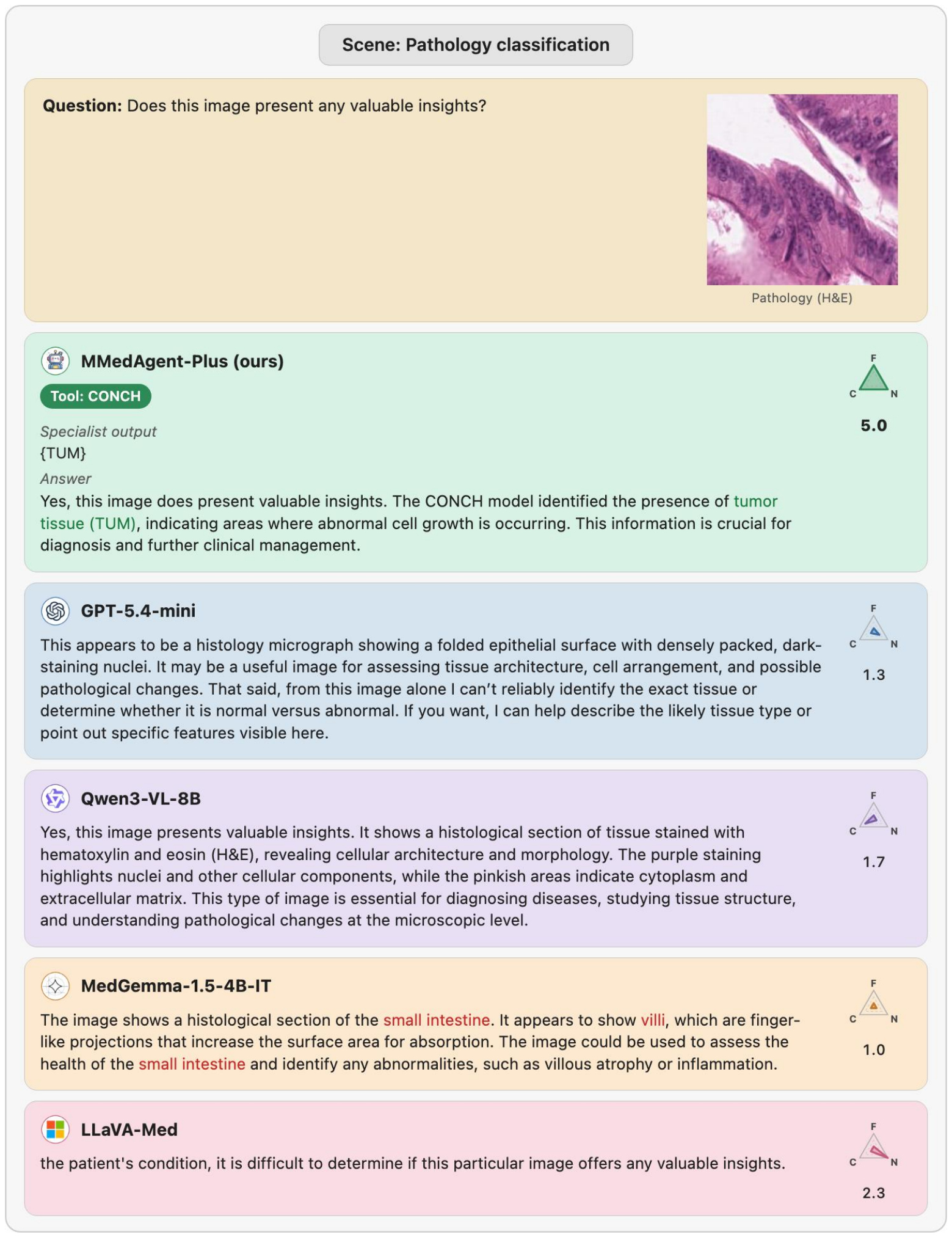}
\caption{\textbf{Pathology classification.} \textsc{MMedAgent-Plus} identifies tumor tissue (TUM) via CONCH; direct-answer baselines describe tissue architecture but fail to assign the pathology class.}
\label{fig:qual_path_2}
\end{figure}

\begin{figure}[!h]
\centering
\includegraphics[width=0.92\linewidth]{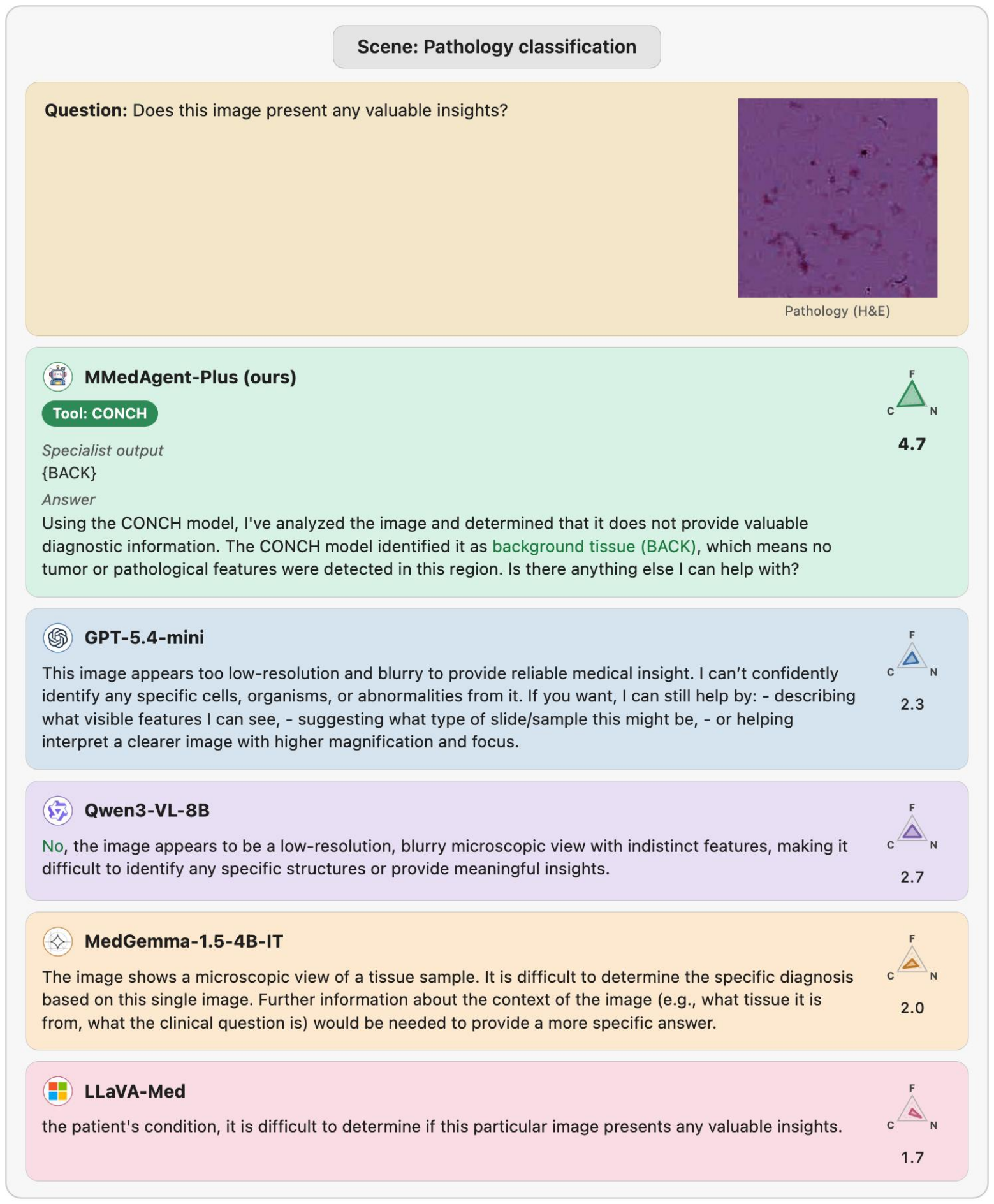}
\caption{\textbf{Pathology classification.} \textsc{MMedAgent-Plus} correctly reports CONCH background-tissue (BACK) with no pathological findings; baselines give non-committal or incorrect interpretations of the patch.}
\label{fig:qual_path_3}
\end{figure}

Figures~\ref{fig:qual_qa}-\ref{fig:qual_path_3} present representative held-out end-to-end cases across the five task categories included in the direct-answer baseline comparison (Section~\ref{sec:end-to-end}). Each example shows the user request, the output returned by the specialist tool, and side-by-side final responses from \textsc{MMedAgent-Plus} and four direct-answer vision-language models. Response quality is visualized using triangular plots representing faithfulness (F), completeness (C), and conciseness (N), with the mean score displayed below each plot.

